\pdfoutput=1

\documentclass[11pt]{article}

\usepackage{acl}

\usepackage{times}
\usepackage{latexsym}

\usepackage{rotating}
\usepackage{graphicx}
\usepackage{booktabs}
\usepackage{enumitem}
\usepackage{multirow}
\usepackage{makecell}
\usepackage{microtype}
\usepackage{lipsum}
\usepackage{stmaryrd}
\usepackage{amsmath}
\usepackage{multirow}
\usepackage{bbm}
\usepackage{amssymb}
\usepackage{booktabs}
\usepackage{adjustbox}
\usepackage{xspace}
\usepackage{algpseudocode}
\usepackage{algorithm}
\usepackage{caption}
\usepackage{subcaption}
\usepackage{color, colortbl}
\usepackage{tablefootnote}
\usepackage{bm}
\usepackage{colortbl}
\usepackage{paralist}
\usepackage{tabularx}

\usepackage{tikz}

\usetikzlibrary{patterns}

\DeclareRobustCommand{\legendsquare}[1]{%
  \tikz[baseline=(a.south)]{\node[#1, inner sep=.8ex, outer sep=0] (a) {};}%
}

\usepackage[T1]{fontenc}

\usepackage[utf8]{inputenc}

\usepackage{microtype}

\usepackage{inconsolata}

\usepackage{graphicx}
\usepackage{todonotes}

\usepackage{arydshln}

\title{SEACrowd: A Multilingual Multimodal Data Hub\\and Benchmark Suite for Southeast Asian Languages}

\renewcommand{\UrlFont}{\ttfamily\tiny}

\def\quad{\hskip0.35em\relax}

\author{
\normalsize{\textbf{Holy Lovenia$^{\bigstar,1,2}$}}\quad
\normalsize{\textbf{Rahmad Mahendra$^{\bigstar,3,2}$}}\quad
\normalsize{\textbf{Salsabil Maulana Akbar$^{\bigstar,2}$}}\quad
\\
\normalsize{\textbf{Lester James V. Miranda$^{\bigstar,4}$}}\quad
\normalsize{\textbf{Jennifer Santoso$^{\bigstar,5}$}}\quad
\normalsize{\textbf{Elyanah Aco$^{\bigstar,6}$}}\quad
\normalsize{\textbf{Akhdan Fadhilah$^{\bigstar,7}$}}\quad
\\
\normalsize{\textbf{Jonibek Mansurov$^{\bigstar,8}$}}\quad
\normalsize{\textbf{Joseph Marvin Imperial$^{\bigstar,9,10}$}}\quad
\normalsize{\textbf{Onno P. Kampman$^{\bigstar,11}$}}\quad
\\
\normalsize{\textbf{Joel Ruben Antony Moniz$^{\bigstar,6}$}}\quad
\normalsize{\textbf{Muhammad Ravi Shulthan Habibi$^{\bigstar,3,2}$}}\quad
\normalsize{\textbf{Frederikus Hudi$^{\bigstar,12,13}$}}\quad
\\
\normalsize{\textbf{Railey Montalan$^{\bigstar,1}$}}\quad
\normalsize{\textbf{Ryan Ignatius$^{6}$}}\quad
\normalsize{\textbf{Joanito Agili Lopo$^{14}$}}\quad
\normalsize{\textbf{William Nixon$^{15}$}}\quad
\\
\normalsize{\textbf{Börje F. Karlsson$^{16}$}}\quad
\normalsize{\textbf{James Jaya$^{6}$}}\quad
\normalsize{\textbf{Ryandito Diandaru$^{6}$}}\quad
\normalsize{\textbf{Yuze Gao$^{17}$}}\quad
\normalsize{\textbf{Patrick Amadeus$^{15}$}}\quad
\\
\normalsize{\textbf{Bin Wang$^{17}$}}\quad
\normalsize{\textbf{Jan Christian Blaise Cruz$^{8,18}$}}\quad
\normalsize{\textbf{Chenxi Whitehouse$^{19}$}}\quad
\normalsize{\textbf{Ivan Halim Parmonangan$^{20}$}}\quad
\\
\normalsize{\textbf{Maria Khelli$^{15}$}}\quad
\normalsize{\textbf{Wenyu Zhang$^{17}$}}\quad
\normalsize{\textbf{Lucky Susanto$^{21}$}}\quad
\normalsize{\textbf{Reynard Adha Ryanda$^{22}$}}\quad
\\
\normalsize{\textbf{Sonny Lazuardi Hermawan$^{23}$}}\quad
\normalsize{\textbf{Dan John Velasco$^{18}$}}\quad
\normalsize{\textbf{Muhammad Dehan Al Kautsar$^{15}$}}\quad
\\
\normalsize{\textbf{Willy Fitra Hendria$^{6}$}}\quad
\normalsize{\textbf{Yasmin Moslem$^{24}$}}\quad
\normalsize{\textbf{Noah Flynn$^{25}$}}\quad
\normalsize{\textbf{Muhammad Farid Adilazuarda$^{8}$}}\quad
\\
\normalsize{\textbf{Haochen Li$^{6}$}}\quad
\normalsize{\textbf{Johanes Lee$^{15}$}}\quad
\normalsize{\textbf{R. Damanhuri$^{26}$}}\quad
\normalsize{\textbf{Shuo Sun$^{17}$}}\quad
\normalsize{\textbf{Muhammad Reza Qorib$^{27}$}}\quad
\\
\normalsize{\textbf{Amirbek Djanibekov$^{8}$}}\quad
\normalsize{\textbf{Wei Qi Leong$^{1}$}}\quad
\normalsize{\textbf{Quyet V. Do$^{28}$}}\quad
\normalsize{\textbf{Niklas Muennighoff$^{29}$}}\quad
\\
\normalsize{\textbf{Tanrada Pansuwan$^{19}$}}\quad
\normalsize{\textbf{Ilham Firdausi Putra$^{6}$}}\quad
\normalsize{\textbf{Yan Xu$^{30,28}$}}\quad
\normalsize{\textbf{Ngee Chia Tai$^{1}$}}\quad
\\
\normalsize{\textbf{Ayu Purwarianti$^{6,31}$}}\quad
\normalsize{\textbf{Sebastian Ruder$^{32}$}}\quad
\normalsize{\textbf{William Tjhi$^{1}$}}\quad
\normalsize{\textbf{Peerat Limkonchotiwat$^{\bigstar,33}$}}\quad
\\
\normalsize{\textbf{Alham Fikri Aji$^{\bigstar,8}$}}\quad
\normalsize{\textbf{Sedrick Keh$^{\bigstar,34}$}}\quad
\normalsize{\textbf{Genta Indra Winata$^{\bigstar,36,2}$}}\quad
\normalsize{\textbf{Ruochen Zhang$^{\bigstar,35}$}}\quad
\\
\normalsize{\textbf{Fajri Koto$^{\bigstar,8,2}$}}\quad
\normalsize{\textbf{Zheng-Xin Yong$^{\bigstar,35}$}}\quad
\normalsize{\textbf{Samuel Cahyawijaya$^{\bigstar,32,28,2}$}}
\\
\parbox{0.95\textwidth}{\centering
\small{$^{1}$AI Singapore}\quad
\small{$^{2}$IndoNLP}\quad
\small{$^{3}$Universitas Indonesia}\quad
\small{$^{4}$Allen Institute for Artificial Intelligence}\quad
\small{$^{5}$RevComm, Inc.}\quad
\small{$^{6}$Independent Researcher}\quad
\small{$^{7}$Tohoku University}\quad
\small{$^{8}$MBZUAI}\quad
\small{$^{9}$University of Bath}\quad
\small{$^{10}$National University Philippines}\quad
\small{$^{11}$MOH Office for Healthcare Transformation (MOHT)}\quad
\small{$^{12}$NAIST}\quad
\small{$^{13}$Works Applications Lab}\quad
\small{$^{14}$Universitas Gadjah Mada}\quad
\small{$^{15}$Institut Teknologi Bandung}\quad
\small{$^{16}$Beijing Academy of Artificial Intelligence (BAAI)}\quad
\small{$^{17}$Institute for Infocomm Research, A*STAR}
\small{$^{18}$Samsung Research Philippines}\quad
\small{$^{19}$University of Cambridge}\quad
\small{$^{20}$Queensland University of Technology}\quad
\small{$^{21}$Monash University Indonesia}\quad
\small{$^{22}$Imperial College London}\quad
\small{$^{23}$Independent Design Engineer}\quad
\small{$^{24}$Bering Lab}\quad
\small{$^{25}$Amazon}\quad
\small{$^{26}$Universitas Diponegoro}\quad
\small{$^{27}$NUS}\quad
\small{$^{28}$HKUST}\quad
\small{$^{29}$Contextual AI}\quad
\small{$^{30}$Huawei Noah's Ark Lab}\quad
\small{$^{31}$Prosa.ai}\quad
\small{$^{32}$Cohere}\quad
\small{$^{33}$VISTEC}\quad
\small{$^{34}$Toyota Research Institute}\quad
\small{$^{35}$Brown University}\quad
\small{$^{36}$Capital One}
}
\\
\small{\textbf{$^{\bigstar}$Major contributors}} \\
\\
}

\begin{document}
\maketitle
\begin{abstract}

Southeast Asia (SEA) is a region characterized by rich linguistic diversity and cultural variety, with over 1,300 indigenous languages and a population of 671 million people.
However, the performance of contemporary AI models for SEA languages is compromised by a significant lack of representation of texts, images, and auditory datasets from SEA.
Evaluating models for SEA languages is challenging due to the scarcity of high-quality datasets, compounded by the predominance of English training data, which raises concerns regarding potential cultural misrepresentation. To address these challenges, we introduce SEACrowd, a collaborative initiative that consolidates a comprehensive resource hub\footnote{\url{https://seacrowd.github.io/seacrowd-catalogue/}} to bridge the resource gap by providing standardized corpora and benchmarks\footnote{\url{https://github.com/SEACrowd/seacrowd-datahub/}} in nearly 1,000 SEA languages across three modalities.
We assess the performance of AI models on 36 indigenous languages across 13 tasks included in SEACrowd, offering valuable insights into the current AI landscape in SEA.
Furthermore, we propose strategies to facilitate greater AI advancements, maximizing potential utility and resource equity for the future of AI in Southeast Asia.

\end{abstract}

\section{Introduction}
\label{sec:introduction}

Despite Southeast Asia (SEA) being home to 1,300 indigenous languages (18\% of the world's languages) and 671 million people (8.75\% of the world's population), the representation of texts, images, and audio datasets from this region is significantly lacking in machine learning models. This deficiency adversely affects the model quality for SEA languages. The language coverage of SEA languages in two common pre-training resources, Common Crawl\footnote{\url{https://commoncrawl.github.io/cc-crawl-statistics/plots/languages}} and C4~\cite{xue-etal-2021-mt5}, is extremely limited, with only 2.36\% (in 11 languages) and 10.62\% (in 11 languages), respectively.
In modalities beyond text, the representation is even more limited.
For instance, Common Voice, one of the largest multilingual speech corpora, includes six SEA indigenous languages~\cite{conneau21_interspeech, ardila-etal-2020-common}, and LAION-5B, one of the largest multilingual vision-language (VL) corpora, includes 12 SEA indigenous languages~\cite{schuhmann2022laion}.
Datasets for other SEA indigenous languages exist, but are often scattered, insufficiently documented, or varied in quality and formatting, thereby making access and usage challenging~\cite{cahyawijaya-etal-2023-nusacrowd,joshi-etal-2020-state,aji-etal-2023-current}.



In terms of evaluation, the sparse availability of high-quality test sets for these languages also complicates evaluating models 
for SEA languages.
Despite there being 1,300+ languages in the SEA region, prior works ~\cite{winata-etal-2023-nusax,cahyawijaya-etal-2021-indonlg,koto-koto-2020-towards,zhang2024m3exam,SeaEval,damonlpsg2023seallm,leong2023bhasa,yong-etal-2023-prompting} have only evaluated fewer than 10 SEA languages collectively.
The actual performance of current models on most SEA languages remains largely unknown.


Moreover, the dominance of Anglocentric training data may result in cultural bias when generating texts, images, or audio in underrepresented SEA languages~\cite{sogaard2022ban, talat2022reap}. 
Further, \citet{durmus2023towards, alkhamissi2024investigating, cahyawijaya2024high} have shown that the learned representations in large language models (LLMs) often fail to reflect local cultural values in SEA~\cite{koto2024indoculture, liu2024multilingual, adilazuarda2024measuring}. This raises concerns about the ability of current LLMs to generate natural, high-quality texts for this region. In addition, the discrepancy in language support
creates language barriers in technological access and risks marginalizing minority groups who do not speak the dominant language.

\begin{figure*}[t]
    \centering
    \includegraphics[width=0.8\linewidth]{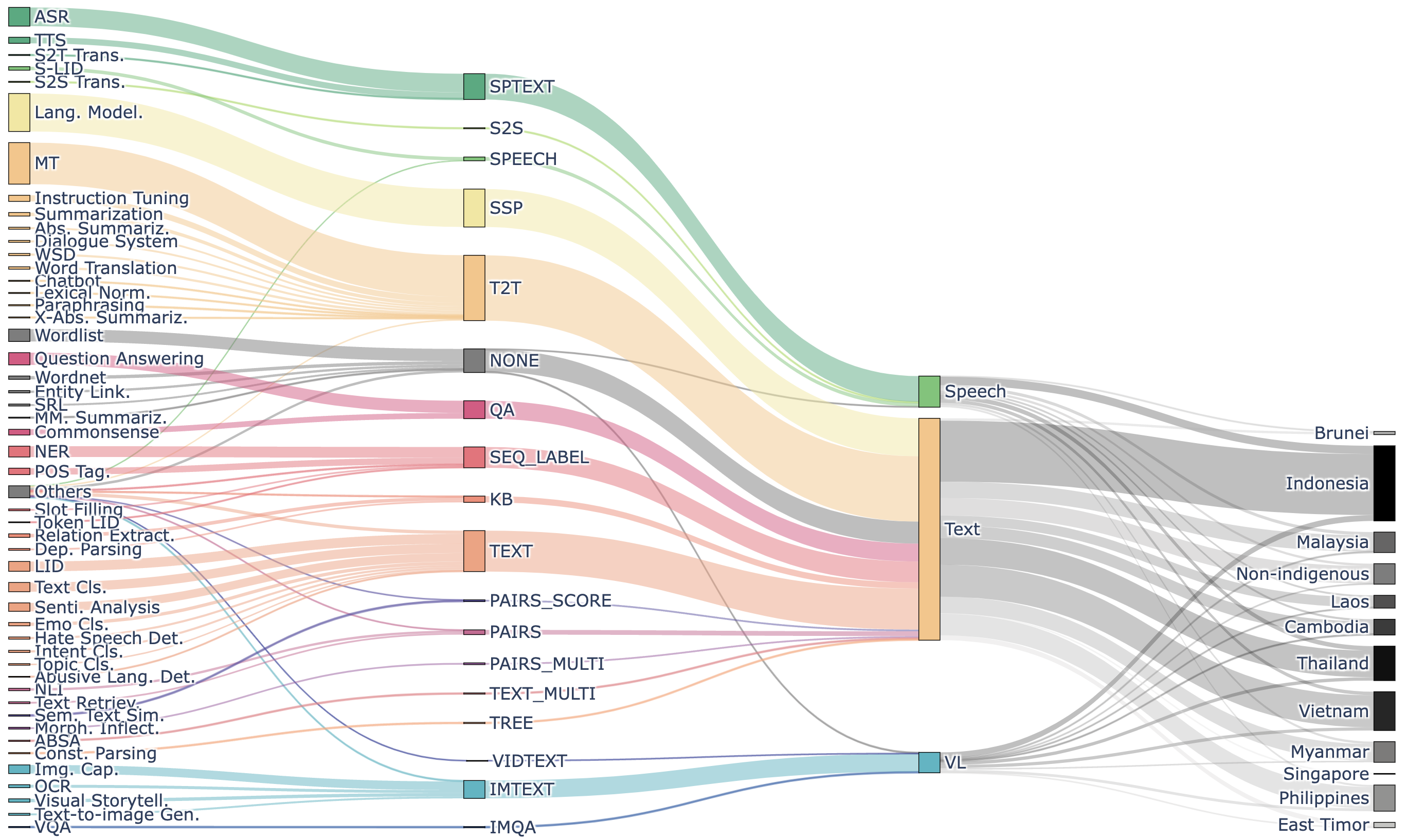}
    \caption{Mapping between tasks, schemas, modalities, and language regions across 498 datasheets in SEACrowd.}
    \label{fig:datasets-in-seacrowd-sankey}
    \vspace{-6pt}
\end{figure*}

In this work, we investigate the current AI progress for SEA languages by addressing the challenges of resources, evaluation, and generation quality. Our contributions are three-fold:

\begin{compactitem}
    \item We bridge the resource gap by centralizing and standardizing $\sim$500 corpora in nearly 1,000 SEA languages in SEACrowd, a comprehensive and standardized resource center, across three modalities: text, image, and audio. 
    \item We close the evaluation gap in SEA languages with the SEACrowd Benchmarks, which cover 38 SEA indigenous languages on 13 tasks across 3 modalities, providing insights into the performance of a diverse spectrum of AI models. Further, our study reveals that the generative outputs of existing LLMs exhibit a closer resemblance to ``translationese'' rather than natural data in nine SEA languages.
    \item We offer insights and strategies for the future development of AI in SEA.
\end{compactitem}


\section{SEACrowd}
\label{sec:seacrowd-initiative}

SEACrowd represents the first comprehensive AI dataset collection initiative for SEA, developed through a collaborative effort among researchers and engineers primarily based in the SEA region. As addressed in \S\ref{sec:introduction}, resource scarcity and the scattered nature of the data are crucial challenges in SEA. SEACrowd addresses these issues through two primary contributions: 1) \textbf{consolidating datasheets} to enhance data discoverability; and 2) \textbf{standardizing dataloaders} for easier use, especially in multiple dataset loading. We also follow data provenance practices \citep{longpre2023data} to preserve the proprietary rights of dataset owners.

\paragraph{Consolidating datasheets}
We invited contributors to submit datasheet forms \cite{gebru2021datasheets} for publicly available datasets across all modalities including text, audio, and image in SEA languages and/or cultures.
These datasheets include detailed information about each dataset, such as data subset(s), description, task, language, license, URL access, annotation method(s), annotation validation, relevant publications, publication venue, and data splits. For each submission, we manually verify and correct it as necessary to ensure datasheet accuracy.

\paragraph{Standardizing dataloaders} 
For each approved datasheet, we created a standardized dataloader wrapper to facilitate ready-to-use data access since only 38.4\% of the consolidated data sources were originally hosted on Hugging Face\footnote{\url{https://huggingface.co/}}. To support diverse task types, we carefully designed the standardized \texttt{seacrowd} schema to support different data structures and modalities (see Appendix~\ref{app:task-schema}). We also adhere to data provenance practices \citep{longpre2023data} and document the relevant metadata (e.g., license) in the dataloaders. Furthermore, we engaged with data owners and successfully converted three private datasets into public ones. 

These efforts have culminated in 498 datasheets in SEACrowd Catalogue and 399 dataloaders in SEACrowd Data Hub (\S\ref{sec:seacrowd-framework}). Notably, our centralized data repository 
covers $\sim$1,000 SEA languages, underscoring the extensive linguistic diversity captured by SEACrowd. We elaborate on the SEACrowd dataset statistics in \S\ref{sec:datasets-in-seacrowd}. SEACrowd's contribution guidelines, progression details, and reviewing procedure are in Appendix~\ref{app:contributing-to-seacrowd}, \ref{app:seacrowd-progression}, and \ref{app:reviewing-in-seacrowd}.

\subsection{SEACrowd Catalogue \& Data Hub}
\label{sec:seacrowd-framework}

SEACrowd comprises two interconnected platforms: \href{https://seacrowd.github.io/seacrowd-catalogue/}{SEACrowd Catalogue}\footnote{SEACrowd Catalogue is also present in \href{https://docs.google.com/spreadsheets/d/1ibbywsC1tQ_sLPX8bUAjC-vrTrUqZgZA46W_sxWw4Ss/edit?usp=sharing}{\texttt{csv} format}.}
and \href{https://github.com/SEACrowd/seacrowd-datahub/}{SEACrowd Data Hub}.
These platforms work in tandem to consolidate the datasheet submissions and provide a standardized pipeline for SEACrowd. Specifically, Catalogue houses the datasheets (metadata), while Data Hub stores the standardized dataloaders and \href{https://pypi.org/project/seacrowd/}{the \texttt{seacrowd} library}\footnote{All codes are available under Apache License 2.0.} for the schemas and configurations (Appendix~\ref{app:task-schema}). These systems share information on the datasheets and dataloaders, allowing users to seamlessly explore and utilize them.



\begin{figure*}[t]
    \centering
    \begin{minipage}[b]{0.75\linewidth}
        \centering
        \begin{subfigure}[b]{\linewidth}
            \centering
            \includegraphics[width=\linewidth,trim={0, 2em, 0, 0}, clip]{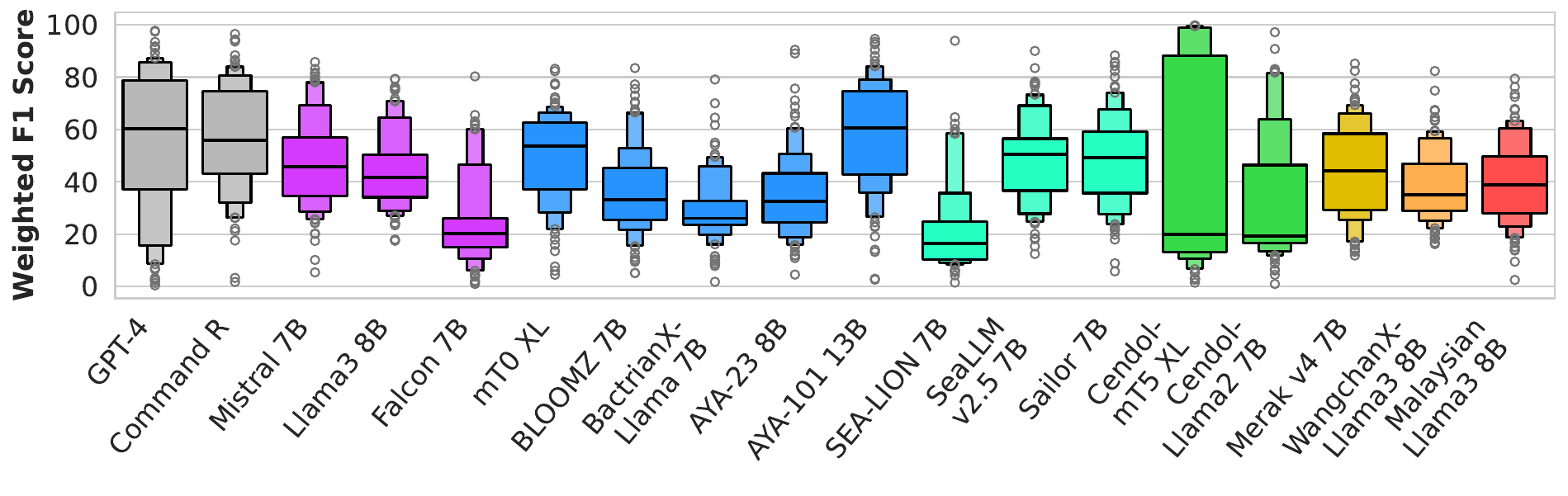}
            \caption{NLU evaluation}
            \label{fig:sea-nlu-overall-results}
        \end{subfigure}
        \begin{subfigure}[b]{\linewidth}
            \centering
            \includegraphics[width=\linewidth]{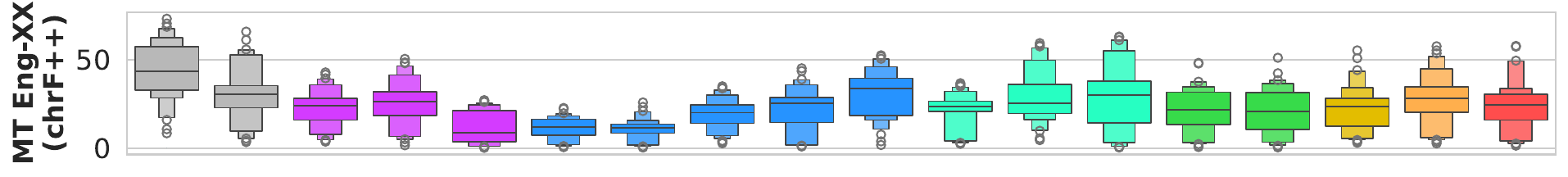}
            \includegraphics[width=\linewidth]{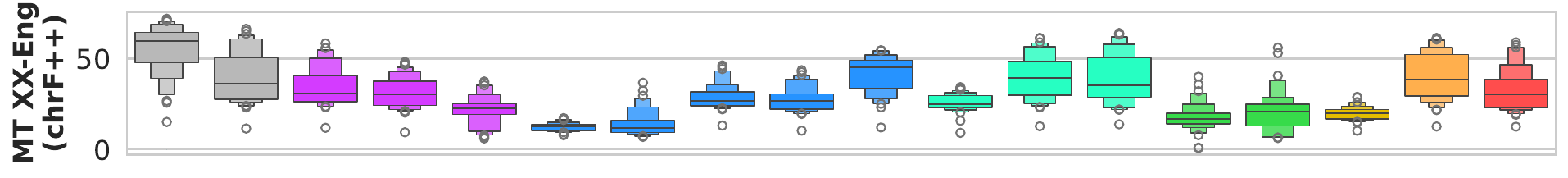}
            \includegraphics[width=\linewidth]{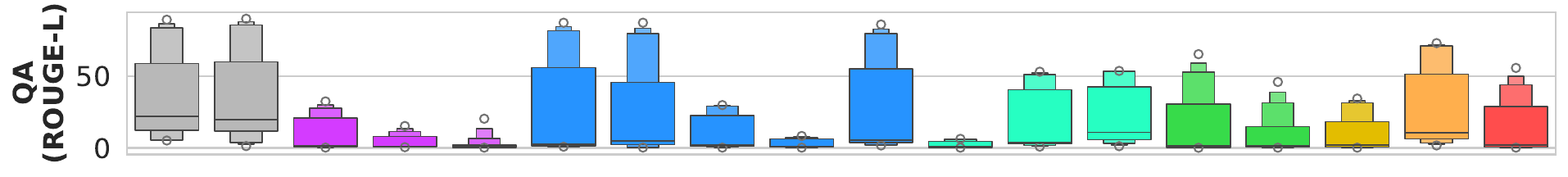}
            \hspace*{0.2cm}\includegraphics[width=\linewidth,trim={0, 2em, 0, 0}, clip]{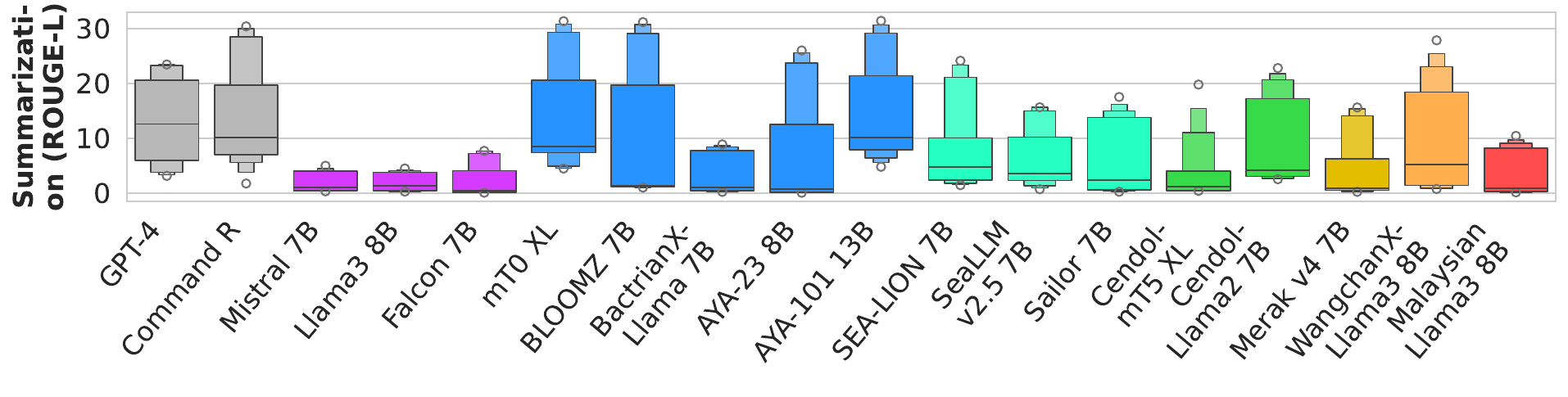}
            \caption{NLG evaluation}
            \label{fig:sea-nlg-overall-results}
        \end{subfigure}
        \caption{Zero-shot model performance across NLU and NLG tasks in SEA languages.}
        \label{fig:sea-nlp-overall-results}
        \vspace{-6pt}
    \end{minipage}
    \hfill
    \begin{minipage}[b]{0.22\linewidth}
    \centering
        \resizebox{1.15\linewidth}{!}{
        \begin{tabular}{l p{1.2cm}}
          \toprule
          \textbf{Model} & \textbf{Gini} $\downarrow$ \\
          \toprule
            \textit{Commercial} & \\
            \hspace{0.4cm}GPT-4 & \underline{0.155} \\
            \hspace{0.4cm}Command-R & 0.184 \\
            \midrule
            \textit{English} & \\
            \hspace{0.4cm}Mistral & 0.159 \\
            \hspace{0.4cm}Llama3 & \underline{0.131} \\
            \hspace{0.4cm}Falcon & 0.238 \\
            \midrule
            \textit{Multilingual} & \\
            \hspace{0.4cm}mT0 & 0.131 \\
            \hspace{0.4cm}BLOOMZ & 0.228 \\
            \hspace{0.4cm}BactrianX-Llama & 0.163 \\
            \hspace{0.4cm}AYA-23 & 0.183 \\
            \hspace{0.4cm}AYA-101 & \textbf{\underline{0.095}} \\
            \midrule
            \textit{SEA regional} & \\
            \hspace{0.4cm}SEA-LION & 0.204 \\
            \hspace{0.4cm}SeaLLM v2.5 & \underline{0.116} \\
            \hspace{0.4cm}Sailor & 0.145 \\
            \midrule
            \textit{SEA country} & \\
            \hspace{0.4cm}Cendol-mT5 & 0.378 \\
            \hspace{0.4cm}Cendol-Llama2 & 0.267 \\
            \hspace{0.4cm}Merak v4 & 0.199 \\
            \hspace{0.4cm}WangchanX-Llama3 & \underline{0.153} \\
            \hspace{0.4cm}Malaysian Llama3 & 0.179 \\
          \bottomrule
        \end{tabular}
        }
        \captionof{table}{Language equity across baselines based on Gini coefficient weighted by population ($\tau = 0.5$).}
        \label{tab:lang-equity}
        \vspace{-6pt}
    \end{minipage}
\end{figure*}

\subsection{Datasets in SEACrowd}
\label{sec:datasets-in-seacrowd}

SEACrowd consolidates 498 datasheets with diverse tasks in SEA languages and provides standardized access through dataloaders to 399 of them. As shown in Figure~\ref{fig:datasets-in-seacrowd-sankey}, approximately 81\% of the datasets in SEACrowd are textual data, with the remaining $\sim$8\% and $\sim$11\% being VL and speech, respectively. 
The complete list of SEA indigenous languages covered by SEACrowd and their mapping to the relevant SEA regions are provided in Appendix~\ref{app:lang-under-study}. Around $\sim$53\% of the datasets have a commercially permissive license.

A total of 83 tasks are provided in SEACrowd with a breakdown of 66 in NLP (e.g., abusive language detection, intent classification, instruction tuning, named entity recognition, etc.), 10 in VL (image-to-text generation, sign language recognition, video captioning, etc.), and 7 in speech (e.g., automatic speech recognition, text-to-speech, speech emotion recognition, and others). These tasks are then standardized into 20 dataloader schemas described in Appendix~\ref{app:task-schema}. Further discussion regarding resources in SEACrowd is in \S\ref{sec:resource-gaps}.

\section{SEACrowd Benchmarks}
\label{sec:sea-evaluation}

To understand the capability of state-of-the-art models, we conduct comprehensive evaluations of existing LLMs, VLMs, and speech models from various architectures and training approaches. To construct a benchmark suite\footnote{\url{https://github.com/SEACrowd/seacrowd-experiments}}, we select a subset of the dataset that has been manually annotated and/or validated from the data presented in \S\ref{sec:datasets-in-seacrowd}. More details regarding the data subsets, baselines, and prompts used for the evaluations are given in Appendix~\ref{app:sea-eval-data}, \ref{app:sea-eval-baselines}, and \ref{app:sea-eval-prompts}.

\subsection{Datasets}

\paragraph{NLP}
Our natural language understanding (NLU) benchmark consists of 131 data subsets and 7 tasks: sentiment analysis, topic classification, natural language inference (NLI), commonsense reasoning, exam-style multiple-choice question answering (QA), culture understanding, and reading comprehension. It covers English (\textsc{eng}) and 33 SEA indigenous languages.

We utilize 100 data subsets for the natural language generation (NLG) benchmark, which covers machine translation (MT) between English and SEA languages from both directions, summarization, as well as extractive or abstractive question answering, covering 27 SEA indigenous languages.

\paragraph{Speech}
We employ 19 automatic speech recognition (ASR) data subsets to evaluate the capability of speech models in 15 SEA indigenous languages.

\paragraph{VL}
We assess the models on image captioning using four data subsets in 4 SEA indigenous languages, i.e., Filipino (\textsc{fil}), Indonesian (\textsc{ind}), Thai (\textsc{tha}), and Vietnamese (\textsc{vie}). This disparity in the evaluation scale is due to the fact that only a few datasets in SEACrowd are VL datasets, and even fewer are annotated by humans.

\subsection{Baselines}

Complete details regarding the model architectures, model sizes, seen languages, corresponding publications, and other aspects are in Appendix~\ref{app:sea-eval-baselines}.

\paragraph{NLP}

To evaluate the zero-shot performance of instruction-tuned LLMs on SEA languages, we benchmark two commercial, i.e., GPT-4~\cite{openai2024gpt4} and Command-R\footnote{\url{https://docs.cohere.com/docs/command-r}}, and 17 open-source baselines, the majority of which are $\sim$7B-13B parameters. We categorize the open-source baselines according to the language(s) coverage in pre-training and/or instruction tuning, i.e., 1) \textbf{English}: Llama3~\cite{touvron2023llama}, Mistral~\cite{jiang2023mistral}, and Falcon~\cite{falcon40b}; 2) \textbf{Multilingual}:
AYA-101, AYA-23~\cite{ustun2024aya}, mT0, BLOOMZ~\cite{muennighoff2022crosslingual}, and BactrianX-Llama~\cite{li2023bactrian}; 3) \textbf{SEA regional}: SEA-LION~\cite{sea_lion_2023}, Sailor~\cite{dou2024sailor}, and SeaLLM~\cite{damonlpsg2023seallm}; and 4) \textbf{SEA country-specific}: Cendol-mT5, Cendol-Llama2~\cite{cahyawijaya2024cendol}, and Merak~\cite{Merak} from Indonesia, WangchanX-Llama3~\cite{phatthiyaphaibun2024wangchanlion} from Thailand, and Malaysian-Llama3\footnote{\url{https://huggingface.co/mesolitica/malaysian-llama-3-8b-instruct-16k}} from Malaysia.

\paragraph{Speech}
We evaluate the zero-shot performance of state-of-the-art \textbf{multilingual pre-trained} speech models in transcribing speech in SEA languages. Specifically, we consider Whisper v3~\cite{pmlr-v202-radford23a}, MMS 1B~\cite{pratap2024scaling}, and Seamless M4T v2~\cite{communication2023seamlessm4t}, which have shown proficiency in accurately transcribing multiple languages without fine-tuning. Additionally, we include models that are \textbf{fine-tuned on specific language(s)}, SEA or English, based on 1) Wav2Vec2 XLSR~\cite{conneau21_interspeech} and 2) XLS-R~\cite{babu2021xlsr}, known for their cross-lingual speech representation learning by pre-training on raw speech waveforms across diverse languages, with XLS-R offering broader language coverage, and 3) Whisper, which leverages weakly supervised pre-training on spectrograms of speech in diverse languages. The specific fine-tuned models are evaluated: XLSR on \textsc{ind}, \textsc{jav}, \textsc{sun}; XLSR and Whisper on Indonesian (\textsc{ind}); XLSR and Whisper on Thai (\textsc{tha}); XLS-R on Tagalog (\textsc{tgl}); XLS-R on Burmese (\textsc{mya}); XLS-R and Whisper on Khmer (\textsc{khm}); and XLSR on English (\textsc{eng}). See Appendix~\ref{app:sea-eval-baselines} for details.

\paragraph{VL}
We consider state-of-the-art VLMs primarily trained on \textbf{English} pre-training and instruction-following data: LLaVA~\cite{liu2023llava, liu2023improvedllava}, InstructBLIP~\cite{dai2024instructblip}, and Idefics2~\cite{laurençon2024matters}, and VLMs trained in a \textbf{multilingual} manner: mBLIP~\cite{geigle2023mblip} and PaliGemma~\cite{gemmateam2024gemma}, to assess their image captioning ability in SEA languages.

\subsection{Experimental Settings}

We conduct all evaluations in a zero-shot fashion. We employ 3 prompt templates in English for each NLU task and 1 for each NLG task. We utilize the weighted F1 score to measure the model performance on NLU tasks and n-gram reference-based metrics, i.e., chrF++~\cite{popovic-2015-chrf, popovic-2017-chrf} and ROUGE-L~\cite{lin-2004-rouge}, on NLG tasks. As for VL, aside from a prompt template in English, we also use a prompt template in the respective SEA indigenous language per data subset. We report
CIDEr~\cite{vedantam2015cider}
for the image captioning task. For ASR, we use word error rate (WER) for languages with Latin script and character error rate (CER) for those with non-Latin script.
        





\section{Result \& Analysis}
\label{sec:result}

\begin{figure}[!t]
  \includegraphics[width=\linewidth,trim={0, 0.9em, 0, 0}, clip]{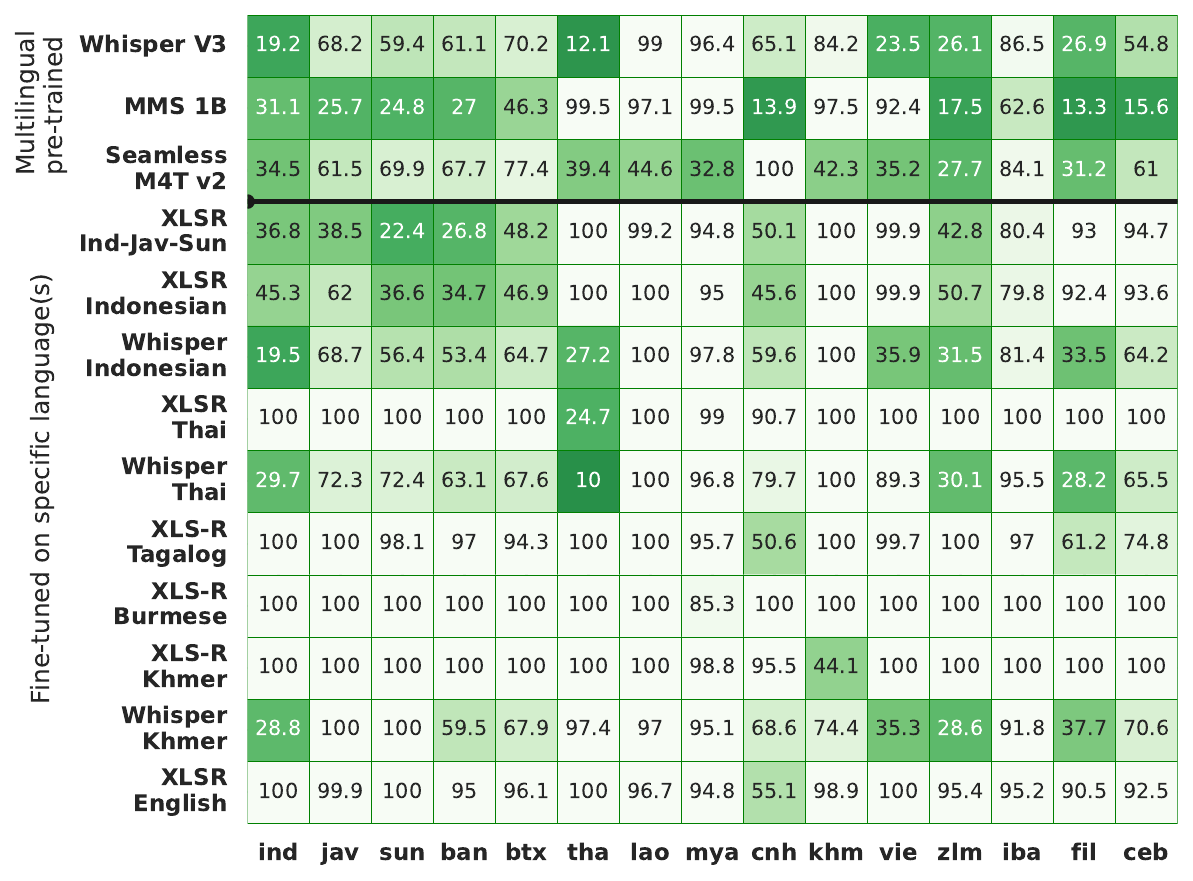}
  \caption{Speech model error rate (\%$\downarrow$) across existing ASR tasks in SEA languages.}
  \label{fig:sea-eval-speech-results}
    \vspace{-6pt}
\end{figure}

\subsection{State-of-the-Art Models on SEA languages}


\paragraph{LLMs}
Figure~\ref{fig:sea-nlu-overall-results} and \ref{fig:sea-nlg-overall-results} illustrate the overall model performance of the LLM baselines in SEA languages for both NLU tasks and NLG tasks. In our NLU evaluation, AYA-101, a large multilingual instruction-tuned language model covering 101 languages, demonstrates the best zero-shot performance. It is followed by the commercial baselines, which achieve a median of $\sim$0.6 weighted F1-score. Sailor and SeaLLM, models specifically trained with SEA languages, also display competitive performance. Similarly, mT0 exhibits strong generalization abilities due to its exposure to $\sim$100 languages in pre-training, including those from the SEA region~\cite{muennighoff2022crosslingual}. In contrast, most English and SEA country-specific baselines perform less effectively, likely due to their narrow focus on English or a limited set of SEA languages, such as Indonesian languages for Cendol and Thai for WangchanX-Llama3. Similar and consistent trends are observed on MT task, while the baselines' poorer scores on abstractive/extractive QA and summarization indicate their ineffectiveness in producing acceptable outputs in SEA languages for these tasks, which is especially pronounced in the open-source baselines. Appendix~\ref{app:sea-eval-results} describes the performance of LLMs per language.

To analyze the equality in model performance across SEA languages, following~\citet{khanuja-etal-2023-evaluating}, we utilize the Gini coefficient---originally used to observe income equality~\cite{dorfman1979formula}---weighted by demand and parameterized by $\tau$. Here, $\tau=1$ corresponds to a demographic notion of demand, considering language population size, while $\tau=0$ does not take population size into account~\cite{blasi-etal-2022-systematic}. Table~\ref{tab:lang-equity} shows that models trained on more SEA languages, such as multilingual and SEA regional baselines, generally exhibit greater language equity. For instance, although Command-R and GPT-4 are competitive performance-wise against AYA-101 and mT0, AYA-101 and mT0 demonstrate higher equality across all SEA languages under study. This trend is consistent across different $\tau$ (see Appendix~\ref{app:sea-eval-lang-equity}).

\begin{figure}[!t]
  \includegraphics[width=0.9\linewidth,trim={0, 0.9em, 0, 0}, clip]{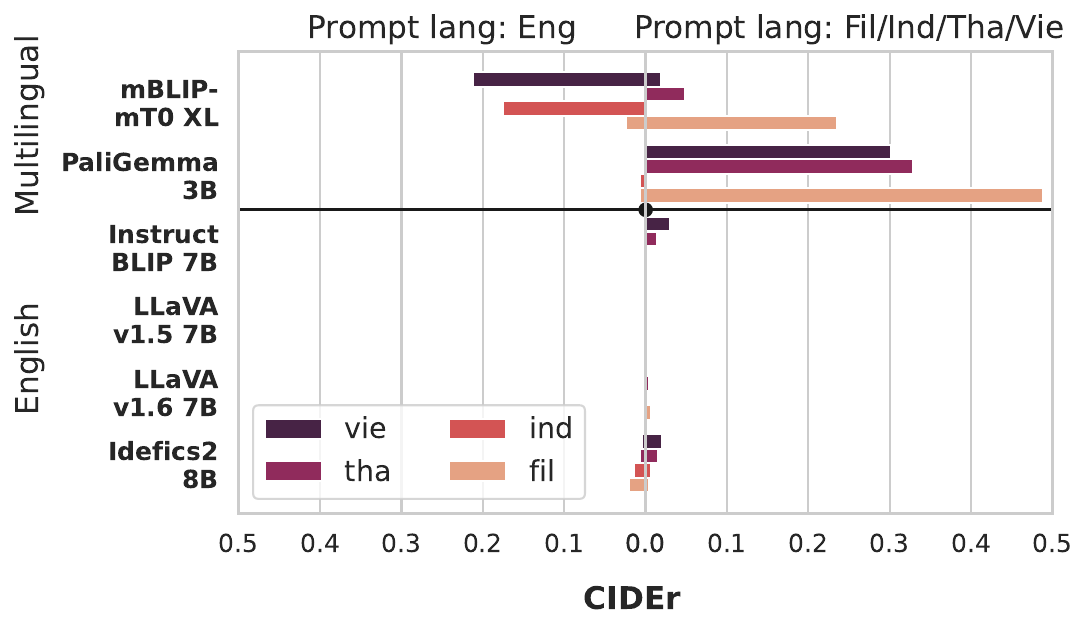}
  \caption{Existing VLMs produce subpar image captions in SEA languages. We report CIDEr~\cite{vedantam2015cider}.}
  \label{fig:sea-eval-vl-results}
  \vspace{-8pt}
\end{figure}

\paragraph{Speech models}

Figure~\ref{fig:sea-eval-speech-results} presents the off-the-shelf speech model performance on ASR across languages in SEA, measured by the error rate percentage.
9 of the 15 SEA languages in our speech evaluation belong to the Austronesian language family.
The other 6 are \textsc{khm} and \textsc{vie}, which belong to Austro-Asiatic, \textsc{cnh} and \textsc{mya} belong to Sino-Tibetan, and \textsc{tha} and \textsc{vie} belong to the Kra-Dai language family.
The multilingual pre-trained baselines have a competitive generalization capability across languages, although it varies by language.
For instance, Whisper v3 demonstrates significantly higher effectiveness for national languages such as \textsc{ind}, \textsc{zlm}, \textsc{fil}, \textsc{tha}, and \textsc{vie}, while performing less optimally for other indigenous languages.
Conversely, Seamless M4T v2 shows a more balanced performance across the languages.
Regarding fine-tuned baselines, error rates decrease for their seen languages.
The fine-tuned Whisper models, however, manage to better optimize for the target language while retaining their original capabilities in other SEA languages compared to their Wav2Vec2 XLSR and XLS-R counterparts, despite both having been pre-trained in a multilingual manner.
This observation aligns with the findings of \citet{rouditchenko23_interspeech}, who find that the number of hours seen per language and language family during pre-training is predictive of how the models compare, in which Whisper's pre-training data duration for these four language families exceeds that of XLSR.

\paragraph{VLMs}

Figure~\ref{fig:sea-eval-vl-results} depicts the zero-shot performance of off-the-shelf VLMs on image captioning in SEA indigenous languages. Despite the capability of LLMs for zero-shot cross-lingual generalization~\cite{huang-etal-2021-multilingual, tackstrom-etal-2012-cross, neubig-hu-2018-rapid, artetxe-etal-2020-cross}, VLMs trained only in English (i.e., InstructBLIP, LLaVA, and Idefics2) fail to exhibit this capability, struggling to generate adequate image captions in SEA languages. Multilingual VL pre-training is crucial to achieving aligned multilingual representations~\cite{burns2020learning, li-etal-2023-unifying, huang-etal-2021-multilingual}. For instance, PaliGemma and mBLIP generate better image captions in \textsc{tha} and \textsc{fil} when prompted in the relevant SEA languages.

\begin{table}[t]
    \centering
    \begin{subtable}[b]{0.45\linewidth}
        \centering
        \resizebox{\linewidth}{!}{
        \begin{tabular}{l p{1.3cm}}
          \toprule
          \textbf{Model} & \textbf{Natural outputs} \\
          \toprule
            \textbf{SEA-LION} & \textbf{58.57\%} \\
            AYA-23 & 43.57\% \\
            Sailor & 37.86\% \\
            Cendol-Llama2 & 37.37\% \\
            Malaysian Llama3 & 36.90\% \\
            WangchanX-Llama3 & 30.24\% \\
            Falcon & 29.52\% \\
            BactrianX-Llama & 28.10\% \\
            SeaLLM & 27.38\% \\
            Merak & 26.19\% \\
            BLOOMZ & 25.00\% \\
            Cendol-MT5 & 24.05\% \\
            Command-R & 20.95\% \\
            mT0-XL & 19.76\% \\
            Mistral & 19.52\% \\
            GPT-4 & 16.67\% \\
            Llama3 & 14.05\% \\
            AYA-101 & 8.33\% \\
          \bottomrule
        \end{tabular}
        }
        \caption{Avg. by models}
        \label{tab:natural-generative-results-by-models}
    \end{subtable}
    \hfill
    \begin{subtable}[b]{0.43\linewidth}
    \centering
        \resizebox{\linewidth}{!}{
        \begin{tabular}{p{2.8cm} p{1.3cm}}
          \toprule
          \textbf{Language} & \textbf{Natural outputs} \\
          \toprule
            \textbf{Indonesian (\textsc{ind})} & \textbf{41.58\%} \\
            Vietnamese (\textsc{vie}) & 37.31\% \\
            Thai (\textsc{tha}) & 34.21\% \\
            Khmer (\textsc{khm}) & 29.21\% \\
            Lao (\textsc{lao}) & 28.42\% \\
            Malay (\textsc{zlm}) & 22.24\% \\
            Burmese (\textsc{mya}) & 19.47\% \\
            Filipino (\textsc{fil}) & 12.22\% \\
            English (\textsc{eng})$^\dagger$ & 8.95\% \\
          \bottomrule
        \end{tabular}
        }
        \caption{Avg. by languages}
        \label{tab:natural-generative-results-by-langs}
    \end{subtable}
    \caption{Current LLMs are still incapable of generating natural texts in SEA languages. $^\dagger$As spoken in SEA regions, not worldwide.}
    \label{tab:natural-generative-results}
    \vspace{-6pt}
\end{table}

\begin{figure*}[!t]
  \centering
  \begin{subfigure}[t]{0.31\linewidth}
      \includegraphics[trim={0, 0, 0, 0}, clip, width=\linewidth]{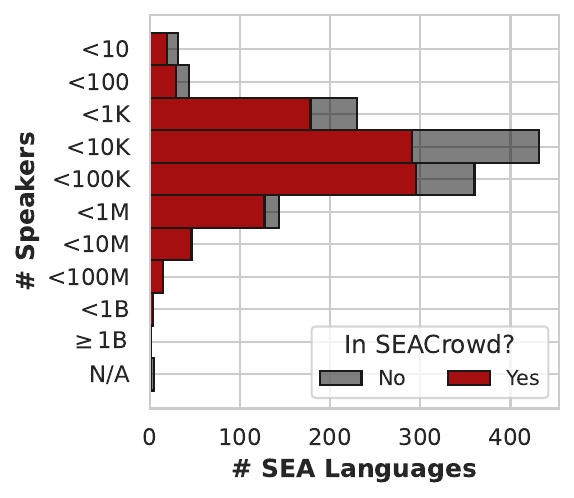}
      \caption{Language coverage}
      \label{fig:resource-gaps-coverage}
  \end{subfigure}
  \begin{subfigure}[t]{0.34\linewidth}
      \includegraphics[trim={0, 0, 0, 0}, clip, width=\linewidth]{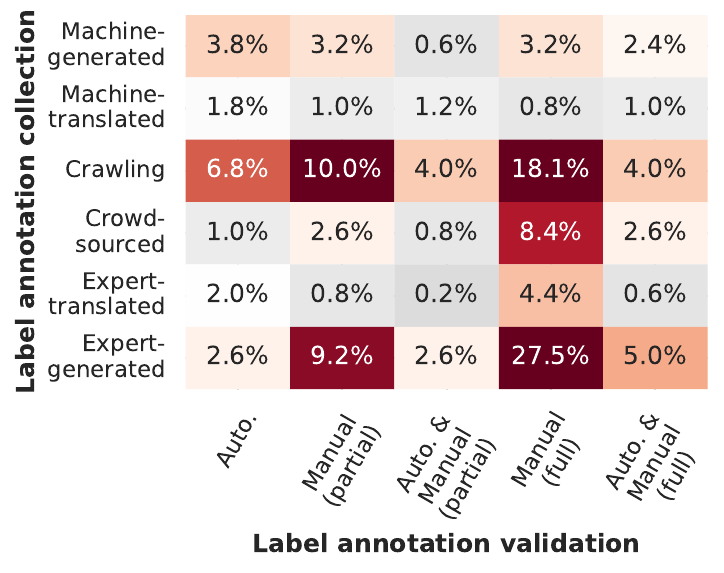}
      \caption{Annotation quality}
      \label{fig:resource-gaps-quality}
  \end{subfigure}
  \begin{subfigure}[t]{0.335\linewidth}
      \includegraphics[trim={0, 0, 0, 0}, clip, width=\linewidth]{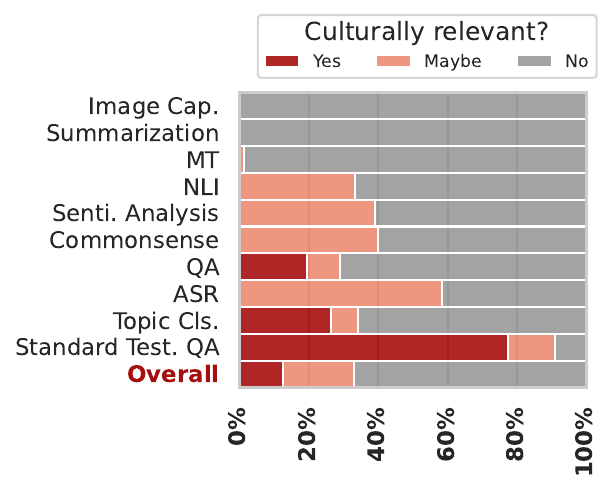}
      \caption{Cultural relevance}
      \label{fig:resource-gaps-culture}
  \end{subfigure}
  \caption{The resource gap in SEA in terms of language coverage, annotation quality, and cultural relevance.}
  \label{fig:resource-gaps}
  \vspace{-8pt}
\end{figure*}

However, when prompted in \textsc{eng}, the performance of these multilingual baselines varies notably. PaliGemma's performance collapses completely, while mBLIP's performance shows both increases and decreases across different SEA languages. This raises the question of whether the multilingual VLMs can maintain consistent performance across different languages used in the instructions and the tasks. It highlights the need for further research into the mechanisms that drive these variations and how to achieve robust multilingual performance in VLMs across diverse linguistic contexts. Understanding these dynamics is crucial for improving VLMs' generalization capabilities and ensuring equitable performance across all languages, despite most related works focusing on monolingual visual instruction tuning~\cite{liu2023llava, gong2023multimodal, zhu2023minigpt}.

\subsection{Generation Quality in SEA Languages: Translationese vs. Natural Language}
\label{sec:result-naturalness}

\paragraph{Classifying Translationese in SEA Languages}

To analyze the generation quality of LLMs in SEA languages, we build a text classifier to discriminate between translationese and natural texts~\cite{riley-etal-2020-translationese}. We construct a translationese classification training and testing dataset using 49 and 62 data subsets, respectively, covering approximately 39.9k and 51.5k sentences across English (\textsc{eng}) and 8 SEA languages: Indonesian (\textsc{ind}), Khmer (\textsc{khm}), Lao (\textsc{lao}), Burmese (\textsc{mya}), Filipino (\textsc{fil}), Thai (\textsc{tha}), Vietnamese (\textsc{vie}), and Malay (\textsc{zlm}). The training and test data are detailed in Appendix~\ref{app:translationese-natural-classifier-data}.

We fine-tune a classifier from mDeBERTaV3~\cite{he2020deberta,he2022debertav3}\footnote{\url{https://huggingface.co/microsoft/mdeberta-v3-base}} using these data and achieve 79.08\% accuracy on the test set in predicting translationese across these 9 languages. The detailed results and ablation studies of our translationese classifier experiments are provided in Appendix~\ref{app:translationese-natural-classifier-exp}. This classifier enables us to assess the generation quality of LLMs by distinguishing between translationese and naturally occurring text, providing insights into the models' performance in producing authentic language output.


\paragraph{Generation Quality of LLMs}

We evaluate the generation quality of LLMs in 9 SEA languages by generating answers to natural, general, and safety questions from Sea-Bench~\cite{damonlpsg2023seallm}. As shown in Table~\ref{tab:natural-generative-results-by-models}, LLMs with extensive language coverage but less focus on SEA languages, e.g., AYA-101~\cite{ustun2024aya}, GPT-4~\cite{openai2024gpt4}, mT0~\cite{muennighoff-etal-2023-crosslingual,xue-etal-2021-mt5}, and Llama3~\cite{llama3modelcard}, tend to produce natural sentences less than 20\% of the time. In contrast, models with narrower language coverage but a greater focus on SEA languages, such as Cendol-Llama2~\cite{cahyawijaya2024cendol}, Sailor~\cite{dou2024sailor}, AYA-23~\cite{aryabumi2024aya}, and SEA-LION~\cite{sea_lion_2023}, generate natural sentences over 35\% of the time.

However, even the LLM with the least translationese generation, SEA-LION, only produces natural SEA sentences 57.71\% of the time, highlighting a significant quality gap in generating natural sentences in SEA languages. As displayed in Table~\ref{tab:natural-generative-results-by-langs}, the translationese issue varies across SEA languages. Languages such as Tagalog (\textsc{tgl}), Burmese (\textsc{mya}), and Malay (\textsc{zlm}) have more severe translationese problems, with existing LLMs producing natural sentences only 11.58\%, 19.47\%, and 22.24\% of the time, respectively. This underscores the need for further improvements in LLMs to more effectively address the linguistic diversity and complexity of SEA languages.

\section{Discussions}
\label{sec:discussions}

\subsection{Resource Gaps in SEA}
\label{sec:resource-gaps}

\paragraph{Coverage} 

SEACrowd covers 980 out of the 1,308 languages spoken in SEA (74.9\%). Despite this high coverage, language representation in SEACrowd exhibits a very long-tail distribution, with over 700 languages having only 1 or 2 datasets, and only 23 languages having 20 datasets or more. These less represented languages typically exist only in the form of lexicons~\cite{asgari-etal-2020-unisent,List2022} or unlabeled data~\cite{leong-etal-2022-bloom,kudugunta2023,nguyen-etal-2024-culturax-cleaned}. Existing tasks in SEACrowd still cover only a small portion of languages. For instance, sentiment analysis data is available for only 22 languages, and named entity recognition (NER) data is available for just 17 languages. Furthermore, for modalities beyond text, SEA resources are extremely underrepresented. Approximately 90\% of SEA indigenous languages lack both speech and VL datasets.

\begin{figure*}[t]
  \centering
  \includegraphics[trim={0.4cm, 1.1cm, 0.4cm, 0.3cm}, clip, width=0.875\linewidth]{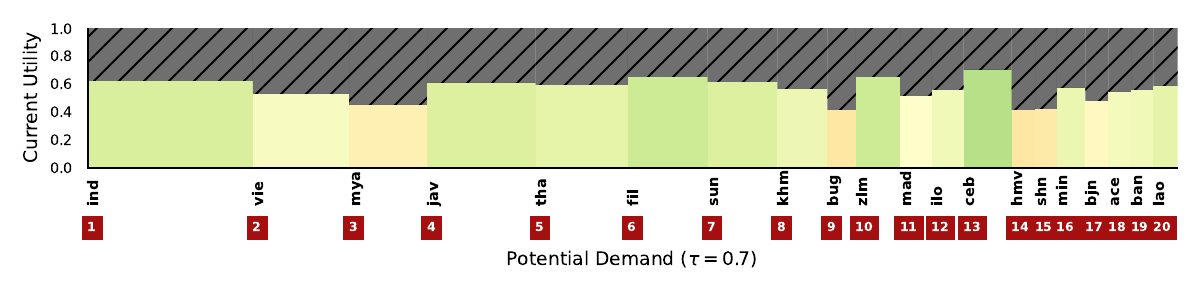}
  \includegraphics[trim={0.4cm, 0.55cm, 0.4cm, 0.3cm}, clip, width=0.875\linewidth]{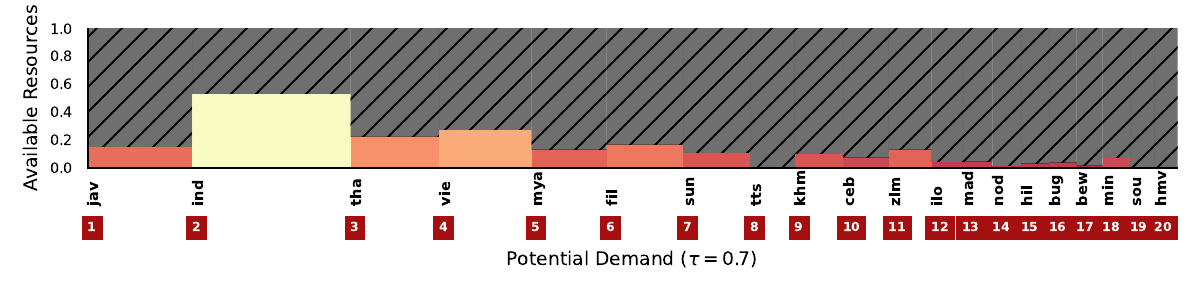}
  \caption{SEA languages prioritization
  based on \textbf{(top)} current utility and \textbf{(bottom)} resource availability. The languages are \colorbox[RGB]{166, 15, 15}{\textcolor{white}{ranked}} based on the descending order of the area size of their missing potential \colorbox{gray}{\legendsquare{pattern=north east lines}}.}
  \label{fig:next-step}
  \vspace{-6pt}
\end{figure*}

\paragraph{Quality} 


78.7\% of the datasets in SEACrowd are published in peer-reviewed venues, and most of the data has undergone external validation. The overall quality of the datasets in SEACrowd is depicted in Figure~\ref{fig:resource-gaps-quality}. We compile the reported data construction methods by the authors, considering both the data collection method (i.e., data source) and label annotation validation (i.e., quality control). Nearly 19\% of the datasets in SEACrowd have machine-generated and machine-translated annotations, while
more than 80\% were obtained from online texts (e.g., web crawling) and expert generation. In terms of label annotation validation, 62.4\% of the datasets have been fully manually checked, while the remaining portion is partially validated and automatically checked. Note that these statistics only provide an initial indication of dataset collection quality on the surface and do not necessarily reflect the exact quality. Only a few datasets (6\%) in SEACrowd report their detailed quality metrics (e.g., inter-annotator agreement scores). A deeper investigation is required for future work.


\paragraph{Cultural Relevance} 

The resource gap in SEA extends to the cultural aspect, where misrepresentation can lead to offensive behaviors, e.g., cultural appropriation and stereotyping~\cite{evans2020good,glotov2023intercultural}. 
As a proxy of the cultural relevance of SEA datasets, we manually curated 259 data subsets used in SEACrowd evaluation based on their data source. Specifically, we categorize them whether they are 1) translated from another language, 2) crawled from local sources, or 3) hand-crafted to capture cultural relevance. In Figure~\ref{fig:resource-gaps-culture}, approximately 70\% lack cultural relevance, as many are machine-translated from English sources. About 20\% are taken from local news, social media, or other local outlets, which potentially contain some culturally relevant data. Only the remaining 10\% are designed to consider cultural relevance, derived from studies highlighting serious deficiencies in cultural understanding by LLMs for underrepresented languages~\cite{kabra-etal-2023-multi,koto-etal-2023-indommlu,wibowo2023copal,liu2024multilingual,koto2024indoculture}.

\subsection{Conclusion \& Future Work}



Southeast Asia is home to highly diverse languages and cultures; the majority of its people do not use English as their primary language.
The utility of English-first AI is limited for the majority of Southeast Asian users, especially in critical sectors like healthcare and education.
Through SEACrowd, we have explored the AI landscape in SEA and bridged the gaps in resources, evaluation, and naturalness analysis of AI models in SEA languages.
Further, our initiative has nurtured an open-source research community, which will actively continue to add and maintain datasheets and dataloaders, as well as drive AI research and developments in SEA.

Nonetheless, AI development in SEA requires concentrated efforts by a range of stakeholders, who may prioritize differently when it comes to incorporating the region's 1,300+ languages into AI models.
Moving forward, our work suggests AI development in SEA should prioritize two key metrics: 1) potential utility and 2) resource equity.\footnote{\url{https://github.com/SEACrowd/globalutility}}

\paragraph{Potential utility} Potential utility is defined as the gap between current utility and ideal utility, in which model capability acts as a proxy for utility.
Based on potential utility, unsurprisingly the development of the national languages (except for English and Chinese used in Singapore), i.e., Indonesian (\textsc{ind}), Burmese (\textsc{mya}), Vietnamese (\textsc{vie}), Thai (\textsc{tha}), Filipino (\textsc{fil}), Khmer (\textsc{khm}), Malay (\textsc{zlm}), and Lao (\textsc{lao}) in Figure~\ref{fig:next-step}, will bring the biggest benefit. 
Among them, we identify notable gaps in the naturalness of Malay, Burmese, and Filipino AI-generated outputs (\S\ref{sec:result-naturalness}).
Focused efforts in resource building for these languages may move the needle the most for utility.
Beyond the national languages, growing local languages or dialects with large speaker bases, e.g., Javanese (\textsc{jav}), Sundanese (\textsc{sun}), and Hmong (\textsc{hmn}), is key.

\paragraph{Resource equity} Resource equity is defined as the gap between existing and ideal resource availability (Figure~\ref{fig:next-step}). We found that many local languages or dialects still fall short of the expected level of resources. 
These include Northeastern Thai (\textsc{tts}), Northern Thai (\textsc{nod}), Hmong Do (\textsc{hmv}), Southern Thai (\textsc{sou}), Cebuano (\textsc{ceb}), Ilocano (\textsc{ilo}), and others.
Efforts to narrow these gaps would not only help preserve these languages but also ensure the continuation of the cultural heritage of the speakers of these languages. More details on SEA language prioritization for different weightings of demand can be found in Appendix~\ref{app:lang-prioritization}.



To improve these metrics, governments, and industry leaders in the region should invest in R\&D activities to improve regional language capability for both the national languages and local dialects. 
This could include funding for open data collection and collaborations with local communities to address the resource gap in local languages. This also requires long-term sustainable strategies, such as catalyzing profitable use cases based on inclusive AI models, promoting fair and responsible compensation schemes for data workers, and orchestrating win-win exemplar collaborations between data owners, AI, and application developers.



\section*{Acknowledgments}
\label{sec:acknowledgments}

We would like to thank our amazing contributors: Joshua Spergel, Tiezheng Yu, Parinthapat Pengpun, Ishan Jindal, Muhammad Satrio, Jipeng Zhang, Bhavish Pahwa, Haryo Akbarianto Wibowo, Hiroki Nomoto, Yohanes Sigit Purnomo W.P., Ahmad Fathan Hidayatullah, Bryan Wilie, Ruhiyah Faradishi Widiaputri, Rafif Rabbani, Fawwaz Mayda, Manoj Khatri, Supryadi Supryadi, Virach Sornlertlamvanich, Pavaris Ruangchutiphophan, Erland Hilman Fuadi, Mega Fransiska, Richardy Sapan, and Camilla Johnine Cosme, for their hard work in submitting datasheets and implementing dataloaders for SEACrowd. 

This work is supported by the National Research Foundation, Singapore under its AI Singapore Programme; PhD Fellowship Award, the Hong Kong University of Science and Technology; and PF20-43679 Hong Kong PhD Fellowship Scheme, Research Grant Council, Hong Kong. JMI is funded by National University Philippines and the UKRI Centre for Doctoral
Training in Accountable, Responsible and Transparent AI [EP/S023437/1] of the University of Bath. In addition, we would like to express our gratitude to Cohere For AI for providing research grants that enabled us to perform our experiments using a commercial baseline, specifically Command-R.


\section*{Limitations}
\label{sec:limitations}

While our work covers nearly 1,000 SEA languages, many dialects, which are considered as belonging to a parent language, are missing from our evaluation benchmark. For instance, for the Malay language, only Standard Malay (\textsc{zsm}) is evaluated, but not other dialects such as Sarawak Malay (\textsc{zlm-sar}). Furthermore, the majority of our datasets also do not contain code-switched texts, which is a common linguistic phenomenon of SEA language usage \citep{aji-etal-2023-current}. Moreover, the language coverage of different evaluation tasks varies significantly. For instance, NLP tasks cover 34 languages in total, whereas VL tasks only cover 4 languages.

\section*{Ethics Statement}
\label{sec:ethics-statement}
In developing an evaluation benchmark for SEA languages, we have taken several steps to ensure ethical considerations are addressed comprehensively. First, the data used for this benchmark is sourced from publicly available resources, ensuring compliance with legal and ethical standards regarding data privacy. Where applicable, explicit consent was obtained from data contributors. Furthermore, all the datasets and resources utilized in this benchmark are used in accordance with their respective licenses. Second, our benchmark aims to be inclusive, representing a wide range of SEA languages, including those that are underrepresented in current linguistic resources. Lastly, our research process, including data collection, benchmark development, and evaluation methodologies, is entirely open-sourced and is documented transparently to enable reproducibility and accountability.

\newpage
\bibliography{custom,anthology}

\newpage

\appendix

\renewcommand{\UrlFont}{\ttfamily\small}

\section{Key Takeaways of SEACrowd}
\label{app:conclusion}


\begin{table*}[t]
    \centering
    \resizebox{\linewidth}{!}{
    \begin{tabular}{lccccc}
      \toprule
      \textbf{Benchmark} & 
      \textbf{\# Languages} & 
      \textbf{\# Indigenous SEA Languages} & \textbf{\# Datasets} & \textbf{\# Tasks} \\ \midrule
      SEACrowd (\textbf{ours})$^\dagger$ & 39 & 38 & 254 & \multicolumn{1}{l}{13 (11 text, 1 speech, 1 vision)} \\
      NusaCrowd$^\dagger$~\cite{cahyawijaya-etal-2023-nusacrowd} & 19 & 19 & 137 & \multicolumn{1}{l}{12 (11 text, 1 speech)} \\
      BUFFET~\cite{asai2023buffet} & 54 & N/A & 15 & \multicolumn{1}{l}{8 (8 text)}\\
      XTREME-UP~\cite{ruder2023xtreme} & 88 & 11 & 269 & \multicolumn{1}{l}{9 (7 text, 1 speech, 1 vision)}\\
      \bottomrule
    \end{tabular}
    }
    \caption{Benchmark comparison. $^\dagger$The numbers in SEACrowd and NusaCrowd are the numbers of datasets included in the evaluation.}
    \label{tab:benchmark_comparison}
\end{table*}

Key findings include:

\paragraph{Model Performance.}
\begin{compactitem}
    \item \textbf{LLMs:} SEA-specific models, such as AYA-101 and mT0, show strong performance on zero-shot tasks, outperforming English or country-specific models in the region. However, tasks like abstractive QA and summarization reveal limitations in existing models' ability to handle SEA languages effectively.
    \item \textbf{Speech:} Off-the-shelf models like Whisper v3 show competitive ASR performance for major SEA languages but struggle with indigenous languages. In contrast, Seamless M4T v2 offers more balanced results across SEA languages.
    \item \textbf{VLMs:} Current VLMs fail to generate high-quality image captions in SEA languages, highlighting the need for more effective multilingual pre-training.
\end{compactitem}

\paragraph{LLM Generation Quality.} SEA language outputs by LLMs are often plagued by translationese, with models like SEA-LION v1 producing natural sentences only 57.71\% of the time. Languages like Tagalog, Burmese, and Malay suffer from unnatural generation.

\paragraph{Resource Gaps.} SEACrowd covers 74.9\% of SEA languages but reveals a long-tail distribution, where most languages lack comprehensive datasets. SEA languages also face cultural misrepresentation, with 70\% of datasets being translations rather than culturally relevant sources.

\paragraph{Prioritizing Development.} Focus should be placed on SEA national languages with significant gaps in naturalness (e.g., Malay, Burmese, Filipino), as well as under-resourced local languages like Javanese and Cebuano.

\paragraph{Collaboration.} Governments, industries, and local communities must invest in R\&D, data collection, and open collaboration to address resource equity and improve SEA AI development.

\section{Related Work}
\label{app:related-work}

\paragraph{SEA data resources} 
LLM research efforts for SEA languages are limited by the lack of available datasets and benchmarks. Up to this day, resources for SEA NLP tasks are concentrated on relatively higher-resource SEA indigenous languages, such as Indonesian \citep{mahendra-etal-2021-indonli,wilie-etal-2020-indonlu,cahyawijaya-etal-2021-indonlg,cahyawijaya-etal-2023-nusacrowd} and Vietnamese \citep{nguyen-etal-2020-vietnamese,huynh-etal-2022-vinli,le-luu-2023-parallel,van2022new}. NusaCrowd~\cite{cahyawijaya-etal-2023-nusacrowd} introduce the first multimodal benchmark for Indonesian languages, including text and speech.~\citet{ruder2023xtreme} introduce a multimodal benchmark encompassing 11 indigenous languages from SEA, spanning a wide array of languages totaling 88.

Additionally, \citet{asai2023buffet} present an LLM benchmark for cross-lingual few-shot transfer, comprising 15 distinct tasks and 54 languages sourced from varied multilingual datasets. Furthermore, \citet{dou2024sailor} find that publicly available pre-training data for SEA languages suffer from quality issues such as textual duplicates and excessive occurrences of Unicode escapes. On the other hand, pre-trained LLMs specifically for SEA languages suffer from limited language coverage; for instance, Cendol~\cite{cahyawijaya2024cendol}, Sailor~\cite{dou2024sailor}, SEA-LION~\citep{sea_lion_2023}, and SeaLLMs~\cite{damonlpsg2023seallm} have only covered up to 11 different SEA languages, including English and Chinese.

\paragraph{Open-source Community Initiatives in NLP}
Open-source and open-science communities play a crucial role in engaging native speakers to curate large-scale multilingual NLP resources. In the past, collaborative efforts have been organized to collect data and train multilingual language models either on a global scale \citep{workshop2022bloom,singh2024aya,ustun2024aya} or on a regional level, e.g., Masakhane for African languages \citep{adelani-etal-2021-masakhaner,adelani-etal-2022-masakhaner,adelani-etal-2022-thousand,adelani-etal-2023-masakhanews}, AI4Bharat for Indian languages \citep[inter alia]{kakwani-etal-2020-indicnlpsuite,kumar-etal-2022-indicnlg,dabre-etal-2022-indicbart}, and AmericasNLP for Latin American languages~\cite{americasnlp-2021-natural,ebrahimi-etal-2022-americasnli}.

In the SEA region, there have been community-based initiatives, e.g., IndoNLP, PyThaiNLP, and RojakNLP, to study NLP on Indonesian languages \citep{aji-etal-2022-one,wilie-etal-2020-indonlu,cahyawijaya-etal-2021-indonlg,cahyawijaya-etal-2023-nusacrowd}, Thai language \citep{phatthiyaphaibun-etal-2023-pythainlp},
and the code-switching phenomenon in SEA \citep{aji-etal-2023-current,yong-etal-2023-prompting,winata2024miners}, respectively.

\begin{table}[t]
    \centering
    \resizebox{\linewidth}{!}{
    \begin{tabular}{lll}
    \toprule
    \textbf{Submission} & \textbf{Points} & \textbf{Max points} \\
    \toprule
    Public datasheet & 2+bonus & 6 \\
    Dataloader & 3 & 6 if difficult \\
    Private datasheet & 1 & - \\
    Access to private data & 4+bonus & 10 if high-quality \\
    \midrule
    Datasheet review & 1 & 1 \\
    Dataloader review & 2 & 4 if difficult \\
    Private datasheet review & 0.5 & - \\
    Private data contact & 1 & 5 if succeeds \\
    \bottomrule
    \end{tabular}
    }
    \caption{Amount of points obtained for contributions related to datasheet, dataloader, and private data.}
    \label{tab:contribution-points}
\end{table}

\section{Contributing to SEACrowd}
\label{app:contributing-to-seacrowd}

\subsection{Open Contributions}
\label{app:seacrowd-open-contributions}

We identify four tasks for open contribution in SEACrowd.\footnote{Landing page: \url{https://github.com/SEACrowd}.} These tasks and the workflow of SEACrowd are heavily influenced by and extended upon NusaCrowd~\cite{cahyawijaya-etal-2023-nusacrowd, cahyawijaya2022nusacrowd_procedure}, a collaborative effort to pool data resources for Indonesian NLP.

\begin{figure*}[t]
\centering
  \includegraphics[width=0.7\linewidth]{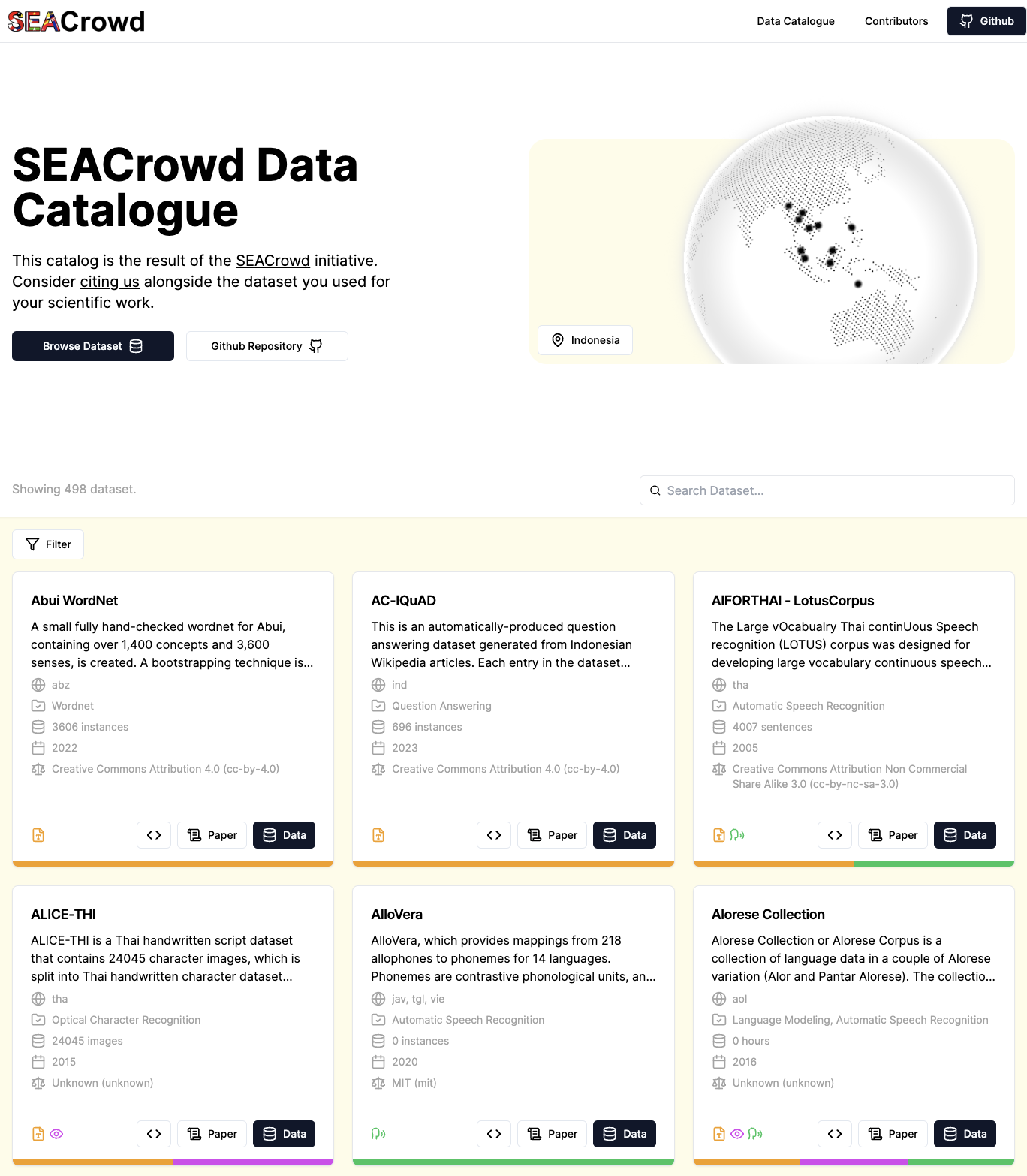}
  \caption{A glimpse of SEACrowd Catalogue.}
  \label{fig:seacrowd-catalogue}
\end{figure*}

\begin{compactitem}

\item \textbf{Submitting Metadata for Existing Public Datasets.}
Contributors can submit detailed datasheets for existing datasets through this form.\footnote{Public datasheet form: \url{https://form.jotform.com/team/232952680898069/seacrowd-sea-datasets}.} Contributors must provide important information such as data license, size, language and dialect, annotation method, and so on. The approved datasheets, as well as under review datasheets, will show up and be indexed in a monitor spreadsheet and the SEACrowd Catalogue (Figure~\ref{fig:seacrowd-catalogue}).

\item \textbf{Building a Dataloader.}
From the approved datasheets from the previous task, contributors can further contribute by building a HuggingFace dataset loader to ensure that all datasets in SEACrowd are standardized in terms of formatting and usage. Contributors can follow a dataloader guide and examples available\footnote{Dataloader guide: \url{https://github.com/SEACrowd/seacrowd-datahub/blob/master/DATALOADER.md}.} in the SEACrowd Data Hub. Dataloader maintainers and reviewers also monitor the self-assigned dataloader issues after 2 weeks of inactivity and ping contributors in case of a blocking impediment.

\item \textbf{Identifying Private AI Datasets for SEA Languages, Cultures, and/or Regions.}
Unfortunately, a number of prior works involving SEA languages are still not publicly available. These may be due to several different reasons, including (but not limited to): non-release contracts related to funding, inclusion of private and personally identifiable data, and the use of explicitly private data such as those used by for-profit companies.

In this task, contributors can search for works that contain private data and fill out a corresponding record form.\footnote{Papers with private dataset form: \url{https://form.jotform.com/team/232952680898069/seacrowd-paper-with-private-dataset}.} The SEACrowd team then attempts to contact the original data owners and negotiate the open-sourcing of their resources.


\item \textbf{Opening a Private AI Dataset of SEA.}
If a contributor has previous work with closed data (or has been contacted by the SEACrowd team regarding closed-source data), they can decide to release their resources and register them in the collection via the public datasheet form. The resource will still be owned by the original contributor and is still tied to the contributor's previous work, as SEACrowd simply catalogs it and records its now open-source license.

\end{compactitem}

\subsection{Measuring Contributions}
\label{app:seacrowd-contribution-points}

\begin{figure*}[t]
\centering
  \includegraphics[width=\linewidth]{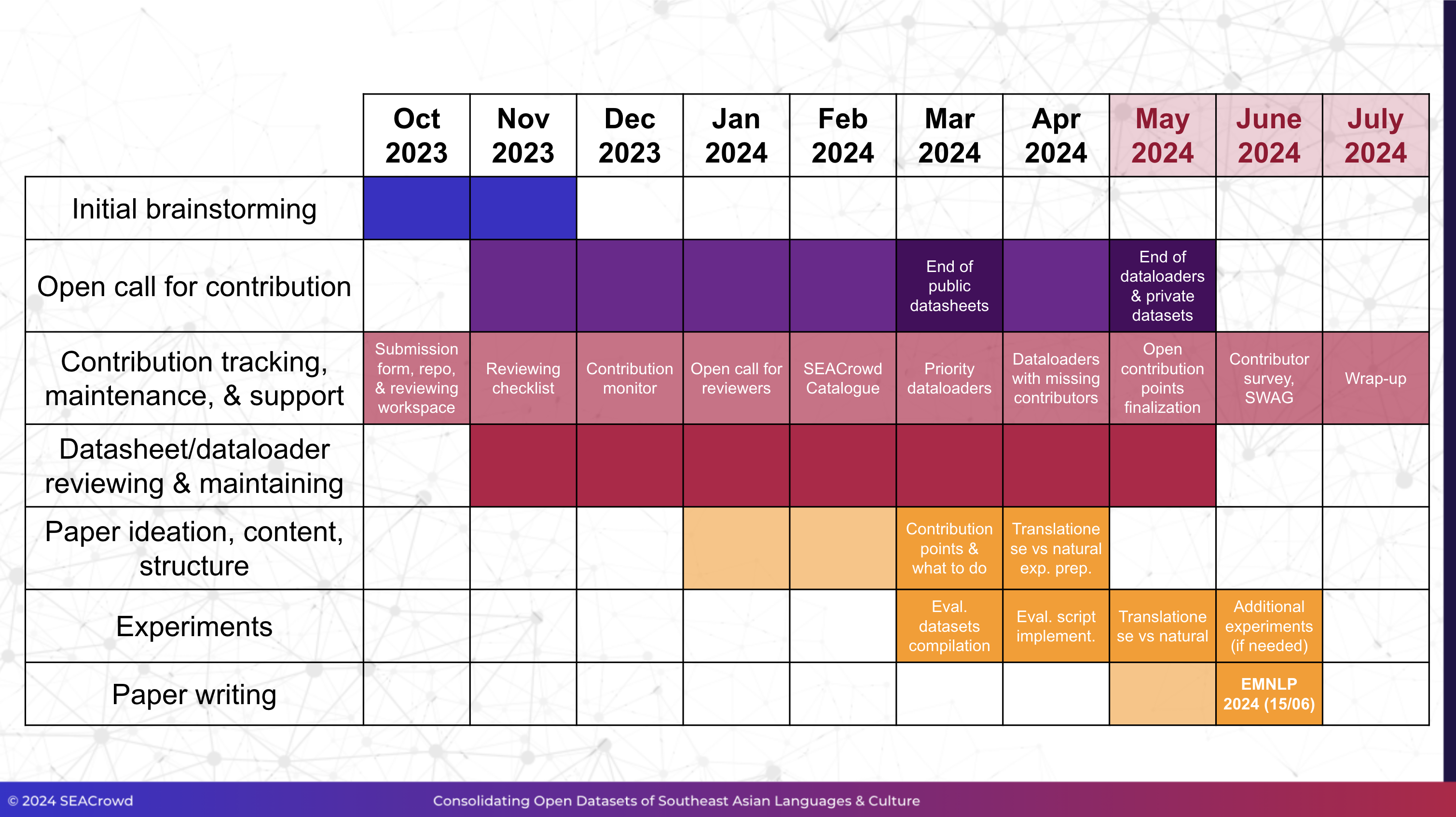}
  \caption{The timeline of SEACrowd's entire run.}
  \label{fig:seacrowd-timeline}
\end{figure*}

To be considered as a co-author, 20 contribution points are required.\footnote{Submissions past the deadlines (see Appendix~\ref{app:seacrowd-timeline}) are still recorded, but contribution points are no longer given.} To monitor how many points the contributors have obtained, \href{https://docs.google.com/spreadsheets/d/e/2PACX-1vQDZtJjA6i7JsxS5IlMtVuwOYjr2Pbl_b47yMSH4aAdHDBIpf-CiJQjNQAzcJPEu_aE7kwH4ZvKvPm0/pubhtml?gid=225616890&single=true}{the contribution point tracking} is provided and updated regularly. The purpose of the point system is not to barrier collaboration but to reward rare and high-quality dataset entries. Table~\ref{tab:contribution-points} describes the contribution points.\footnote{Contribution point guidelines: \url{https://github.com/SEACrowd/seacrowd-datahub/blob/master/POINTS.md}.} A bonus of 1 point is given if the dataset modality is speech or vision. We also provide a bonus based on the language rarity in terms of available resources as defined by~\citet{joshi-etal-2020-state}\footnote{\url{https://microsoft.github.io/linguisticdiversity/assets/lang2tax.txt}}, consisting of 1 point for languages in level 1 and 2, and 2 points for languages in level 0 or absent from the list. For other contributions not mentioned in Table~\ref{tab:contribution-points} (e.g., maintenance, design, experiment, paper writing, etc.), the amount of contribution points is adjusted to the bulk and the complexity of the relevant work.

\section{Progression of SEACrowd}
\label{app:seacrowd-progression}

\subsection{Timeline}
\label{app:seacrowd-timeline}

SEACrowd released the open call for contributions on 1 November 2023. This lasted until 31 March 2024, for datasheet submissions, and until 15 May 2024 for both dataloaders and private dataset submissions. SEACrowd contributors have a biweekly discussion regarding the challenges they face while contributing, the next steps they should take to proceed, and/or experiment and research ideas for the paper. The detailed timeline can be seen in Figure~\ref{fig:seacrowd-timeline}.

\subsection{Contribution Progress}
\label{app:seacrowd-contribution-progress}

\begin{figure*}[t]
\centering
  \includegraphics[width=0.8\linewidth]{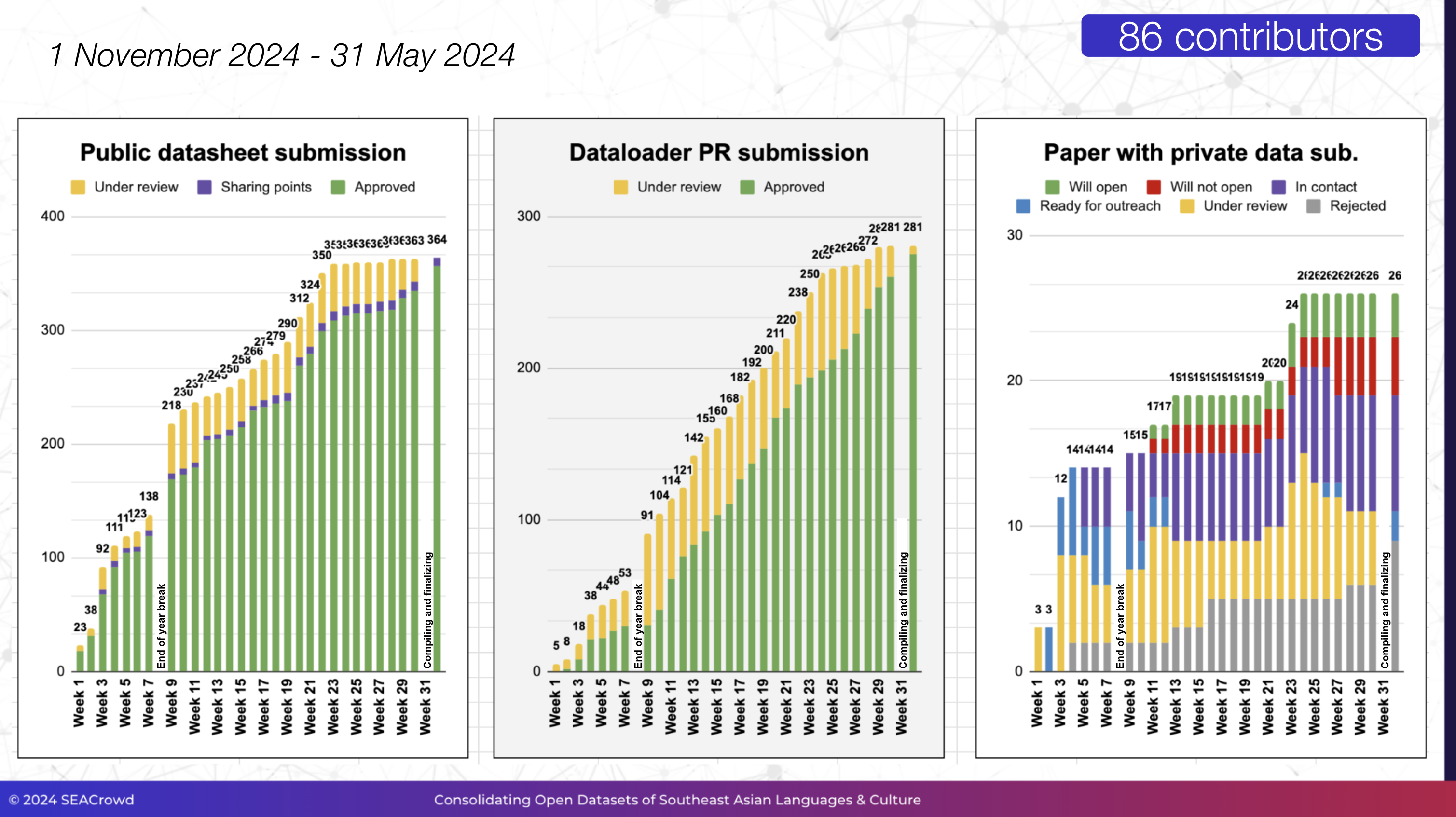}
  \caption{Weekly status update of the cumulative number of submissions in SEACrowd.}
  \label{fig:seacrowd-contribution-progress}
\end{figure*}

Figure~\ref{fig:seacrowd-contribution-progress} shows the number of submissions for public datasheets, dataloader pull requests, and papers with private datasets in SEACrowd.

\section{Reviewing SEACrowd's Submissions}
\label{app:reviewing-in-seacrowd}

We provide the complete reviewing guidelines in our Data Hub.\footnote{Reviewer SOP: \url{https://github.com/SEACrowd/seacrowd-datahub/blob/master/REVIEWING.md}}

\subsection{Datasheet Reviewing}

The datasheet reviewing standard operating procedure (SOP)
ensures the integrity and completeness of datasets submitted to SEACrowd. It outlines procedures for verifying dataset availability, avoiding duplicates, and ensuring correctness and relevance to the SEA region. The SOP includes FAQs addressing common issues such as dataset duplicates and incorrect information, along with an approval checklist covering aspects like data availability, dataset splits, and licensing. Reviewers are instructed on how to handle various scenarios, including correcting errors and determining points allocation for multiple contributors. For instance, if the datasheet submitted has incorrect or missing information, the reviewer can either ask the contributor to fix it (with some guidance) or fix it themself. Upon completion of the review, reviewers update the status, add notes and points, and await the generation of a GitHub issue for the approved datasheet.

\subsection{Dataloader Reviewing}

The dataloader reviewing SOP
governs the review process for dataloaders in SEACrowd, ensuring adherence to the data structure and \texttt{seacrowd} schema and config standards. It specifies checks for metadata correctness, subset implementation, test script passing, and adherence to coding conventions. Additionally, it outlines dataloader config rules based on dataset types and provides guidelines for multilingual datasets. The SOP emphasizes the importance of reviewer collaboration, with each dataloader requiring two reviewers per submitted pull request, and outlines the approval and reviewer assignment process, either by allocation or by self-assignment based on availability and promptness.

\section{Schemas in SEACrowd}
\label{app:task-schema}


Schemas define and format the attributes of the dataset returned by a dataloader. For each dataloader, we implement 2 schema types: the source schema and the \texttt{seacrowd} schema. The source schema presents the dataset in a format similar to its original structure, while the \texttt{seacrowd} schema standardizes the data structure across similar tasks.

The following subsections define the \texttt{seacrowd} schemas in NLP (\ref{app:schemas-nlp}), speech (\ref{app:schemas-speech}), and VL (\ref{app:schemas-vl}).

\begin{table}[t]
    \centering
    \begin{subtable}[t]{\linewidth}
        \centering
        \resizebox{\linewidth}{!}{
        \begin{tabular}{lclr}
          \toprule
          \textbf{Subset ID} & \textbf{Language} & \textbf{Region} & \textbf{\# Samples} \\
          \toprule
            \multicolumn{4}{c}{\cellcolor[HTML]{D9D9D9}\textit{\textbf{Sentiment Analysis} $\rightarrow$ \texttt{*\_seacrowd\_text}}} \\
            \texttt{lazada\_review\_filipino} & \textsc{fil} & Philippines & 1001 \\
            \texttt{gklmip\_sentiment} & \textsc{mya} & Myanmar & 716 \\
            \texttt{indolem\_sentiment} & \textsc{ind} & Indonesia & 1011 \\
            \texttt{id\_sentiment\_analysis} & \textsc{ind} & Indonesia & 10806 \\
            \texttt{karonese\_sentiment} & \textsc{btx} & Indonesia & 1000 \\
            \texttt{wisesight\_thai\_sentiment} & \textsc{tha} & Thailand & 2671 \\
            \texttt{wongnai\_reviews} & \textsc{tha} & Thailand & 6203 \\
            \texttt{typhoon\_yolanda\_tweets} & \textsc{fil} & Philippines & 153 \\
            \texttt{smsa} & \textsc{ind} & Indonesia & 500 \\
            \texttt{prdect\_id\_sentiment} & \textsc{ind} & Indonesia & 5400 \\
            \texttt{id\_sent\_emo\_mobile\_apps\_sentiment} & \textsc{ind} & Indonesia & 21696 \\
            \texttt{shopee\_reviews\_tagalog} & \textsc{fil} & Philippines & 2250 \\
            \texttt{nusatranslation\_senti\_abs} & \textsc{abs} & Indonesia & 500 \\
            \texttt{nusatranslation\_senti\_btk} & \textsc{btx} & Indonesia & 1200 \\
            \texttt{nusatranslation\_senti\_bew} & \textsc{bew} & Indonesia & 1200 \\
            \texttt{nusatranslation\_senti\_bhp} & \textsc{bhp} & Indonesia & 500 \\
            \texttt{nusatranslation\_senti\_jav} & \textsc{jav} & Indonesia & 1200 \\
            \texttt{nusatranslation\_senti\_mad} & \textsc{mad} & Indonesia & 1200 \\
            \texttt{nusatranslation\_senti\_mak} & \textsc{mak} & Indonesia & 1200 \\
            \texttt{nusatranslation\_senti\_min} & \textsc{min} & Indonesia & 1200 \\
            \texttt{nusatranslation\_senti\_mui} & \textsc{mui} & Indonesia & 500 \\
            \texttt{nusatranslation\_senti\_rej} & \textsc{rej} & Indonesia & 500 \\
            \texttt{nusatranslation\_senti\_sun} & \textsc{sun} & Indonesia & 1200 \\
            \texttt{nusax\_senti\_ind} & \textsc{ind} & Indonesia & 400 \\
            \texttt{nusax\_senti\_ace} & \textsc{ace} & Indonesia & 400 \\
            \texttt{nusax\_senti\_jav} & \textsc{jav} & Indonesia & 400 \\
            \texttt{nusax\_senti\_sun} & \textsc{sun} & Indonesia & 400 \\
            \texttt{nusax\_senti\_min} & \textsc{min} & Indonesia & 400 \\
            \texttt{nusax\_senti\_bug} & \textsc{bug} & Indonesia & 400 \\
            \texttt{nusax\_senti\_bbc} & \textsc{bbc} & Indonesia & 400 \\
            \texttt{nusax\_senti\_ban} & \textsc{ban} & Indonesia & 400 \\
            \texttt{nusax\_senti\_nij} & \textsc{nij} & Indonesia & 400 \\
            \texttt{nusax\_senti\_mad} & \textsc{mad} & Indonesia & 400 \\
            \texttt{nusax\_senti\_bjn} & \textsc{bjn} & Indonesia & 400 \\
            \texttt{nusax\_senti\_eng} & \textsc{eng} & Non-indigenous & 400 \\
            \texttt{indonglish} & \textsc{ind} & Indonesia & 1011 \\
        \bottomrule
        \end{tabular}
        }
    \end{subtable}
    \caption{Sentiment analysis data subsets used in SEACrowd NLU evaluation.}
    \label{tab:sea-eval-nlu-senti-subset}
\end{table}

\begin{table}[!]
    \centering
    \begin{subtable}[t]{\linewidth}
        \centering
        \resizebox{0.75\linewidth}{!}{
        \begin{tabular}{lclr}
          \toprule
          \textbf{Subset ID} & \textbf{Language} & \textbf{Region} & \textbf{\# Samples} \\
          \toprule
            \multicolumn{4}{c}{\cellcolor[HTML]{D9D9D9}\textit{\textbf{NLI} $\rightarrow$ \texttt{*\_seacrowd\_pairs}}} \\
            \texttt{indonli} & \textsc{ind} & Indonesia & 5183 \\
            \texttt{wrete} & \textsc{ind} & Indonesia & 100 \\
            \texttt{snli\_indo} & \textsc{ind} & Indonesia & 9823 \\
            \texttt{myxnli} & \textsc{mya} & Myanmar & 5010 \\
            \texttt{xnli.tha} & \textsc{tha} & Thailand & 5010 \\
            \texttt{xnli.vie} & \textsc{vie} & Vietnam & 5010 \\
        \bottomrule
        \end{tabular}
        }
    \end{subtable}
    \caption{NLI data subsets used in SEACrowd NLU evaluation.}
    \label{tab:sea-eval-nlu-nli-subset}
    \vspace{-0.5cm}
\end{table}

\begin{table}[t]
    \centering
    \begin{subtable}[t]{\linewidth}
        \centering
        \resizebox{\linewidth}{!}{
        \begin{tabular}{lclr}
          \toprule
          \textbf{Subset ID} & \textbf{Language} & \textbf{Region} & \textbf{\# Samples} \\
          \toprule
            \multicolumn{4}{c}{\cellcolor[HTML]{D9D9D9}\textit{\textbf{Topic Classification} $\rightarrow$ \texttt{*\_seacrowd\_text}}} \\
            \texttt{gklmip\_newsclass} & \textsc{khm} & Cambodia & 1436 \\
            \texttt{indonesian\_news\_dataset} & \textsc{ind} & Indonesia & 2627 \\
            \texttt{uit\_vion} & \textsc{vie} & Vietnam & 26000 \\
            \texttt{sib\_200\_ace\_Arab} & \textsc{ace} & Indonesia & 204 \\
            \texttt{sib\_200\_ace\_Latn} & \textsc{ace} & Indonesia & 204 \\
            \texttt{sib\_200\_ban\_Latn} & \textsc{ban} & Indonesia & 204 \\
            \texttt{sib\_200\_bjn\_Arab} & \textsc{bjn} & Indonesia & 204 \\
            \texttt{sib\_200\_bjn\_Latn} & \textsc{bjn} & Indonesia & 204 \\
            \texttt{sib\_200\_bug\_Latn} & \textsc{bug} & Indonesia & 204 \\
            \texttt{sib\_200\_ceb\_Latn} & \textsc{ceb} & Philippines & 204 \\
            \texttt{sib\_200\_ilo\_Latn} & \textsc{ilo} & Philippines & 204 \\
            \texttt{sib\_200\_ind\_Latn} & \textsc{ind} & Indonesia & 204 \\
            \texttt{sib\_200\_jav\_Latn} & \textsc{jav} & Indonesia & 204 \\
            \texttt{sib\_200\_kac\_Latn} & \textsc{kac} & Myanmar & 204 \\
            \texttt{sib\_200\_khm\_Khmr} & \textsc{khm} & Cambodia & 204 \\
            \texttt{sib\_200\_lao\_Laoo} & \textsc{lao} & Laos & 204 \\
            \texttt{sib\_200\_lus\_Latn} & \textsc{lus} & Myanmar & 204 \\
            \texttt{sib\_200\_min\_Arab} & \textsc{min} & Indonesia & 204 \\
            \texttt{sib\_200\_min\_Latn} & \textsc{min} & Indonesia & 204 \\
            \texttt{sib\_200\_mya\_Mymr} & \textsc{mya} & Myanmar & 204 \\
            \texttt{sib\_200\_pag\_Latn} & \textsc{pag} & Philippines & 204 \\
            \texttt{sib\_200\_shn\_Mymr} & \textsc{shn} & Myanmar & 204 \\
            \texttt{sib\_200\_sun\_Latn} & \textsc{sun} & Indonesia & 204 \\
            \texttt{sib\_200\_tgl\_Latn} & \textsc{fil} & Philippines & 204 \\
            \texttt{sib\_200\_tha\_Thai} & \textsc{tha} & Thailand & 204 \\
            \texttt{sib\_200\_vie\_Latn} & \textsc{vie} & Non-indigenous & 204 \\
            \texttt{sib\_200\_war\_Latn} & \textsc{war} & Philippines & 204 \\
            \texttt{sib\_200\_zsm\_Latn} & \textsc{zsm} & Malaysia & 204 \\
            \texttt{nusaparagraph\_topic\_btk} & \textsc{btx} & Indonesia & 500 \\
            \texttt{nusaparagraph\_topic\_bew} & \textsc{bew} & Indonesia & 800 \\
            \texttt{nusaparagraph\_topic\_bug} & \textsc{bug} & Indonesia & 300 \\
            \texttt{nusaparagraph\_topic\_jav} & \textsc{jav} & Indonesia & 800 \\
            \texttt{nusaparagraph\_topic\_mad} & \textsc{mad} & Indonesia & 700 \\
            \texttt{nusaparagraph\_topic\_mak} & \textsc{mak} & Indonesia & 700 \\
            \texttt{nusaparagraph\_topic\_min} & \textsc{min} & Indonesia & 800 \\
            \texttt{nusaparagraph\_topic\_mui} & \textsc{mui} & Indonesia & 400 \\
            \texttt{nusaparagraph\_topic\_rej} & \textsc{rej} & Indonesia & 350 \\
            \texttt{nusaparagraph\_topic\_sun} & \textsc{sun} & Indonesia & 900 \\
        \bottomrule
        \end{tabular}
        }
    \end{subtable}
    \caption{Topic classification data subsets used in SEACrowd NLU evaluation.}
    \label{tab:sea-eval-nlu-topic-subset}
\end{table}

\begin{table}[!]
    \centering
    \begin{subtable}[t]{\linewidth}
        \centering
        \resizebox{0.9\linewidth}{!}{
        \begin{tabular}{lclr}
          \toprule
          \textbf{Subset ID} & \textbf{Language} & \textbf{Region} & \textbf{\# Samples} \\
          \toprule
            \multicolumn{4}{c}{\cellcolor[HTML]{D9D9D9}\textit{\textbf{Commonsense Reasoning} $\rightarrow$ \texttt{*\_seacrowd\_text/qa}}} \\
            \texttt{emotes\_3k\_tgl} & \textsc{fil} & Philippines & 2905 \\
            \texttt{emotes\_3k\_eng} & \textsc{eng} & Non-indigenous & 2905 \\
            \texttt{indo\_story\_cloze} & \textsc{ind} & Indonesia & 1135 \\
            \texttt{xstorycloze\_id} & \textsc{ind} & Indonesia & 1511 \\
            \texttt{xstorycloze\_my} & \textsc{mya} & Myanmar & 1511 \\
        \bottomrule
        \end{tabular}
        }
    \end{subtable}
    \caption{Commonsense reasoning data subsets used in SEACrowd NLU evaluation.}
    \label{tab:sea-eval-nlu-commonsense-subset}
\end{table}

\begin{table}[t]
    \centering
    \begin{subtable}[t]{\linewidth}
        \centering
        \resizebox{\linewidth}{!}{
        \begin{tabular}{lclr}
          \toprule
          \textbf{Subset ID} & \textbf{Language} & \textbf{Region} & \textbf{\# Samples} \\
          \toprule
            \multicolumn{4}{c}{\cellcolor[HTML]{D9D9D9}\textit{\textbf{Standard Testing QA} $\rightarrow$ \texttt{*\_seacrowd\_qa}}} \\
            \texttt{indommlu\_ind} & \textsc{ind} & Indonesia & 14979 \\
            \texttt{indommlu\_ban} & \textsc{ban} & Indonesia & 14979 \\
            \texttt{indommlu\_mad} & \textsc{mad} & Indonesia & 14979 \\
            \texttt{indommlu\_mak} & \textsc{mak} & Indonesia & 14979 \\
            \texttt{indommlu\_sun} & \textsc{sun} & Indonesia & 14979 \\
            \texttt{indommlu\_jav} & \textsc{jav} & Indonesia & 14979 \\
            \texttt{indommlu\_bjn} & \textsc{bjn} & Indonesia & 14979 \\
            \texttt{indommlu\_abl} & \textsc{abl} & Indonesia & 14979 \\
            \texttt{indommlu\_nij} & \textsc{nij} & Indonesia & 14979 \\
            \texttt{seaeval\_cross\_mmlu\_ind} & \textsc{ind} & Indonesia & 150 \\
            \texttt{seaeval\_cross\_mmlu\_vie} & \textsc{vie} & Vietnam & 150 \\
            \texttt{seaeval\_cross\_mmlu\_zlm} & \textsc{zsm} & Malaysia & 150 \\
            \texttt{seaeval\_cross\_mmlu\_fil} & \textsc{fil} & Philippines & 150 \\
            \texttt{seaeval\_cross\_logiqa\_ind} & \textsc{ind} & Indonesia & 176 \\
            \texttt{seaeval\_cross\_logiqa\_vie} & \textsc{vie} & Vietnam & 176 \\
            \texttt{seaeval\_cross\_logiqa\_zlm} & \textsc{zsm} & Malaysia & 176 \\
            \texttt{seaeval\_cross\_logiqa\_fil} & \textsc{fil} & Philippines & 176 \\
            \texttt{m3exam\_jav} & \textsc{jav} & Indonesia & 371 \\
            \texttt{m3exam\_tha} & \textsc{tha} & Thailand & 2168 \\
            \texttt{m3exam\_vie} & \textsc{vie} & Vietnam & 1789 \\
            \texttt{okapi\_m\_arc\_ind} & \textsc{ind} & Indonesia & 1170 \\
            \texttt{okapi\_m\_arc\_vie} & \textsc{vie} & Vietnam & 1170 \\
            
            \multicolumn{4}{c}{\cellcolor[HTML]{D9D9D9}\textit{\textbf{Cultural QA} $\rightarrow$ \texttt{*\_seacrowd\_qa}}} \\
            \texttt{copal\_colloquial} & \textsc{ind} & Indonesia & 559 \\
            \texttt{xcopa\_tha} & \textsc{tha} & Thailand & 500 \\
            \texttt{xcopa\_vie} & \textsc{vie} & Vietnam & 500 \\
            \texttt{xcopa\_ind} & \textsc{ind} & Indonesia & 500 \\
            \texttt{seaeval\_sg\_eval\_eng} & \textsc{eng} & Non-indigenous & 103 \\
            \texttt{seaeval\_ph\_eval\_eng} & \textsc{eng} & Non-indigenous & 100 \\
            \texttt{mabl\_ind} & \textsc{ind} & Indonesia & 1140 \\
            \texttt{mabl\_jav} & \textsc{jav} & Indonesia & 600 \\
            \texttt{mabl\_sun} & \textsc{sun} & Indonesia & 600 \\

            \multicolumn{4}{c}{\cellcolor[HTML]{D9D9D9}\textit{\textbf{Reading Comprehension QA} $\rightarrow$ \texttt{*\_seacrowd\_qa}}} \\
            \texttt{belebele\_ceb\_latn} & \textsc{ceb} & Philippines & 900 \\
            \texttt{belebele\_ilo\_latn} & \textsc{ilo} & Philippines & 900 \\
            \texttt{belebele\_ind\_latn} & \textsc{ind} & Indonesia & 900 \\
            \texttt{belebele\_jav\_latn} & \textsc{jav} & Indonesia & 900 \\
            \texttt{belebele\_kac\_latn} & \textsc{kac} & Myanmar & 900 \\
            \texttt{belebele\_khm\_khmr} & \textsc{khm} & Cambodia & 900 \\
            \texttt{belebele\_lao\_laoo} & \textsc{lao} & Laos & 900 \\
            \texttt{belebele\_mya\_mymr} & \textsc{mya} & Myanmar & 900 \\
            \texttt{belebele\_shn\_mymr} & \textsc{shn} & Myanmar & 900 \\
            \texttt{belebele\_sun\_latn} & \textsc{sun} & Indonesia & 900 \\
            \texttt{belebele\_tgl\_latn} & \textsc{fil} & Philippines & 900 \\
            \texttt{belebele\_tha\_thai} & \textsc{tha} & Thailand & 900 \\
            \texttt{belebele\_vie\_latn} & \textsc{vie} & Vietnam & 900 \\
            \texttt{belebele\_war\_latn} & \textsc{war} & Philippines & 900 \\
            \texttt{belebele\_zsm\_latn} & \textsc{zsm} & Malaysia & 900 \\
        \bottomrule
        \end{tabular}
        }
    \end{subtable}
    \caption{Multiple-choice QA data subsets used in SEACrowd NLU evaluation.}
    \label{tab:sea-eval-nlu-mcqa-subset}
\end{table}

\begin{table}[t]
\begin{subtable}[t]{\linewidth}
    \centering
    \resizebox{0.8\linewidth}{!}{
    \begin{tabular}{lclr}
      \toprule
      \textbf{Subset ID} & \textbf{Language} & \textbf{Region} & \textbf{\# Samples} \\
      \toprule
        \multicolumn{4}{c}{\cellcolor[HTML]{D9D9D9}\textit{\textbf{Extractive \& Abstractive QA} $\rightarrow$ \texttt{*\_seacrowd\_qa}}} \\
        \texttt{facqa} & \textsc{ind} & Indonesia & 311 \\
        \texttt{iapp\_squad} & \textsc{tha} & Thailand & 739 \\
        \texttt{qasina} & \textsc{ind} & Indonesia & 500 \\
        \texttt{mkqa\_khm} & \textsc{khm} & Cambodia & 10000 \\
        \texttt{mkqa\_zsm} & \textsc{zsm} & Malaysia & 10000 \\
        \texttt{mkqa\_tha} & \textsc{tha} & Thailand & 10000 \\
        \texttt{mkqa\_vie} & \textsc{vie} & Vietnam & 10000 \\
      \bottomrule
    \end{tabular}
    }
\end{subtable}
    \caption{Extractive and abstractive QA subsets used in SEACrowd NLG evaluation.}
    \label{tab:sea-eval-nlg-qa-subset}
\end{table}

\begin{table}[t]
    \centering
    \begin{subtable}[t]{\linewidth}
        \centering
        \resizebox{0.8\linewidth}{!}{
        \begin{tabular}{lclr}
          \toprule
          \textbf{Subset ID} & \textbf{Language} & \textbf{Region} & \textbf{\# Samples} \\
          \toprule
            \multicolumn{4}{c}{\cellcolor[HTML]{D9D9D9}\textit{\textbf{Summarization} $\rightarrow$ \texttt{*\_seacrowd\_t2t}}} \\
            \texttt{lr\_sum\_ind} & \textsc{ind} & Indonesia & 500 \\
            \texttt{lr\_sum\_vie} & \textsc{vie} & Vietnam & 1460 \\
            \texttt{lr\_sum\_lao} & \textsc{lao} & Laos & 1496 \\
            \texttt{lr\_sum\_tha} & \textsc{tha} & Thailand & 500 \\
            \texttt{lr\_sum\_khm} & \textsc{khm} & Cambodia & 486 \\
            \texttt{lr\_sum\_mya} & \textsc{mya} & Myanmar & 990 \\
            \texttt{xl\_sum\_mya} & \textsc{mya} & Myanmar & 570 \\
            \texttt{xl\_sum\_ind} & \textsc{ind} & Indonesia & 4780 \\
            \texttt{xl\_sum\_tha} & \textsc{tha} & Thailand & 826 \\
            \texttt{xl\_sum\_vie} & \textsc{vie} & Vietnam & 4013 \\
        \bottomrule
        \end{tabular}
        }
    \end{subtable}
    \caption{Summarization data subsets used in SEACrowd NLG evaluation.}
    \label{tab:sea-eval-nlg-sum-subset}
\end{table}

\subsection{NLP}
\label{app:schemas-nlp}

\begin{compactitem}
    \item \textbf{Unlabeled text (\texttt{SSP})}. This schema could be used for language modeling in self-supervised pre-training. It consists of \texttt{(id, text)}, where \texttt{id} denotes a unique row identifier of the dataset and \texttt{text} denotes an input text.
    
    \item \textbf{Single-label text classification (\texttt{TEXT})}. This schema could be used for sentiment analysis, emotion classification, legal classification, and others. It consists of \texttt{(id, text, label)}, where \texttt{id} denotes a unique row identifier of the dataset, \texttt{text} denotes an input text, and \texttt{label} denotes a deterministic target variable.

    \item \textbf{Multi-label text classification (\texttt{TEXT MULTI})}. This schema could be used for hate speech detection and aspect-based sentiment analysis. It consists of \texttt{(id, text, labels)}, where \texttt{id} denotes a unique row identifier of the dataset, \texttt{text} denotes an input text, and \texttt{labels} denotes a list of deterministic target variables.
    
    \item \textbf{Text-to-text (\texttt{T2T})}. This schema could be used for machine translation, summarization, and paraphrasing. It consists of \texttt{(id, text\_1, text\_2, text\_1\_name, text\_2\_name)}, where \texttt{id} denotes a unique row identifier of the dataset, \texttt{text\_1} and \texttt{text\_2} denote an input text pair, and \texttt{text\_1\_name} and \texttt{text\_2\_name} denote the names of the input text pair (e.g., \texttt{ind} and \texttt{jav} for translation input text pairs, or \texttt{document} and \texttt{summary} for summarization input text pairs).

    \item \textbf{Sequence labeling (\texttt{SEQ LABEL})}. This schema could be used for named entity recognition (NER), POS tagging, and others. It consists of \texttt{(id, tokens, labels)}, where \texttt{id} denotes a unique row identifier of the dataset, \texttt{tokens} denotes a list of tokens of an input text, and \texttt{labels} denotes a list of targets for the tokens.

    \item \textbf{Question answering (\texttt{QA})}. This schema could be used for extractive QA, multiple-choice QA, and others. It consists of \texttt{(id, question\_id, document\_id, question, type, choices, context, answer)}, where \texttt{id} denotes a unique row identifier of the dataset, \texttt{question\_id} denotes a unique identifier of the question, \texttt{document\_id} denotes a unique identifier of the context document, \texttt{question} denotes an input question to be answered, \texttt{type} denotes the type of the QA task (e.g., extractive, multiple-choice, open-generative, closed-generative, etc.), \texttt{choices} denotes a list of answer choices (if required), \texttt{context} denotes a passage that serves as the background information of the question (if required), and \texttt{answer} denotes the gold answer to the question (if required).

    \item \textbf{Single-label text pair classification \texttt{(PAIRS)}}. This could be used for textual entailment and next-sentence prediction. It consists of \texttt{(id, text\_1, text\_2, label)}, where \texttt{id} denotes a unique row identifier of the dataset, \texttt{text\_1} and \texttt{text\_2} denote an input text pair, and \texttt{label} denotes the target variable.

    \item \textbf{Single-label text pair classification with continuous values or regression \texttt{(PAIRS SCORE)}}. This could be used for answer grading and semantic textual similarity. It consists of \texttt{(id, text\_1, text\_2, label)}, where \texttt{id} denotes a unique row identifier of the dataset, \texttt{text\_1} and \texttt{text\_2} denote an input text pair, and \texttt{label} denotes a target variable as a continuous value.

    \item \textbf{Multi-label text pair classification (\texttt{PAIRS MULTI})}. This could be used for morphological inflection. It consists of \texttt{(id, text\_1, text\_2, labels)}, where \texttt{id} denotes a unique row identifier of the dataset, \texttt{text\_1} and \texttt{text\_2} denote an input text pair, and \texttt{labels} denotes a list of target variables.

    \item \textbf{Knowledge base (\texttt{KB})}. This schema could be used for constituency parsing, dependency parsing, coreference resolution, dialogue systems, and other tasks with complex structures. It consists of \texttt{(id, passages, entities, events, coreferences, relations)}. Considering its intricate structure, we encourage readers to take a look at the implementation of the knowledge base schema.

    \item \textbf{Tree (\texttt{TREE})}. This schema could be used for constituency parsing, this schema assumes a document with subnode elements and a tree hierarchy. It consists of \texttt{(id, passage, nodes)}, where \texttt{id} denotes a unique row identifier of the dataset, \texttt{passage} denotes the passage to that particular \texttt{id}; this \texttt{passage} consist of \texttt{(id, type, text, offsets)}, \texttt{nodes} denotes the nodes to that particular \texttt{id}; this \texttt{nodes} consists of \texttt{(id, type, text, offsets, subnodes)}.

    \item \textbf{Conversational Chat (\texttt{CHAT})}. This schema could be used for conversational chat and/or multi-turn conversation. It consists of \texttt{(id, input, output, meta)}, where \texttt{id} denotes a unique row identifier of the dataset, \texttt{input} denotes a sequence that consists of \texttt{content} and \texttt{role} as an input prompt and the role of the entity inputting the prompt, \texttt{output} denotes an answer from that input prompt, and \texttt{meta} denotes relevant details to allow some flexibility of the schema (if required).

    \item \textbf{End-to-end Task Oriented Dialogue (\texttt{TOD})}. This schema could be used for end-to-end task-oriented dialogue. It consists of \texttt{(dialogue\_idx, dialogue)}, where \texttt{dialogue\_idx} denotes a unique row identifier of the dialogue, \texttt{dialogue} denotes some core details such as \texttt{turn label}, \texttt{system utterance}, \texttt{turn id}x, \texttt{belief state} (consist of \texttt{slots} and \texttt{act}), \texttt{user utterance}, and \texttt{system acts}.
\end{compactitem}

\begin{table}[!]
    \centering
    \begin{subtable}[t]{\linewidth}
        \centering
        \resizebox{0.8\linewidth}{!}{
        \begin{tabular}{lclr}
          \toprule
          \textbf{Subset ID} & \textbf{Language} & \textbf{Region} & \textbf{\# Samples} \\
          \toprule
            \multicolumn{4}{c}{\cellcolor[HTML]{D9D9D9}\textit{\textbf{Image Captioning} $\rightarrow$ \texttt{*\_seacrowd\_imtext}}} \\
            \texttt{xm3600\_fil} & \textsc{fil} & Philippines & 2760 \\
            \texttt{xm3600\_id} & \textsc{ind} & Indonesia & 2775 \\
            \texttt{xm3600\_th} & \textsc{tha} & Thailand & 2798 \\
            \texttt{xm3600\_vi} & \textsc{vie} & Vietnam & 2855 \\
        \bottomrule
        \end{tabular}
        }
    \end{subtable}
    \caption{Image captioning data subsets used in SEACrowd VL evaluation.}
    \label{tab:sea-eval-vl-img-cap-subset}
\end{table}

\subsection{Speech}
\label{app:schemas-speech}

\begin{compactitem}
    \item \textbf{Speech-text (\texttt{SPTEXT})}. This could be used for speech recognition, text-to-speech (TTS) or speech synthesis, and speech-to-text translation. It consists of \texttt{(id, path, audio, text, speaker\_id, metadata)}, where \texttt{id} denotes a unique row identifier of the dataset, \texttt{path} denotes the file path to an input audio source, \texttt{audio} denotes the audio data loaded from the corresponding \texttt{path}, \texttt{text} denotes an input text, \texttt{speaker\_id} denotes a unique identifier of the speaker, \texttt{metadata} denotes relevant details such as the age and gender of the speaker (if required).

    \item \textbf{Speech-to-speech (\texttt{S2S})}. This could be used for speech-to-speech translation. It consists of \texttt{(id, path\_1, audio\_1, text\_1, metadata\_1, path\_2, audio\_2, text\_2, metadata\_2)}, where \texttt{id} denotes a unique row identifier of the dataset, \texttt{path\_1} and \texttt{path\_2} denote the file path to a respective input audio source, \texttt{audio\_1} and \texttt{audio\_2} denote the audio data loaded from the corresponding \texttt{path}, \texttt{text\_1} and \texttt{text\_2} denote input texts, and \texttt{metadata\_1} and \texttt{metadata\_2} denote relevant details such as the age of the speaker and their gender (if required).

    \item \textbf{Speech Classification (\texttt{SPEECH})}. This schema could be used for speech classification, speech-language identification, and speech-emotion recognition for single-label use only. It consists of \texttt{(id, path, audio, speaker\_id, labels, metadata)}, where \texttt{id} denotes a unique row identifier of the dataset, \texttt{path} denotes the file path to an input audio source, \texttt{audio} denotes the audio data loaded from the corresponding \texttt{path}, \texttt{speaker\_id} denotes a unique identifier of the speaker, \texttt{labels} denotes the label of that particular speech (only can be single-label), \texttt{metadata} denotes relevant details such as the age and gender of the speaker (if required).
    
    \item \textbf{Speech Classification for Multilabel (\texttt{SPEECH MULTILABEL})}. This schema could be used for speech classification, speech-language identification, and speech-emotion recognition for multi-label use only. It consists of \texttt{(id, path, audio, speaker\_id, labels, metadata)}, where \texttt{id} denotes a unique row identifier of the dataset, \texttt{path} denotes the file path to an input audio source, \texttt{audio} denotes the audio data loaded from the corresponding \texttt{path}, \texttt{speaker\_id} denotes a unique identifier of the speaker, \texttt{labels} denotes the sequence of labels of that particular speech (only can be multi-label), \texttt{metadata} denotes relevant details such as the age and gender of the speaker (if required).
\end{compactitem}

\begin{table}[!]
    \centering
    \begin{subtable}[t]{\linewidth}
        \centering
        \resizebox{\linewidth}{!}{
        \begin{tabular}{lclr}
          \toprule
          \textbf{Subset ID} & \textbf{Language} & \textbf{Region} & \textbf{\# Samples} \\
          \toprule
            \multicolumn{4}{c}{\cellcolor[HTML]{D9D9D9}\textit{\textbf{ASR} $\rightarrow$ \texttt{*\_seacrowd\_sptext}}} \\
            \texttt{asr\_ibsc} & \textsc{iba} & Brunei & 473 \\
            \texttt{commonvoice\_120\_ind} & \textsc{ind} & Indonesia & 3647 \\
            \texttt{commonvoice\_120\_tha} & \textsc{tha} & Thailand & 10964 \\
            \texttt{commonvoice\_120\_cnh} & \textsc{cnh} & Myanmar & 763 \\
            \texttt{commonvoice\_120\_vie} & \textsc{vie} & Vietnam & 1302 \\
            \texttt{fleurs\_ind} & \textsc{ind} & Indonesia & 687 \\
            \texttt{fleurs\_jav} & \textsc{jav} & Indonesia & 728 \\
            \texttt{fleurs\_tha} & \textsc{tha} & Thailand & 1021 \\
            \texttt{fleurs\_lao} & \textsc{lao} & Laos & 405 \\
            \texttt{fleurs\_mya} & \textsc{mya} & Myanmar & 880 \\
            \texttt{fleurs\_khm} & \textsc{khm} & Cambodia & 771 \\
            \texttt{fleurs\_vie} & \textsc{vie} & Vietnam & 857 \\
            \texttt{fleurs\_zlm} & \textsc{zlm} & Malaysia & 749 \\
            \texttt{fleurs\_fil} & \textsc{fil} & Philippines & 964 \\
            \texttt{fleurs\_ceb} & \textsc{ceb} & Philippines & 541 \\
            \texttt{indspeech\_newstra\_ethnicsr\_nooverlap\_jav} & \textsc{jav} & Indonesia & 1000 \\
            \texttt{indspeech\_newstra\_ethnicsr\_nooverlap\_sun} & \textsc{sun} & Indonesia & 1000 \\
            \texttt{indspeech\_newstra\_ethnicsr\_nooverlap\_ban} & \textsc{ban} & Indonesia & 1000 \\
            \texttt{indspeech\_newstra\_ethnicsr\_nooverlap\_btk} & \textsc{btx} & Indonesia & 1000 \\
        \bottomrule
        \end{tabular}
        }
    \end{subtable}
    \caption{ASR data subsets used in SEACrowd speech evaluation.}
    \label{tab:sea-eval-speech-asr-subset}
\end{table}

\begin{table*}[t]
\begin{subtable}[t]{\linewidth}
    \centering
    \resizebox{.8\linewidth}{!}{
    \begin{tabular}{llclr}
      \toprule
      \multicolumn{2}{c}{\textbf{Subset ID}} & \multirow[c]{2}{*}{\textbf{Language}} & \multirow[c]{2}{*}{\textbf{Region}} & \multirow[c]{2}{*}{\textbf{\# Samples}} \\
      \multicolumn{1}{c}{Eng $\rightarrow$ XX}  &  \multicolumn{1}{c}{XX $\rightarrow$ Eng} & & \\
      \toprule
        \multicolumn{5}{c}{\cellcolor[HTML]{D9D9D9}\textit{\textbf{MT (Eng $\Leftrightarrow$ XX)} $\rightarrow$ \texttt{*\_seacrowd\_t2t}}} \\
        \texttt{lio\_and\_central\_flores\_eng\_ljl} & \texttt{lio\_and\_central\_flores\_ljl\_eng} & \textsc{ljl} & Indonesia & 1658 \\
        \texttt{flores200\_eng\_Latn\_ace\_Latn} & \texttt{flores200\_ace\_Latn\_eng\_Latn} & \textsc{ace} & Indonesia & 1012 \\
        \texttt{flores200\_eng\_Latn\_ban\_Latn} & \texttt{flores200\_ban\_Latn\_eng\_Latn} & \textsc{ban} & Indonesia & 1012 \\
        \texttt{flores200\_eng\_Latn\_bjn\_Latn} & \texttt{flores200\_bjn\_Latn\_eng\_Latn} & \textsc{bjn} & Indonesia & 1012 \\
        \texttt{flores200\_eng\_Latn\_bug\_Latn} & \texttt{flores200\_bug\_Latn\_eng\_Latn} & \textsc{bug} & Indonesia & 1012 \\
        \texttt{flores200\_eng\_Latn\_ceb\_Latn} & \texttt{flores200\_ceb\_Latn\_eng\_Latn} & \textsc{ceb} & Philippines & 1012 \\
        \texttt{flores200\_eng\_Latn\_ilo\_Latn} & \texttt{flores200\_ilo\_Latn\_eng\_Latn} & \textsc{ilo} & Philippines & 1012 \\
        \texttt{flores200\_eng\_Latn\_ind\_Latn} & \texttt{flores200\_ind\_Latn\_eng\_Latn} & \textsc{ind} & Indonesia & 1012 \\
        \texttt{flores200\_eng\_Latn\_jav\_Latn} & \texttt{flores200\_jav\_Latn\_eng\_Latn} & \textsc{jav} & Indonesia & 1012 \\
        \texttt{flores200\_eng\_Latn\_kac\_Latn} & \texttt{flores200\_kac\_Latn\_eng\_Latn} & \textsc{kac} & Myanmar & 1012 \\
        \texttt{flores200\_eng\_Latn\_khm\_Khmr} & \texttt{flores200\_khm\_Khmr\_eng\_Latn} & \textsc{khm} & Cambodia & 1012 \\
        \texttt{flores200\_eng\_Latn\_lao\_Laoo} & \texttt{flores200\_lao\_Laoo\_eng\_Latn} & \textsc{lao} & Laos & 1012 \\
        \texttt{flores200\_eng\_Latn\_lus\_Latn} & \texttt{flores200\_lus\_Latn\_eng\_Latn} & \textsc{lus} & Myanmar & 1012 \\
        \texttt{flores200\_eng\_Latn\_min\_Latn} & \texttt{flores200\_min\_Latn\_eng\_Latn} & \textsc{min} & Indonesia & 1012 \\
        \texttt{flores200\_eng\_Latn\_mya\_Mymr} & \texttt{flores200\_mya\_Mymr\_eng\_Latn} & \textsc{mya} & Myanmar & 1012 \\
        \texttt{flores200\_eng\_Latn\_pag\_Latn} & \texttt{flores200\_pag\_Latn\_eng\_Latn} & \textsc{pag} & Philippines & 1012 \\
        \texttt{flores200\_eng\_Latn\_shn\_Mymr} & \texttt{flores200\_shn\_Mymr\_eng\_Latn} & \textsc{shn} & Myanmar & 1012 \\
        \texttt{flores200\_eng\_Latn\_sun\_Latn} & \texttt{flores200\_sun\_Latn\_eng\_Latn} & \textsc{sun} & Indonesia & 1012 \\
        \texttt{flores200\_eng\_Latn\_tha\_Thai} & \texttt{flores200\_tha\_Thai\_eng\_Latn} & \textsc{tha} & Thailand & 1012 \\
        \texttt{flores200\_eng\_Latn\_vie\_Latn} & \texttt{flores200\_vie\_Latn\_eng\_Latn} & \textsc{vie} & Vietnam & 1012 \\
        \texttt{flores200\_eng\_Latn\_war\_Latn} & \texttt{flores200\_war\_Latn\_eng\_Latn} & \textsc{war} & Philippines & 1012 \\
        \texttt{flores200\_eng\_Latn\_zsm\_Latn} & \texttt{flores200\_zsm\_Latn\_eng\_Latn} & \textsc{zsm} & Malaysia & 1012 \\
        \texttt{ntrex\_128\_eng-US\_ind} & \texttt{ntrex\_128\_ind\_eng-US} & \textsc{ind} & Indonesia & 1997 \\
        \texttt{ntrex\_128\_eng-US\_mya} & \texttt{ntrex\_128\_mya\_eng-US} & \textsc{mya} & Myanmar & 1997 \\
        \texttt{ntrex\_128\_eng-US\_fil} & \texttt{ntrex\_128\_fil\_eng-US} & \textsc{fil} & Philippines & 1997 \\
        \texttt{ntrex\_128\_eng-US\_khm} & \texttt{ntrex\_128\_khm\_eng-US} & \textsc{khm} & Cambodia & 1997 \\
        \texttt{ntrex\_128\_eng-US\_lao} & \texttt{ntrex\_128\_lao\_eng-US} & \textsc{lao} & Laos & 1997 \\
        \texttt{ntrex\_128\_eng-US\_zlm} & \texttt{ntrex\_128\_zlm\_eng-US} & \textsc{zsm} & Malaysia & 1997 \\
        \texttt{ntrex\_128\_eng-US\_tha} & \texttt{ntrex\_128\_tha\_eng-US} & \textsc{tha} & Thailand & 1997 \\
        \texttt{ntrex\_128\_eng-US\_vie} & \texttt{ntrex\_128\_vie\_eng-US} & \textsc{vie} & Vietnam & 1997 \\
        \texttt{ntrex\_128\_eng-US\_hmv} & \texttt{ntrex\_128\_hmv\_eng-US} & \textsc{hmv} & Vietnam & 1997 \\
        \texttt{nusax\_mt\_eng\_ind} & - & \textsc{ind} & Indonesia & 400 \\
        \texttt{nusax\_mt\_eng\_ace} & \texttt{nusax\_mt\_ace\_eng} & \textsc{ace} & Indonesia & 400 \\
        \texttt{nusax\_mt\_eng\_jav} & \texttt{nusax\_mt\_jav\_eng} & \textsc{jav} & Indonesia & 400 \\
        \texttt{nusax\_mt\_eng\_sun} & \texttt{nusax\_mt\_sun\_eng} & \textsc{sun} & Indonesia & 400 \\
        \texttt{nusax\_mt\_eng\_min} & \texttt{nusax\_mt\_min\_eng} & \textsc{min} & Indonesia & 400 \\
        \texttt{nusax\_mt\_eng\_bug} & \texttt{nusax\_mt\_bug\_eng} & \textsc{bug} & Indonesia & 400 \\
        \texttt{nusax\_mt\_eng\_bbc} & \texttt{nusax\_mt\_bbc\_eng} & \textsc{bbc} & Indonesia & 400 \\
        \texttt{nusax\_mt\_eng\_ban} & \texttt{nusax\_mt\_ban\_eng} & \textsc{ban} & Indonesia & 400 \\
        \texttt{nusax\_mt\_eng\_nij} & \texttt{nusax\_mt\_nij\_eng} & \textsc{nij} & Indonesia & 400 \\
        \texttt{nusax\_mt\_eng\_mad} & \texttt{nusax\_mt\_mad\_eng} & \textsc{mad} & Indonesia & 400 \\
        \texttt{nusax\_mt\_eng\_bjn} & \texttt{nusax\_mt\_bjn\_eng} & \textsc{bjn} & Indonesia & 400 \\
    \bottomrule
    \end{tabular}
    }
    \end{subtable}
    \caption{MT between English and SEA languages data subsets used in SEACrowd NLG evaluation.}
    \label{tab:sea-eval-nlg-mt-subset}
\end{table*}

\subsection{VL}
\label{app:schemas-vl}

\begin{compactitem}
    \item \textbf{Image-text (\texttt{IMTEXT})}. This schema could be used for image captioning, text-to-image generation, and vision-language pre-training. It consists of \texttt{(id, text, image\_paths, metadata)}, where \texttt{id} denotes a unique row identifier of the dataset, \texttt{text} denotes an input text, \texttt{image\_paths} denotes a list of paths to the input image sources, and \texttt{metadata} denotes relevant details such as visual concepts and labels (if required).
    
    \item \textbf{General Image Classification (\texttt{IMAGE})}. This schema could be used for image classification both single-label and multi-label. It consists of \texttt{(id, labels, image\_path, metadata)}, where \texttt{id} denotes a unique row identifier of the dataset, \texttt{labels} denotes the label of that particular image (can be single-label and multi-label), \texttt{image\_path} denotes a list of paths to the input image sources, and \texttt{metadata} denotes relevant details such as visual concepts and labels (if required).
    
    \item \textbf{Image Question Answering (\texttt{IMQA})}. This schema could be used for image/visual question answering. It consists of \texttt{(id, question\_id, document\_id, questions, type, choices, context, answer, image\_paths, meta)}, where \texttt{id} denotes a unique row identifier of the dataset, \texttt{question\_id} denotes a unique identifier of the question, \texttt{document\_id} denotes a unique identifier of the context document, \texttt{question} denotes an input question to be answered, \texttt{type} denotes the type of the QA task (e.g., extractive, multiple-choice, open-generative, closed-generative, etc.), \texttt{choices} denotes a list of answer choices (if required), \texttt{context} denotes a passage that serves as the background information of the question (if required), and \texttt{answer} denotes the gold answer to the question (if required), \texttt{image\_path} denotes a list of paths to the input image sources, and \texttt{metadata} denotes relevant details to allow some flexibility of the schema (if required).
    
    \item \textbf{General Video-to-Text (\texttt{VIDEO})}. This schema could be used for video-to-text retrieval and video captioning. It consists of \texttt{(id, video\_path, text, metadata)}, where \texttt{id} denotes a unique row identifier of the dataset, \texttt{video\_path} denotes the file path to an input video source, \texttt{text} denotes the text associated with that particular frame/video, \texttt{metadata} denotes relevant details such as the resolution, duration, and FPS of the video (if required).
\end{compactitem}

\section{Supplementary Details for SEA Evaluation}
\label{app:sea-eval}

\begin{table}[t]
    \centering
    \resizebox{\linewidth}{!}{
    \begin{tabular}{lrrrrr}
      \toprule
      \textbf{Model} & \textbf{$\tau = 0.01$} & \textbf{$\tau = 0.2$} & \textbf{$\tau = 0.5$} & \textbf{$\tau = 0.7$} & \textbf{$\tau = 1.0$} \\
      \toprule
        \textit{Commercial} & & & & & \\
        \hspace{0.4cm}GPT-4 & \underline{0.199} & \underline{0.192} & \underline{0.155} & \underline{0.118} & \textbf{\underline{0.066}} \\
        \hspace{0.4cm}Command-R & 0.201 & 0.198 & 0.185 & 0.168 & 0.126 \\
        \midrule
        \textit{English} & & & & & \\
        \hspace{0.4cm}Mistral & 0.161 & 0.160 & 0.159 & 0.162 & 0.150 \\
        \hspace{0.4cm}Llama3 & \underline{0.138} & \underline{0.137} & \underline{0.131} & \underline{0.129} & \underline{0.113} \\
        \hspace{0.4cm}Falcon & 0.274 & 0.272 & 0.238 & 0.250 & 0.211 \\
        \midrule
        \textit{Multilingual} & & & & & \\
        \hspace{0.4cm}mT0 & 0.151 & 0.148 & 0.131 & 0.112 & 0.074 \\
        \hspace{0.4cm}BLOOMZ & 0.238 & 0.236 & 0.228 & 0.217 & 0.167 \\
        \hspace{0.4cm}BactrianX-Llama & 0.163 & 0.162 & 0.163 & 0.168 & 0.149 \\
        \hspace{0.4cm}AYA-23 & 0.183 & 0.182 & 0.183 & 0.179 & 0.135 \\
        \hspace{0.4cm}AYA-101 & \textbf{\underline{0.112}} & \textbf{\underline{0.109}} & \textbf{\underline{0.095}} & \textbf{\underline{0.085}} & \underline{0.069} \\
        \midrule
        \textit{SEA regional} & & & & & \\
        \hspace{0.4cm}SEA-LION & 0.250 & 0.242 & 0.204 & 0.164 & 0.102 \\
        \hspace{0.4cm}SeaLLM v2.5 & \underline{0.137} & \underline{0.133} & \underline{0.116} & \underline{0.097} & \underline{0.069} \\
        \hspace{0.4cm}Sailor & 0.152 & 0.151 & 0.145 & 0.139 & 0.113 \\
        \midrule
        \textit{SEA country} & & & & & \\
        \hspace{0.4cm}Cendol-mT5 & 0.407 & 0.404 & 0.378 & 0.328 & 0.200 \\
        \hspace{0.4cm}Cendol-Llama2 & 0.294 & 0.290 & 0.267 & 0.232 & 0.149 \\
        \hspace{0.4cm}Merak v4 & 0.209 & 0.207 & 0.199 & 0.190 & 0.155 \\
        \hspace{0.4cm}WangchanX-Llama3 & \underline{0.163} & \underline{0.161} & \underline{0.153} & \underline{0.150} & \underline{0.131} \\
        \hspace{0.4cm}Malaysian Llama3 & 0.181 & 0.181 & 0.179 & 0.176 & 0.143 \\
      \bottomrule
    \end{tabular}
    }
    \caption{Language equity across baselines based on Gini coefficient weighted by population with different $\tau$ values. Lower Gini means higher equity.}
    \label{tab:lang-equity-diff-tau}
\end{table}

\subsection{Datasets}
\label{app:sea-eval-data}

Table~\ref{tab:sea-eval-nlu-senti-subset}, \ref{tab:sea-eval-nlu-nli-subset}, \ref{tab:sea-eval-nlu-topic-subset}, \ref{tab:sea-eval-nlu-commonsense-subset}, and \ref{tab:sea-eval-nlu-mcqa-subset}  provide the details of data subsets used in the NLU evaluation.
Sentiment analysis dataset is originally from NusaX~\cite{winata-etal-2023-nusax}, NusaTranslation~\cite{cahyawijaya-etal-2023-nusawrites}, SentiTaglish\footnote{\url{https://huggingface.co/datasets/ccosme/SentiTaglishProductsAndServices}}, SmSA~\cite{purwarianti2019improving}, PRDECT-ID~\cite{SUTOYO2022108554}, code-mixed Indonesian-English sentiment~\cite{Astuti2023}, Karonese tweet sentiment~\cite{karo2022sentiment}, Typhoon Yolanda sentiment~\cite{imperial2019sentiment}, GKLMIP Khmer sentiment~\cite{gklmip}, Wisesight sentiment corpus\footnote{\url{https://github.com/PyThaiNLP/wisesight-sentiment}}, Filipino-Tagalog product reviews Sentiment\footnote{\url{https://github.com/EricEchemane/Filipino-Tagalog-Product-Reviews-Sentiment-Analysis}}, and multilabel sentiment of Indonesian mobile apps review~\cite{RICCOSAN2023109576}.

Topic classification dataset is originally from NusaParagraph~\cite{cahyawijaya-etal-2023-nusawrites}, UIT-ViON~\cite{tran2021empirical}, SIB-200~\cite{adelani-etal-2024-sib}, GKLMIP Khmer news~\cite{gklmip}, and Indonesian news~\cite{sentrinov}.
Natural Language Inference dataset is originally from IndoNLI~\cite{mahendra-etal-2021-indonli}, WreTe~\cite{setya2018semi}, SNLI Indo~\cite{PUTRA2024109998}, MyXNLI\footnote{\url{https://huggingface.co/datasets/akhtet/myXNLI}}, and XNLI~\cite{conneau-etal-2018-xnli}.
Commonsense reasoning dataset is originally from XStoryCloze~\cite{lin-etal-2022-shot}, IndoCloze~\cite{koto-etal-2022-cloze}, and EMoTES-3K~\cite{catapang-visperas-2023-emotion}.

Open domain QA dataset is originally from IndoMMLU~\cite{koto-etal-2023-large}, SeaEval~\cite{wang2023seaeval}, M3Exam~\cite{M3Exam}, and Okapi~\cite{dac2023okapi}.
Cultural QA dataset is originally from COPAL-ID~\cite{wibowo2023copal}, XCOPA~\cite{ponti-etal-2020-xcopa}, SeaEval~\cite{wang2023seaeval}, and Multilingual Fig-QA~\cite{kabra-etal-2023-multi}.
The reading comprehension dataset is originally from Belebele~\cite{bandarkar2023belebele}.

Table~\ref{tab:sea-eval-nlg-qa-subset}, \ref{tab:sea-eval-nlg-sum-subset}, and \ref{tab:sea-eval-nlg-mt-subset} provide the details of data subsets used in the NLG evaluation. 
The summarization dataset is originally from LR-Sum~\cite{palen-michel-lignos-2023-lr} and XL-Sum~\cite{hasan-etal-2021-xl}.
The machine translation dataset is originally from Lio and the Central Flores corpus~\cite{elias2018lio}, Flores-200~\cite{Costa-jussà2024} and NTREX-128~\cite{federmann-etal-2022-ntrex}.
Question answering dataset is originally from FacQA~\cite{purwarianti2007qa}, QASiNa~\cite{rizqullah2023qasina}, MKQA~\cite{longpre-etal-2021-mkqa}, and Open Thai Wikipedia QA dataset\footnote{\url{https://zenodo.org/records/4539916}}.

Table~\ref{tab:sea-eval-vl-img-cap-subset} and \ref{tab:sea-eval-speech-asr-subset} provide the details of data subsets used in the VL and speech evaluation. 
The image captioning dataset is originally from XM3600~\cite{thapliyal-etal-2022-crossmodal}.
Speech recognition dataset is originally from INDspeech NEWSTRA Ethnic collection~\cite{INDSpeech_ethnic}, ASR Iban~\cite{Juan2015}, FLEURS~\cite{conneau22_interspeech}, and Common Voice~\cite{ardila-etal-2020-common}.

\subsection{Baselines}
\label{app:sea-eval-baselines}

\begin{table}[t]
    \centering
    \resizebox{\linewidth}{!}{
    \begin{tabular}{llc}
      \toprule
      \textbf{Model} & \textbf{Hyperparameter} & \textbf{Value}\\ \midrule
      Logistic Regression & \texttt{max\_iter} & \texttt{100} \\
      & \texttt{C} & \texttt{np.linspace(0.001, 10, 100)}\\ \midrule
      Naive Bayes & \texttt{alpha} & \texttt{np.linspace(0.001, 1, 50)}\\ 
      & \texttt{distribution} & \texttt{MultinomialNB}\\ \midrule
      SVM & \texttt{C} & \texttt{1} \\
      & \texttt{kernel} & \texttt{["rbf", "linear"]}
      \\
      \bottomrule
    \end{tabular}
    }
    \caption{Hyper-parameters of classical models for Translationese prediction through grid search.}
    \label{tab:hyper-params-translationese}
\end{table}

\begin{table}[t]
    \centering
    \resizebox{\linewidth}{!}{
    \begin{tabular}{l|c|ccc}
      \toprule
      \textbf{Model} & \textbf{3-label} & \textbf{HT vs. MT-Nat} & \textbf{MT vs. HT-Nat} & \textbf{Nat vs. HT-MT} \\ \midrule
      LR (TF-IDF) & 39.73 & 53.03 & 56.01 & 75.20 \\ 
      LR (BoW) & 45.63 &  55.90 & 61.39 & 75.60  \\ 
      NB (TF-IDF) & 33.43 & 49.53 & 50.55 & 73.05 \\ 
      NB (BoW) & 33.70 & 49.10 & 50.64 & 71.26 \\ 
      SVM (TF-IDF) & 39.55 & 52.63 & 55.10 & 76.40 \\ 
      SVM (BoW) & 46.84 & 56.85 & \underline{61.40} & 75.65 \\
      \midrule
      mDeBERTa & \underline{51.51} &	\underline{64.77} & 	59.16 & \textbf{79.08} \\
      \bottomrule
    \end{tabular}
    }
    \caption{Results of translationese classifier (accuracy) averaged across languages.}
    \label{tab:translationese-results-avg}
\end{table}

Table~\ref{tab:sea-eval-baselines-nlp}, \ref{tab:sea-eval-baselines-speech}, and \ref{tab:sea-eval-baselines-vl} report the details of baseline models used in SEACrowd evaluation (\S\ref{sec:sea-evaluation}). For each baseline model, we provide information regarding the model size, origin base model, seen languages in the training corpora use, and the URL where the models can be downloaded. In principle, this work does not aim to acquire and fit all available SEA-trained LLMs over the Internet, as this is computationally expensive. Rather, we want to initiate the exploration of select publicly available models to serve as baselines for the evaluation of foundational capabilities on SEA languages through benchmarking on NLU, NLG, speech, and vision tasks aggregated via SEACrowd.

Across the various models explored, as listed in the tables, we prioritized the diversity of model variation in terms of scale, openness, and coverage of SEA languages. In NLP tasks, we covered 5 LLM groups for the main experiments: English-only, multilingual, regional, and country-specific models. Instruction-tuned LLMs demonstrate the ability to generalize to unseen tasks~\cite{wei2021finetuned, sanh2021multitask, ouyang2022training}. Some of these LLMs are based on a multilingual foundation, hence their proficiency in generalizing across languages~\cite{muennighoff2022crosslingual,adilazuarda2023the,zhang-etal-2023-multilingual}. For NLU, we compute the weighted F1-score and obtain the answers via log-likelihood for open-source baselines or string matching for commercial baselines.

For the speech benchmark, only two model families are available: multilingual models and models fine-tuned on specific SEA languages. For vision tasks, we covered English-only and one multilingual model. These models utilize a visual backbone pre-trained on image-text alignment, e.g., CLIP~\cite{pmlr-v139-radford21a}, to project image features into the input space of an existing pre-trained LM. In summary, we mostly explored open models readily accessible on HuggingFace but also included commercial models such as GPT-4 and Whisper V3 for performance benchmarking, reproducibility, and extension by future works.

\begin{table}[t]
    \centering
    \begin{subtable}[t]{\linewidth}
        \centering
        \resizebox{0.6\linewidth}{!}{
        \begin{tabular}{lcc}
          \toprule
          \textbf{Country} & \textbf{Affiliation} & \textbf{Origin} \\
          \toprule
            Indonesia & 16 & 31 \\
            Malaysia & 0 & 1 \\
            Philippines & 3 & 7 \\
            Singapore & 13 & 2 \\
            Thailand & 1 & 2 \\
            Vietnam & 0 & 1 \\
            \midrule
            Australia & 1 & 0 \\
            Brazil/Sweden & 0 & 1 \\
            Canada & 1 & 0 \\
            China & 2 & 8 \\
            Egypt & 0 & 1 \\
            Germany & 0 & 2 \\
            Hong Kong & 2 & 0 \\
            India & 0 & 1 \\
            Ireland & 1 & 0 \\
            Japan & 3 & 0 \\
            The Netherlands & 0 & 1 \\
            UAE & 5 & 0 \\
            UK & 4 & 0 \\
            USA & 9 & 1 \\
            Uzbekistan & 0 & 2 \\
        \bottomrule
        \end{tabular}
        }
    \end{subtable}
    \caption{The demographics of the authors based on affiliation country and origin country.}
    \label{tab:seacrowd-author-demographic}
\end{table}

\subsection{Prompts}
\label{app:sea-eval-prompts}

Tables~\ref{tab:sea-eval-prompts-nlu}, \ref{tab:sea-eval-prompts-nlg}, and \ref{tab:sea-eval-prompts-vl} describe the handwritten prompt templates used in NLU, NLG, and VL evaluation (\S\ref{sec:sea-evaluation}). For all tasks, we used a zero-shot prompting procedure to serve as the baseline setup. Due to the task complexity and distribution of workload from volunteer contributors with available computing resources, we limited the experiment procedure for some setups to ensure the acquisition of results in line with target release dates. For NLU, we explored three prompt styles for each dataset from core tasks, including commonsense reasoning, question-answering, and NLI. For more challenging tasks requiring more intensive computing power such as NLG and VL, we used only one uniform prompt style, but we also explored prompts translated into SEA languages, i.e., Filipino, Indonesian, Thai, and Vietnamese for VL.

\subsection{Evaluation Results}
\label{app:sea-eval-results}

Table~\ref{tab:sea-eval-nlu-results-per-lang} and \ref{tab:sea-eval-nlg-results-per-lang} describes the NLU and NLG results per language.

\subsection{Language Equity Results}
\label{app:sea-eval-lang-equity}

Table~\ref{tab:lang-equity-diff-tau} presents the language equity of LLMs used in the evaluation across different weights of the number of language speakers in the Gini coefficient calculation.

\section{Supplementary Details for Translationese Classifier}
\label{app:translationese-natural-classifier}

\subsection{Training \& Evaluation Data}
\label{app:translationese-natural-classifier-data}

We manually select and validate the text collection method of each data subset for training and evaluating the translationese classifier, in Tables~\ref{tab:translationese-vs-natural-data-train} and \ref{tab:translationese-vs-natural-data-test}, respectively. This validation is done by checking the relevant publication, domain, and annotation method. If the texts in the data subsets are a product of machine or human translation, we regard them as translationese. We label data subsets with human-generated texts as natural data.

\subsection{Experiments}
\label{app:translationese-natural-classifier-exp}

We aim to assess the capability of ML models to differentiate between human-generated/natural samples (\texttt{Nat}), human-translated samples (\texttt{HT}), and machine-translated samples (\texttt{MT}). Our approach involves training classifiers using classical ML techniques and fine-tuning mDeBERTa models to enhance learning. Furthermore, we experiment by combining two label classes into one to evaluate the predictive difficulty of distinguishing between these labels. This analysis provides valuable insights into the relative similarity of the samples across these categories. The following section provides a comprehensive overview of our methodology for this study.

\paragraph{Classical ML}
We use three classical machine learning methods: 1) Logistic Regression (LR), 2) Naive Bayes (NB), and 3) Support Vector Machine (SVM) with two different features, including TF-IDF and Bag-of-words (BoW). We run hyper-parameter tuning with grid search to find the best hyper-parameters for each method on validation set, and report the results on test set in Table~\ref{tab:hyper-params-translationese}.

\paragraph{Encoder LM}
We explore fine-tuning encoder-only LM for developing a translationese classifier. We utilize mDeBERTa-v3$_{base}$ model\footnote{\url{https://huggingface.co/microsoft/mdeberta-v3-base}}~\cite{he2020deberta,he2022debertav3}--a multilingual encoder-only LM--as our backbone. We train the model with AdamW~\cite{loshchilov2018decoupled} optimizer using a learning rate of 1e-5, batch size of 256, and warming up steps of 500 for a maximum of 10 epochs. We apply an early stopping of 3 epochs based on the validation accuracy. We show the results in Table \ref{tab:translationese-results-avg}.

\begin{table}[t!]
    \centering
    \resizebox{0.75\linewidth}{!}{
    \begin{tabular}{clr}
        \toprule
        \textbf{No.} & \textbf{Name} & \textbf{C. Points} \\
        \toprule
        1 & Holy Lovenia & 549 \\
        2 & Samuel Cahyawijaya & 480 \\
        3 & Rahmad Mahendra & 317 \\
        4 & Salsabil Maulana Akbar & 243 \\
        5 & Lester James V. Miranda & 234 \\
        6 & Zheng-Xin Yong & 164 \\
        7 & Jennifer Santoso & 164 \\
        8 & Elyanah Aco & 158 \\
        9 & Akhdan Fadhilah & 157 \\
        10 & Jonibek Mansurov & 132 \\
        11 & Fajri Koto & 121 \\
        12 & Joseph Marvin Imperial & 118 \\
        13 & Ruochen Zhang & 114 \\
        14 & Genta Indra Winata & 108 \\
        15 & Onno P. Kampman & 107 \\
        16 & Joel Ruben Antony Moniz & 93 \\
        17 & Muhammad Ravi Shulthan Habibi & 92 \\
        18 & Frederikus Hudi & 83 \\
        19 & Sedrick Keh & 81 \\
        20 & Alham Fikri Aji & 80 \\
        21 & Railey Montalan & 78 \\
        22 & Peerat Limkonchotiwat & 72 \\
        \midrule
        23 & Ryan Ignatius & 56 \\
        24 & Joanito Agili Lopo & 50 \\
        25 & William Nixon & 50 \\
        26 & Börje F. Karlsson & 49 \\
        27 & James Jaya & 48 \\
        28 & Ryandito Diandaru & 48 \\
        29 & Yuze Gao & 48 \\
        30 & William Tjhi & 46 \\
        31 & Patrick Amadeus & 46 \\
        32 & Bin Wang & 44 \\
        33 & Jan Christian Blaise Cruz & 43 \\
        34 & Chenxi Whitehouse & 36 \\
        35 & Ivan Halim Parmonangan & 36 \\
        36 & Maria Khelli & 36 \\
        37 & Sebastian Ruder & 35 \\
        38 & Wenyu Zhang & 34 \\
        39 & Lucky Susanto & 33 \\
        40 & Reynard Adha Ryanda & 32 \\
        41 & Sonny Lazuardi Hermawan & 30 \\
        42 & Dan John Velasco & 29 \\
        43 & Muhammad Dehan Al Kautsar & 29 \\
        44 & Willy Fitra Hendria & 29 \\
        45 & Yasmin Moslem & 29 \\
        46 & Noah Flynn & 28 \\
        47 & Muhammad Farid Adilazuarda & 27 \\
        48 & Haochen Li & 27 \\
        49 & Johanes Lee & 27 \\
        50 & R. Damanhuri & 27 \\
        51 & Shuo Sun & 27 \\
        52 & Muhammad Reza Qorib & 26 \\
        53 & Amirbek Djanibekov & 25 \\
        54 & Wei Qi Leong & 25 \\
        55 & Quyet V. Do & 24 \\
        56 & Niklas Muennighoff & 24 \\
        57 & Tanrada Pansuwan & 22 \\
        58 & Ilham Firdausi Putra & 21 \\
        59 & Yan Xu & 21 \\
        60 & Ayu Purwarianti & 20 \\
        61 & Ngee Chia Tai & 20 \\
        \bottomrule
    \end{tabular}
    }
    \caption{Co-authors ordered by their amount of contribution points.}
    \label{tab:coauthors-by-contributions}
\end{table}

\section{Supplementary Details for SEA Language Prioritization}
\label{app:lang-prioritization}

\begin{figure*}[t]
  \centering
  \begin{subfigure}[t]{\linewidth}
    \centering
      \includegraphics[trim={0.4cm, 1.1cm, 0.4cm, 0.3cm}, clip, width=0.9\linewidth]{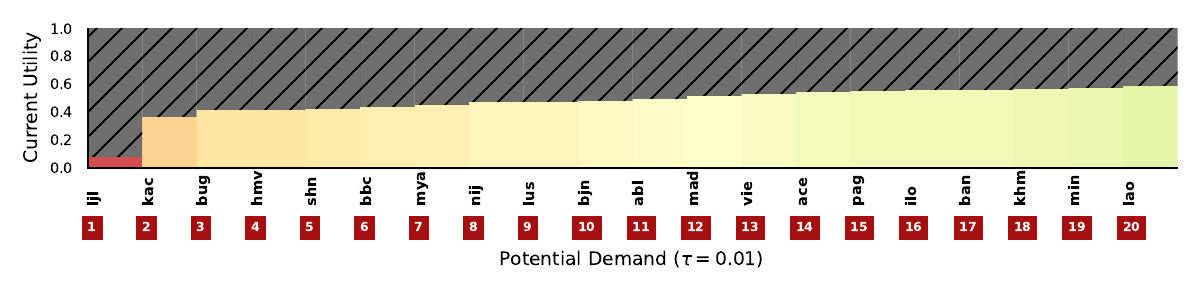}
      \caption{$\tau = 0.01$}
  \end{subfigure}
  \begin{subfigure}[t]{\linewidth}
    \centering
      \includegraphics[trim={0.4cm, 1.1cm, 0.4cm, 0.3cm}, clip, width=0.9\linewidth]{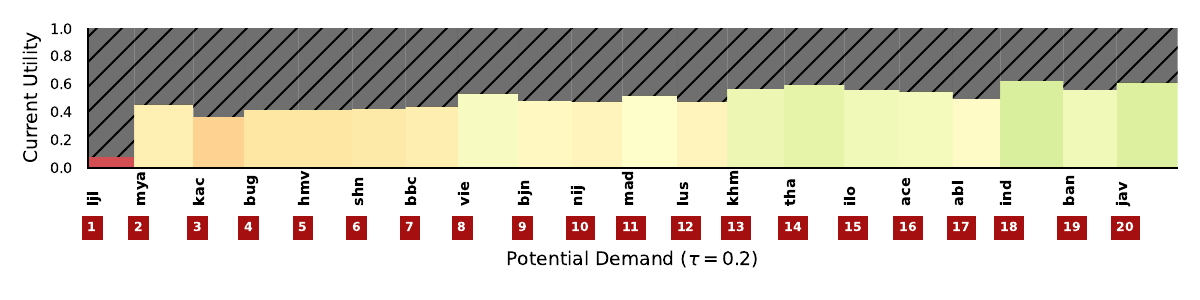}
      \caption{$\tau = 0.2$}
  \end{subfigure}
  \begin{subfigure}[t]{\linewidth}
    \centering
      \includegraphics[trim={0.4cm, 1.1cm, 0.4cm, 0.3cm}, clip, width=0.9\linewidth]{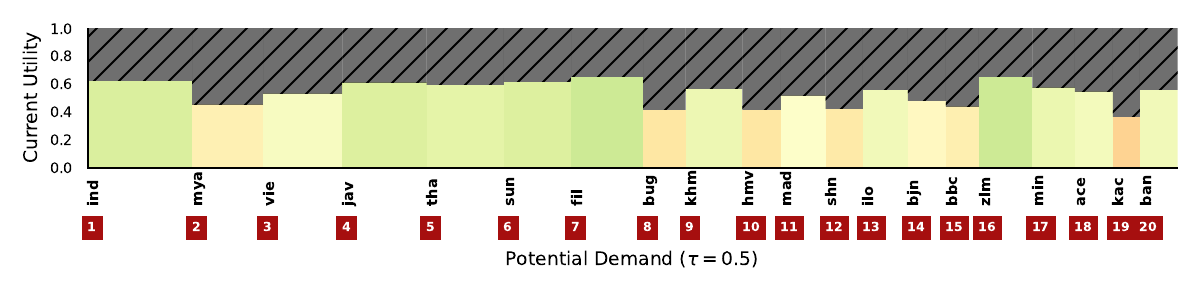}
      \caption{$\tau = 0.5$}
  \end{subfigure}
  \begin{subfigure}[t]{\linewidth}
    \centering
      \includegraphics[trim={0.4cm, 1.1cm, 0.4cm, 0.3cm}, clip, width=0.9\linewidth]{figures/next-steps/averaged_0_7.pdf}
      \caption{$\tau = 0.7$}
  \end{subfigure}
  \begin{subfigure}[t]{\linewidth}
    \centering
      \includegraphics[trim={0.4cm, 1.1cm, 0.4cm, 0.3cm}, clip, width=0.9\linewidth]{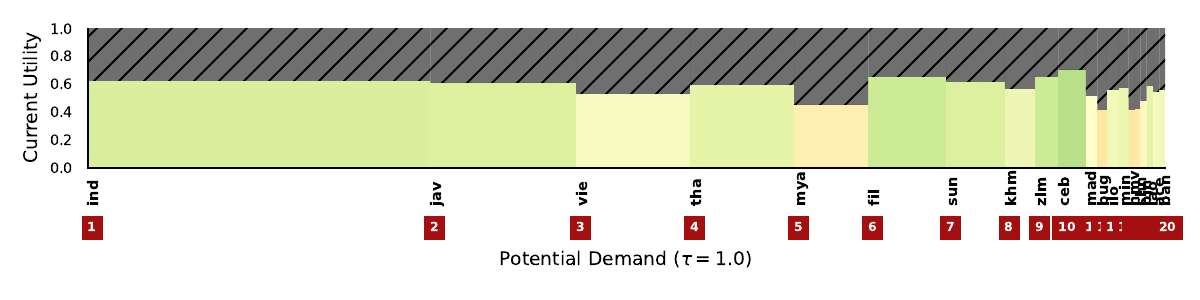}
      \caption{$\tau = 1.0$}
  \end{subfigure}
  \caption{Top-20 SEA indigenous languages to be prioritized based on their potential demand and current utility.}
  \label{fig:lang-prioritization-capability}
\end{figure*}

\begin{figure*}[t]
  \centering
  \begin{subfigure}[t]{\linewidth}
    \centering
      \includegraphics[trim={0.4cm, 1.1cm, 0.4cm, 0.3cm}, clip, width=0.9\linewidth]{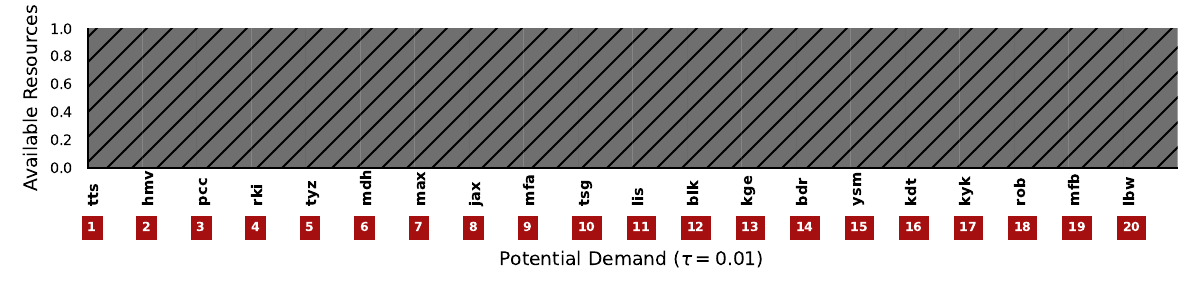}
      \caption{$\tau = 0.01$}
  \end{subfigure}
  \begin{subfigure}[t]{\linewidth}
    \centering
      \includegraphics[trim={0.4cm, 1.1cm, 0.4cm, 0.3cm}, clip, width=0.9\linewidth]{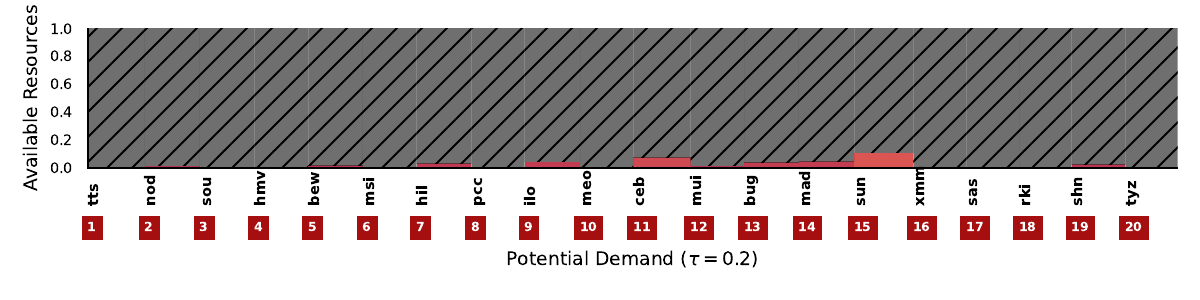}
      \caption{$\tau = 0.2$}
  \end{subfigure}
  \begin{subfigure}[t]{\linewidth}
    \centering
      \includegraphics[trim={0.4cm, 1.1cm, 0.4cm, 0.3cm}, clip, width=0.9\linewidth]{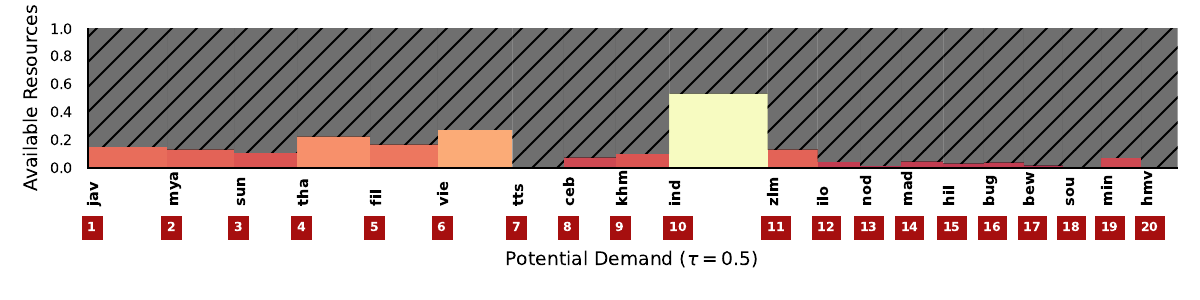}
      \caption{$\tau = 0.5$}
  \end{subfigure}
  \begin{subfigure}[t]{\linewidth}
    \centering
      \includegraphics[trim={0.4cm, 1.1cm, 0.4cm, 0.3cm}, clip, width=0.9\linewidth]{figures/next-steps/data_0_7.pdf}
      \caption{$\tau = 0.7$}
  \end{subfigure}
  \begin{subfigure}[t]{\linewidth}
    \centering
      \includegraphics[trim={0.4cm, 1.1cm, 0.4cm, 0.3cm}, clip, width=0.9\linewidth]{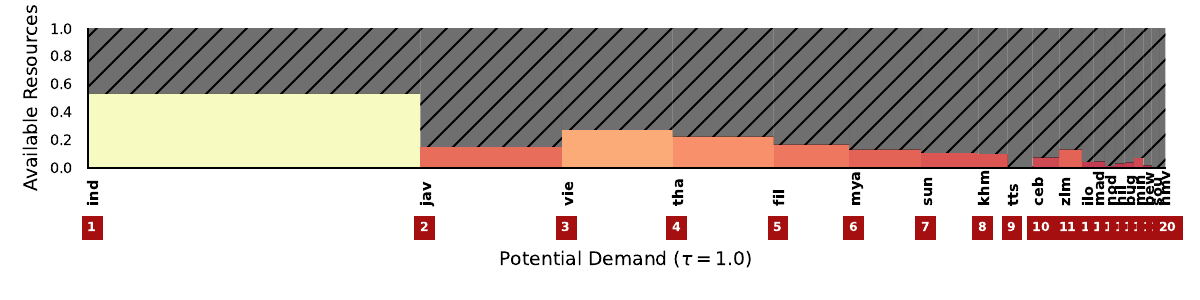}
      \caption{$\tau = 1.0$}
  \end{subfigure}
  \caption{Top-20 SEA indigenous languages to be prioritized based on their potential demand and data availability.}
  \label{fig:lang-prioritization-availability}
\end{figure*}

Based on the results of the global utility metric~\cite{blasi-etal-2022-systematic}, we provide the top-20 SEA indigenous languages to be prioritized based on their demand (i.e., the number of SEA language speakers) and current utility (Figure~\ref{fig:lang-prioritization-capability}) or resource availability (Figure~\ref{fig:lang-prioritization-availability}).\footnote{\url{https://github.com/SEACrowd/globalutility}} We use the performance scores of AYA-101 as one of the best-performing models on SEA languages for the current utility. While the current utility, also known as the model capability, is relative to the model performance on \textsc{eng}, the resource availability is relative to 500, which is approximately the number of datasets in Korean language available in HuggingFace. The Korean language is chosen as the pivot because it is considered a higher-resource language than most by~\citet{joshi-etal-2020-state}.

\section{Contributor Demographics}
\label{app:contributor-demographic}

Table~\ref{tab:seacrowd-author-demographic} describes the geographical distribution of the authors in SEACrowd.

\section{Languages Under Study}
\label{app:lang-under-study}

Table~\ref{tab:sea-langs-under-100m-speakers}-\ref{tab:sea-langs-under-100-speakers} present the list of SEA indigenous languages covered by SEACrowd. Information regarding the ISO 639-3 code, language name, region, and population is obtained from~\cite{ethnologue, glottolog, thejoshuaproject, wals} and Wikipedia\footnote{\url{https://www.wikipedia.org/}}.

\section{Amount of Contributions by Co-Authors}

Table~\ref{tab:coauthors-by-contributions} provides a list of co-authors sorted by their amount of contributions in SEACrowd. The full details of their contributions can be seen in \href{https://docs.google.com/spreadsheets/d/e/2PACX-1vQDZtJjA6i7JsxS5IlMtVuwOYjr2Pbl_b47yMSH4aAdHDBIpf-CiJQjNQAzcJPEu_aE7kwH4ZvKvPm0/pubhtml?gid=225616890&single=true}{our contribution tracking}.


\begin{table*}[t]
    \centering
    \begin{subtable}[t]{\linewidth}
        \centering
        \resizebox{\linewidth}{!}{
\begin{tabular}{lcclp{.6\linewidth}}
          \toprule
          \textbf{Model name} & \textbf{Model size} & \textbf{Backbone} & \textbf{Seen langs} & \textbf{URL} \\
          \toprule
            \multicolumn{5}{c}{\cellcolor[HTML]{D9D9D9}\textit{\textbf{Commercial}}} \\
             GPT-4 & N/A & GPT-4 & N/A & \url{https://openai.com/index/gpt-4/}. We used \texttt{turbo-2024-04-09} for NLU and \texttt{gpt-4o-2024-05-13} for NLG.\\
             Command-R & 36B & Command-R & 2 SEA langs (\textsc{vie}, \textsc{ind}),  22 non-SEA langs & \url{https://cohere.com/blog/command-r}\\
             \multicolumn{5}{c}{\cellcolor[HTML]{D9D9D9}\textit{\textbf{English}}} \\
             Mistral & 7B & Mistral & N/A & \texttt{mistralai/Mistral-7B-Instruct-v0.3}\\
             Llama3 & 8B & Llama3 & N/A & \texttt{meta-llama/Meta-Llama-3-8B-Instruct}\\
             Falcon & 7B & Falcon & 0 SEA langs (mainly English) & \texttt{tiiuae/falcon-7b-instruct}\\
\bottomrule
        \end{tabular}
        }
    \end{subtable}
    \caption{LLMs used in SEACrowd NLU and NLG evaluation.}
    \label{tab:sea-eval-baselines-nlp}
\end{table*}

\begin{table*}[t]
  \centering
  \begin{subtable}[t]{\linewidth}
      \centering
      \resizebox{\linewidth}{!}{
      \begin{tabular}{lcclp{.6\linewidth}}  
        \toprule
        \textbf{Model name} & \textbf{Model size} & \textbf{Backbone} & \textbf{Seen langs} & \textbf{URL} \\
        \toprule

          \multicolumn{5}{c}{\cellcolor[HTML]{D9D9D9}\textit{\textbf{Multilingual}}} \\
          Whisper v3 & 1.54B & Whisper v3 & 89 non-SEA \& 9 SEA (\textsc{ind}, \textsc{jav}, \textsc{lao}, \textsc{zlm}, \textsc{mya}, \textsc{tgl}, \textsc{tha}, \textsc{sun}, \textsc{vie}) & \texttt{openai/whisper-large-v3}  \\
\bottomrule
      \end{tabular}
      }
  \end{subtable}
  \caption{Speech models used in SEACrowd speech evaluation.}
  \label{tab:sea-eval-baselines-speech}
\end{table*}

\begin{table*}[t]
    \centering
    \begin{subtable}[t]{\linewidth}
        \centering
        \resizebox{0.8\linewidth}{!}{
        \begin{tabular}{lcccp{.4\linewidth}}
          \toprule
          \textbf{Model name} & \textbf{Model size} & \textbf{Backbone} & \textbf{Pre-training images} & \textbf{URL} \\
          \toprule
             \multicolumn{5}{c}{\cellcolor[HTML]{D9D9D9}\textit{\textbf{English}}} \\
             LLaVA 1.5 & N/A & N/A & N/A & N/A \\
             LLaVA 1.6 & 7B & Mistral-7B & N/A & \texttt{liuhaotian/llava-v1.6-mistral-7b} \\
             Idefics2 & 8B & Mistral-7B-v0.1 & 1.5B & \texttt{HuggingFaceM4/idefics2-8b} \\
             PaliGemma & 2B & Gemma-2B & N/A & \texttt{google/paligemma-3b-pt-224} \\
             
             \multicolumn{5}{c}{\cellcolor[HTML]{D9D9D9}\textit{\textbf{Multilingual}}} \\
             mBLIP & N/A & blip2-flan-t5-xl & N/A & \texttt{Gregor/mblip-mt0-xl} \\

        \bottomrule
        \end{tabular}
        }
    \end{subtable}
    \caption{VLMs used in SEACrowd VL evaluation.}
    \label{tab:sea-eval-baselines-vl}
\end{table*}

\begin{table*}[t]
    \centering
    \begin{subtable}[t]{\linewidth}
        \centering
        \resizebox{\linewidth}{!}{
        \begin{tabular}{cp{1.3\linewidth}}
          \toprule
          \textbf{No.} & \textbf{Prompt template} \\
          \toprule
             \multicolumn{2}{c}{\cellcolor[HTML]{D9D9D9}\textit{\textbf{Sentiment Analysis}}} \\
             1 & \texttt{Classify the sentiment of the text below.\symbol{92}n[INPUT] => Sentiment ([OPTIONS]): [LABEL\_CHOICE]} \\
             2 & \texttt{Predict the sentiment of the following text.\symbol{92}nText: [INPUT]\symbol{92}nAnswer with [OPTIONS]: [LABEL\_CHOICE]} \\
             3 & \texttt{[INPUT]\symbol{92}nWhat would be the sentiment of the text above? [OPTIONS]? [LABEL\_CHOICE]} \\
             \midrule
             \multicolumn{2}{c}{\cellcolor[HTML]{D9D9D9}\textit{\textbf{Topic Classification}}} \\
             1 & \texttt{Classify the topic of the text below.\symbol{92}n[INPUT] => Topic ([OPTIONS]): [LABEL\_CHOICE]} \\
             2 & \texttt{Predict the topic of the following text.\symbol{92}nText: [INPUT]\symbol{92}nAnswer with [OPTIONS]: [LABEL\_CHOICE]} \\
\bottomrule
        \end{tabular}
        }
    \end{subtable}
    \caption{Prompt templates used for NLU tasks.}
    \label{tab:sea-eval-prompts-nlu}
\end{table*}

\begin{table*}[t]
    \centering
    \begin{subtable}[t]{\linewidth}
        \centering
        \resizebox{\linewidth}{!}{
        \begin{tabular}{cp{1.3\linewidth}}
          \toprule
          \textbf{No.} & \textbf{Prompt template} \\
          \toprule
             \multicolumn{2}{c}{\cellcolor[HTML]{D9D9D9}\textit{\textbf{Machine Translation (MT)}}} \\
             1 & \texttt{Translate the following text from [SOURCE] to [TARGET]. Give your translation directly.\symbol{92}nText: [INPUT]\symbol{92}nTranslation:} \\
            \midrule
            \multicolumn{2}{c}{\cellcolor[HTML]{D9D9D9}\textit{\textbf{Summarization}}} \\
            1 & \texttt{Write a summary from the following text.\symbol{92}nText: [INPUT]\symbol{92}nSummary:} \\
            \midrule
            \multicolumn{2}{c}{\cellcolor[HTML]{D9D9D9}\textit{\textbf{Abstractive \& Extractive QA}}} \\
            1 & \texttt{Refer to the passage below and answer the following question:\symbol{92}nPassage: [CONTEXT]\symbol{92}nQuestion: [QUESTION]\symbol{92}nAnswer:} \\
        \bottomrule
        \end{tabular}
        }
    \end{subtable}
    \caption{Prompt templates used for NLG tasks.}
    \label{tab:sea-eval-prompts-nlg}
\end{table*}

\begin{table*}[t]
    \centering
    \begin{subtable}[t]{\linewidth}
        \centering
        \resizebox{\linewidth}{!}{
        \begin{tabular}{cp{1.3\linewidth}}
          \toprule
          \textbf{Lang.} & \textbf{Prompt template} \\
          \toprule
             \multicolumn{2}{c}{\cellcolor[HTML]{D9D9D9}\textit{\textbf{Image Captioning}}} \\
             \textsc{eng} & \texttt{Caption the following image in [LANGUAGE].} \\
             \textsc{fil} & \texttt{Ilarawan ang sumusunod na larawan.} \\
             \textsc{ind} & \texttt{Deskripsikan gambar berikut.} \\
        \bottomrule
        \end{tabular}
        }
    \end{subtable}
    \caption{Prompt templates used for the image captioning task in VL evaluation.}
    \label{tab:sea-eval-prompts-vl}
\end{table*}

\begin{sidewaystable*}[t]
    \centering
    \begin{subtable}[t]{\linewidth}
        \centering
        \resizebox{0.9\linewidth}{!}{
        \hspace{-6.em}
        \begin{tabular}{lrrrrrrrrrrrrrrrrrrrrrrrrrrrrrrrrrrrr}
          \toprule
         & \textbf{\textsc{abl}} & \textbf{\textsc{abs}} & \textbf{\textsc{ace}} & \textbf{\textsc{ban}} & \textbf{\textsc{bbc}} & \textbf{\textsc{bew}} & \textbf{\textsc{bhp}} & \textbf{\textsc{bjn}} & \textbf{\textsc{btx}} & \textbf{\textsc{bug}} & \textbf{\textsc{ceb}} & \textbf{\textsc{eng}} & \textbf{\textsc{fil}} & \textbf{\textsc{ilo}} & \textbf{\textsc{ind}} & \textbf{\textsc{jav}} & \textbf{\textsc{kac}} & \textbf{\textsc{khm}} & \textbf{\textsc{lao}} & \textbf{\textsc{lus}} & \textbf{\textsc{mad}} & \textbf{\textsc{mak}} & \textbf{\textsc{min}} & \textbf{\textsc{mui}} & \textbf{\textsc{mya}} & \textbf{\textsc{nij}} & \textbf{\textsc{pag}} & \textbf{\textsc{rej}} & \textbf{\textsc{shn}} & \textbf{\textsc{sun}} & \textbf{\textsc{tha}} & \textbf{\textsc{vie}} & \textbf{\textsc{war}} & \textbf{\textsc{zsm}} & \textbf{Overall} \\
        \toprule
        GPT-4 & 63.3 & 39.0 & 39.3 & 60.3 & 7.1 & 68.5 & 2.8 & 60.4 & 27.8 & 40.4 & 85.6 & 52.1 & 55.9 & 69.5 & 60.7 & 59.7 & 30.8 & 66.4 & 51.8 & 70.0 & 37.1 & 44.3 & 57.9 & 71.8 & 47.6 & 40.2 & 79.4 & 34.0 & 21.7 & 58.5 & 59.6 & 56.1 & 84.9 & 61.6 & 51.9 \\
        Command-R & 50.1 & 80.8 & 57.6 & 62.8 & 47.4 & 81.8 & 58.2 & 57.1 & 57.3 & 57.9 & 66.7 & 69.4 & 51.1 & 56.8 & 58.3 & 61.2 & 36.5 & 41.5 & 33.8 & 63.9 & 61.9 & 58.4 & 66.4 & 81.7 & 34.8 & 53.3 & 75.6 & 69.6 & 35.4 & 63.2 & 42.7 & 55.9 & 67.6 & 55.7 & 58.0 \\
        \midrule
        Mistral & 36.7 & 53.6 & 46.4 & 49.6 & 33.0 & 59.3 & 44.3 & 44.6 & 44.3 & 48.8 & 53.5 & 69.2 & 48.4 & 49.1 & 52.5 & 46.7 & 33.2 & 29.8 & 30.7 & 56.1 & 45.7 & 44.8 & 51.2 & 62.6 & 27.4 & 40.1 & 69.2 & 48.6 & 31.9 & 48.3 & 40.8 & 45.2 & 54.4 & 49.6 & 46.8 \\
        Llama3 & 37.3 & 40.3 & 43.2 & 48.9 & 34.8 & 44.5 & 32.6 & 42.2 & 38.5 & 42.9 & 51.2 & 59.5 & 45.2 & 46.7 & 49.2 & 44.4 & 28.5 & 34.6 & 30.3 & 46.8 & 39.0 & 38.0 & 43.6 & 49.2 & 35.2 & 39.6 & 60.5 & 38.5 & 31.1 & 45.2 & 43.8 & 45.5 & 50.3 & 49.0 & 42.6 \\
\bottomrule
        \end{tabular}
        }
    \end{subtable}
    \caption{NLU evaluation results in weighted F1-score per language.}
    \label{tab:sea-eval-nlu-results-per-lang}
\end{sidewaystable*}

\begin{sidewaystable*}
    \centering
    \begin{subtable}[t]{\linewidth}
        \centering
        \resizebox{\linewidth}{!}{
        \begin{tabular}{lrrrrrrrrrrrrrrrrrrrrrrrrrrrrr}
        \toprule
        & \textbf{\textsc{ace}} & \textbf{\textsc{ban}} & \textbf{\textsc{bbc}} & \textbf{\textsc{bjn}} & \textbf{\textsc{bug}} & \textbf{\textsc{ceb}} & \textbf{\textsc{fil}} & \textbf{\textsc{hmv}} & \textbf{\textsc{ilo}} & \textbf{\textsc{ind}} & \textbf{\textsc{jav}} & \textbf{\textsc{kac}} & \textbf{\textsc{khm}} & \textbf{\textsc{lao}} & \textbf{\textsc{ljl}} & \textbf{\textsc{lus}} & \textbf{\textsc{mad}} & \textbf{\textsc{min}} & \textbf{\textsc{mya}} & \textbf{\textsc{nij}} & \textbf{\textsc{pag}} & \textbf{\textsc{shn}} & \textbf{\textsc{sun}} & \textbf{\textsc{tha}} & \textbf{\textsc{vie}} & \textbf{\textsc{war}} & \textbf{\textsc{zsm}} & \textbf{Overall} \\
        \toprule
        GPT-4 & 32.9 & 40.7 & 28.8 & 42.0 & 24.1 & 66.5 & 65.6 & 50.0 & 52.9 & 59.3 & 54.2 & 16.7 & 29.8 & 41.9 & 10.0 & 33.3 & 29.2 & 46.1 & 21.5 & 27.7 & 37.0 & 14.5 & 50.0 & 28.8 & 47.5 & 66.4 & 59.6 & 39.9 \\
        Command-R & 19.6 & 26.1 & 16.4 & 30.0 & 16.0 & 44.3 & 52.5 & 16.8 & 29.4 & 57.9 & 32.6 & 8.8 & 8.7 & 14.2 & 6.0 & 19.5 & 17.2 & 31.6 & 9.5 & 18.4 & 20.4 & 8.9 & 27.5 & 24.3 & 46.8 & 34.4 & 50.1 & 25.5 \\
        \midrule
        Mistral & 12.4 & 15.0 & 10.0 & 13.9 & 11.1 & 28.5 & 37.2 & 10.2 & 15.9 & 28.6 & 15.4 & 7.3 & 8.7 & 10.8 & 4.2 & 11.7 & 9.5 & 18.0 & 5.7 & 12.4 & 17.5 & 9.5 & 14.8 & 15.1 & 25.1 & 22.4 & 31.1 & 15.6 \\
        Llama3 & 11.0 & 12.3 & 8.1 & 13.8 & 7.6 & 25.1 & 33.2 & 7.6 & 18.4 & 21.9 & 17.0 & 4.8 & 6.5 & 5.8 & 3.2 & 9.6 & 8.5 & 16.4 & 4.5 & 9.5 & 11.8 & 6.3 & 15.1 & 9.6 & 21.7 & 20.5 & 25.2 & 13.2 \\
        Falcon & 7.3 & 9.5 & 8.2 & 8.3 & 7.9 & 18.6 & 23.6 & 6.6 & 9.7 & 15.3 & 7.7 & 6.0 & 3.1 & 3.1 & 4.2 & 9.3 & 6.6 & 11.8 & 1.8 & 8.7 & 12.9 & 4.5 & 7.7 & 2.4 & 13.5 & 13.5 & 17.0 & 9.2 \\
        \midrule
        mT0 & 4.8 & 5.6 & 3.7 & 5.7 & 3.1 & 4.6 & 6.8 & 4.5 & 3.8 & 29.3 & 5.8 & 2.1 & 4.3 & 6.1 & 1.7 & 3.4 & 3.6 & 6.5 & 5.0 & 3.5 & 3.6 & 3.5 & 6.8 & 9.4 & 19.6 & 6.1 & 9.1 & 6.4 \\
\bottomrule
        \end{tabular}
        }
    \end{subtable}
    \caption{NLG evaluation results in ROUGE-L per language.}
    \label{tab:sea-eval-nlg-results-per-lang}
\end{sidewaystable*}

\begin{table*}[t]
    \centering
    \begin{subtable}[t]{\linewidth}
        \centering
        \resizebox{\linewidth}{!}{
        \begin{tabular}{clllr}
          \toprule
          \textbf{Lang.} & \textbf{Subset} & \textbf{Original Task} & \textbf{Domain} & \textbf{\# Samples} \\
          \toprule
            \multicolumn{5}{c}{\cellcolor[HTML]{D9D9D9}\textit{\textbf{Translationese}}} \\
            \textsc{eng} & \texttt{emotes\_3k\_eng\_seacrowd\_t2t} & Commonsense Reasoning & Ethics & 2000 \\
            \textsc{eng} & \texttt{aya\_evaluation\_suite\_eng\_seacrowd\_t2t} & Instruction Tuning & General & 400 \\
            \midrule
            \textsc{ind} & \texttt{belebele\_ind\_latn\_seacrowd\_qa} & QA & General & 1969 \\
            \textsc{ind} & \texttt{parallel\_asian\_treebank\_ind\_eng\_seacrowd\_t2t} & Machine Translation & News & 31 \\
            \textsc{ind} & \texttt{aya\_evaluation\_suite\_ind\_seacrowd\_t2t} & Instruction Tuning & General & 4 \\
            \textsc{ind} & \texttt{bactrian\_x\_id\_seacrowd\_t2t} & Instruction Tuning & Mixed, Multi-domain, Wikipedia & 1972 \\
\bottomrule
        \end{tabular}
        }
    \end{subtable}
    \caption{Train data used in the translationese classifier experiment.}
    \label{tab:translationese-vs-natural-data-train}
\end{table*}

\begin{table*}[t]
    \centering
    \begin{subtable}[t]{\linewidth}
        \centering
        \resizebox{\linewidth}{!}{
        \begin{tabular}{clllr}
          \toprule
          \textbf{Lang.} & \textbf{Subset} & \textbf{Original Task} & \textbf{Domain} & \textbf{\# Samples} \\
          \toprule
            \multicolumn{5}{c}{\cellcolor[HTML]{D9D9D9}\textit{\textbf{Translationese}}} \\
            \textsc{eng} & \texttt{emotes\_3k\_eng\_seacrowd\_t2t} & Commonsense Reasoning & Ethics & 2000 \\
            \textsc{eng} & \texttt{aya\_evaluation\_suite\_eng\_seacrowd\_t2t} & Instruction Tuning & General & 400 \\
            \midrule
            \textsc{ind} & \texttt{belebele\_ind\_latn\_seacrowd\_qa} & QA & General & 1969 \\
            \textsc{ind} & \texttt{parallel\_asian\_treebank\_ind\_eng\_seacrowd\_t2t} & MT & News & 31 \\
            \textsc{ind} & \texttt{aya\_evaluation\_suite\_ind\_seacrowd\_t2t} & Instruction Tuning & General & 4 \\
            \textsc{ind} & \texttt{bactrian\_x\_id\_seacrowd\_t2t} & Instruction Tuning & Mixed, Multi-domain, Wikipedia & 1972 \\
\bottomrule
        \end{tabular}
        }
    \end{subtable}
    \caption{Test data used in the translationese classifier experiment.}
    \label{tab:translationese-vs-natural-data-test}
\end{table*}

\begin{table}[t]
    \centering
    \begin{subtable}[t]{\linewidth}
        \centering
        \resizebox{\linewidth}{!}{
\begin{tabular}{rcllr}
          \toprule
          \textbf{No.} & \textbf{ISO 639-3} & \textbf{Language} & \textbf{Region(s)} & \textbf{Population} \\
          \toprule
             \multicolumn{5}{c}{\cellcolor[HTML]{117b36}\textit{\textbf{\textcolor{white}{In SEACrowd}}}} \\
             1 & \textsc{ind} & Indonesian & Indonesia & <1B \\
            2 & \textsc{jav} & Javanese & Indonesia & <100M \\
            3 & \textsc{vie} & Vietnamese & Vietnam & <100M \\
            4 & \textsc{tha} & Thai & Thailand, Cambodia & <100M \\
            5 & \textsc{fil} & Filipino & Philippines & <100M \\
            6 & \textsc{mya} & Burmese & Myanmar & <100M \\
            7 & \textsc{sun} & Sunda & Indonesia & <100M \\
            8 & \textsc{tgl} & Tagalog & Philippines & <100M \\
            9 & \textsc{khm} & Khmer & Cambodia, Vietnam & <100M \\
            10 & \textsc{ceb} & Cebuano & Philippines & <100M \\
            11 & \textsc{tts} & Northeastern Thai & Thailand & <100M \\
            12 & \textsc{zlm} & Malay & Malaysia & <100M \\
            13 & \textsc{zsm} & Standard Malay & Malaysia, Brunei, Singapore & <100M \\             
        \bottomrule
        \end{tabular}
        }
    \end{subtable}
    \caption{SEA indigenous languages with $\geq$10M speakers.}
    \label{tab:sea-langs-under-100m-speakers}
\end{table}

\begin{table}[t]
    \centering
    \begin{subtable}[t]{\linewidth}
        \centering
        \resizebox{\linewidth}{!}{
        \begin{tabular}{rcllr}
          \toprule
          \textbf{No.} & \textbf{ISO 639-3} & \textbf{Language} & \textbf{Region(s)} & \textbf{Population} \\
          \toprule
             \multicolumn{5}{c}{\cellcolor[HTML]{117b36}\textit{\textbf{\textcolor{white}{In SEACrowd}}}} \\
             1 & \textsc{ilo} & Ilocano & Philippines & <10M \\
            2 & \textsc{mad} & Madura & Indonesia & <10M \\
            3 & \textsc{nod} & Northern Thai & Laos, Thailand & <10M \\
            4 & \textsc{hil} & Hiligaynon & Philippines & <10M \\
            5 & \textsc{min} & Minangkabau & Indonesia & <10M \\
            6 & \textsc{bug} & Bugis & Indonesia & <10M \\
\bottomrule
        \end{tabular}
        }
    \end{subtable}
    \caption{SEA indigenous languages with <10M speakers.}
    \label{tab:sea-langs-under-10m-speakers}
\end{table}

\begin{table}[t]
    \centering
    \begin{subtable}[t]{\linewidth}
        \centering
        \resizebox{\linewidth}{!}{
        \begin{tabular}{rcllr}
          \toprule
          \textbf{No.} & \textbf{ISO 639-3} & \textbf{Language} & \textbf{Region(s)} & \textbf{Population} \\
          \toprule
             \multicolumn{5}{c}{\cellcolor[HTML]{117b36}\textit{\textbf{\textcolor{white}{In SEACrowd}}}} \\
                1 & \textsc{nut} & Nung & Vietnam & <1M \\
                2 & \textsc{kac} & Jingpho & Myanmar & <1M \\
                3 & \textsc{tsg} & Tausug & Philippines & <1M \\
                4 & \textsc{nij} & Ngaju & Indonesia & <1M \\
                5 & \textsc{ljp} & Lampung Api & Indonesia & <1M \\
                6 & \textsc{mqy} & Manggarai & Indonesia & <1M \\
\bottomrule
        \end{tabular}
        }
    \end{subtable}
    \caption{(1/2) SEA indigenous languages with <1M speakers.}
    \label{tab:sea-langs-under-1m-speakers-1}
\end{table}

\begin{table}[t]
    \centering
    \begin{subtable}[t]{\linewidth}
        \centering
        \resizebox{\linewidth}{!}{
        \begin{tabular}{rcllr}
          \toprule
          \textbf{No.} & \textbf{ISO 639-3} & \textbf{Language} & \textbf{Region(s)} & \textbf{Population} \\
          \toprule
             \multicolumn{5}{c}{\cellcolor[HTML]{117b36}\textit{\textbf{\textcolor{white}{In SEACrowd}}}} \\
                71 & \textsc{btz} & Batak Alas-Kluet & Indonesia & <1M \\
                72 & \textsc{ctd} & Tedim Chin & Myanmar & <1M \\
                73 & \textsc{srv} & Southern Sorsoganon & Philippines & <1M \\
                74 & \textsc{abl} & Lampung Nyo & Indonesia & <1M \\
                75 & \textsc{dnw} & Western Dani & Indonesia & <1M \\
                76 & \textsc{ktp} & Kaduo & Laos & <1M \\
\bottomrule
        \end{tabular}
        }
    \end{subtable}
    \caption{(2/2) SEA indigenous languages with <1M speakers.}
    \label{tab:sea-langs-under-1m-speakers-2}
\end{table}

\begin{table}[t]
    \centering
    \begin{subtable}[t]{\linewidth}
        \centering
        \resizebox{\linewidth}{!}{
        \begin{tabular}{rcllr}
          \toprule
          \textbf{No.} & \textbf{ISO 639-3} & \textbf{Language} & \textbf{Region(s)} & \textbf{Population} \\
          \toprule
             \multicolumn{5}{c}{\cellcolor[HTML]{117b36}\textit{\textbf{\textcolor{white}{In SEACrowd}}}} \\
                1 & \textsc{adr} & Adonara & Indonesia & <100K \\
                2 & \textsc{sed} & Sedang & Vietnam & <100K \\
                3 & \textsc{blf} & Buol & Indonesia & <100K \\
                4 & \textsc{tbl} & Tboli & Philippines & <100K \\
                5 & \textsc{hre} & Hre & Vietnam & <100K \\
                6 & \textsc{rol} & Romblomanon & Philippines & <100K \\
\bottomrule
        \end{tabular}
        }
    \end{subtable}
    \caption{(1/5) SEA indigenous languages with <100K speakers.}
    \label{tab:sea-langs-under-100k-speakers-1}
\end{table}

\begin{table}[t]
    \centering
    \begin{subtable}[t]{\linewidth}
        \centering
        \resizebox{\linewidth}{!}{
        \begin{tabular}{rcllr}
          \toprule
          \textbf{No.} & \textbf{ISO 639-3} & \textbf{Language} & \textbf{Region(s)} & \textbf{Population} \\
          \toprule
             \multicolumn{5}{c}{\cellcolor[HTML]{117b36}\textit{\textbf{\textcolor{white}{In SEACrowd}}}} \\
                72 & \textsc{tiy} & Teduray & Philippines & <100K \\
                73 & \textsc{tjg} & Tunjung & Indonesia & <100K \\
                74 & \textsc{wmm} & Maiwa & Indonesia & <100K \\
                75 & \textsc{sdo} & Bukar-Sadong Bidayuh & Indonesia, Malaysia & <100K \\
                76 & \textsc{kyp} & Kang & Laos & <100K \\
                77 & \textsc{tvo} & Tidore & Indonesia & <100K \\
\bottomrule
        \end{tabular}
        }
    \end{subtable}
    \caption{(2/5) SEA indigenous languages with <100K speakers.}
    \label{tab:sea-langs-under-100k-speakers-2}
\end{table}

\begin{table}[t]
    \centering
    \begin{subtable}[t]{\linewidth}
        \centering
        \resizebox{\linewidth}{!}{
        \begin{tabular}{rcllr}
          \toprule
          \textbf{No.} & \textbf{ISO 639-3} & \textbf{Language} & \textbf{Region(s)} & \textbf{Population} \\
          \toprule
             \multicolumn{5}{c}{\cellcolor[HTML]{117b36}\textit{\textbf{\textcolor{white}{In SEACrowd}}}} \\
                143 & \textsc{aol} & Alor & Indonesia & <100K \\
                144 & \textsc{jmd} & Yamdena & Indonesia & <100K \\
                145 & \textsc{laa} & Southern Subanen & Philippines & <100K \\
                146 & \textsc{lmy} & Lamboya & Indonesia & <100K \\
                147 & \textsc{txe} & Totoli & Indonesia & <100K \\
                148 & \textsc{oyb} & Oy & Laos & <100K \\
\bottomrule
        \end{tabular}
        }
    \end{subtable}
    \caption{(3/5) SEA indigenous languages with <100K speakers.}
    \label{tab:sea-langs-under-100k-speakers-3}
\end{table}

\begin{table}[t]
    \centering
    \begin{subtable}[t]{\linewidth}
        \centering
        \resizebox{\linewidth}{!}{
        \begin{tabular}{rcllr}
          \toprule
          \textbf{No.} & \textbf{ISO 639-3} & \textbf{Language} & \textbf{Region(s)} & \textbf{Population} \\
          \toprule
             \multicolumn{5}{c}{\cellcolor[HTML]{117b36}\textit{\textbf{\textcolor{white}{In SEACrowd}}}} \\
                217 & \textsc{snl} & Sangil & Philippines & <100K \\
                218 & \textsc{tby} & Tabaru & Indonesia & <100K \\
                219 & \textsc{tea} & Temiar & Malaysia & <100K \\
                220 & \textsc{yli} & Angguruk Yali & Indonesia & <100K \\
                221 & \textsc{mej} & Meyah & Indonesia & <100K \\
                222 & \textsc{mbi} & Ilianen Manobo & Philippines & <100K \\
\bottomrule
        \end{tabular}
        }
    \end{subtable}
    \caption{(4/5) SEA indigenous languages with <100K speakers.}
    \label{tab:sea-langs-under-100k-speakers-4}
\end{table}

\begin{table}[t]
    \centering
    \begin{subtable}[t]{\linewidth}
        \centering
        \resizebox{\linewidth}{!}{
        \begin{tabular}{rcllr}
          \toprule
          \textbf{No.} & \textbf{ISO 639-3} & \textbf{Language} & \textbf{Region(s)} & \textbf{Population} \\
          \toprule
             \multicolumn{5}{c}{\cellcolor[HTML]{117b36}\textit{\textbf{\textcolor{white}{In SEACrowd}}}} \\
            290 & \textsc{szb} & Ngalum & Indonesia & <100K \\
            291 & \textsc{tbk} & Calamian Tagbanwa & Philippines & <100K \\
            292 & \textsc{tbw} & Tagbanwa & Philippines & <100K \\
            293 & \textsc{txx} & Tatana & Malaysia & <100K \\
            294 & \textsc{wnk} & Wanukaka & Indonesia & <100K \\
            295 & \textsc{yva} & Yawa & Indonesia & <100K \\
\bottomrule
        \end{tabular}
        }
    \end{subtable}
    \caption{(5/5) SEA indigenous languages with <100K speakers.}
    \label{tab:sea-langs-under-100k-speakers-5}
\end{table}

\begin{table}[t]
    \centering
    \begin{subtable}[t]{\linewidth}
        \centering
        \resizebox{\linewidth}{!}{
        \begin{tabular}{rcllr}
          \toprule
          \textbf{No.} & \textbf{ISO 639-3} & \textbf{Language} & \textbf{Region(s)} & \textbf{Population} \\
          \toprule
             \multicolumn{5}{c}{\cellcolor[HTML]{117b36}\textit{\textbf{\textcolor{white}{In SEACrowd}}}} \\
                1 & \textsc{xte} & Ketengban & Indonesia & <10K \\
                2 & \textsc{bna} & Bonerate & Indonesia & <10K \\
                3 & \textsc{bku} & Buhid & Philippines & <10K \\
                4 & \textsc{aws} & South Awyu & Indonesia & <10K \\
                5 & \textsc{woo} & Manombai & Indonesia & <10K \\
                6 & \textsc{asc} & Casuarina Coast Asmat & Indonesia & <10K \\
\bottomrule
        \end{tabular}
        }
    \end{subtable}
    \caption{(1/6) SEA indigenous languages with <10K speakers.}
    \label{tab:sea-langs-under-10k-speakers-1}
\end{table}

\begin{table}[t]
    \centering
    \begin{subtable}[t]{\linewidth}
        \centering
        \resizebox{0.945\linewidth}{!}{
        \begin{tabular}{rcllr}
          \toprule
          \textbf{No.} & \textbf{ISO 639-3} & \textbf{Language} & \textbf{Region(s)} & \textbf{Population} \\
          \toprule
             \multicolumn{5}{c}{\cellcolor[HTML]{117b36}\textit{\textbf{\textcolor{white}{In SEACrowd}}}} \\
                81 & \textsc{yac} & Pass Valley Yali & Indonesia & <10K \\
                82 & \textsc{yoy} & Yoy & Laos, Thailand & <10K \\
                83 & \textsc{and} & Ansus & Indonesia & <10K \\
                84 & \textsc{mxn} & Moi Kelim & Indonesia & <10K \\
                85 & \textsc{tlv} & Taliabu & Indonesia & <10K \\
                86 & \textsc{bty} & Bobot & Indonesia & <10K \\
\bottomrule
        \end{tabular}
        }
    \end{subtable}
    \caption{(2/6) SEA indigenous languages with <10k speakers.}
    \label{tab:sea-langs-under-10k-speakers-2}
\end{table}

\begin{table}[t]
    \centering
    \begin{subtable}[t]{\linewidth}
        \centering
        \resizebox{\linewidth}{!}{
        \begin{tabular}{rcllr}
          \toprule
          \textbf{No.} & \textbf{ISO 639-3} & \textbf{Language} & \textbf{Region(s)} & \textbf{Population} \\
          \toprule
             \multicolumn{5}{c}{\cellcolor[HTML]{117b36}\textit{\textbf{\textcolor{white}{In SEACrowd}}}} \\
                161 & \textsc{bpv} & Bian Marind & Indonesia & <10K \\
                162 & \textsc{gzn} & Gane & Indonesia & <10K \\
                163 & \textsc{dmr} & East Damar & Indonesia & <10K \\
                164 & \textsc{obk} & Southern Bontok & Philippines & <10K \\
                165 & \textsc{bzl} & Boano & Indonesia & <10K \\
                166 & \textsc{hbu} & Habun & East Timor & <10K \\
\bottomrule
        \end{tabular}
        }
    \end{subtable}
    \caption{(3/6) SEA indigenous languages with <10K speakers.}
    \label{tab:sea-langs-under-10k-speakers-3}
\end{table}

\begin{table}[t]
    \centering
    \begin{subtable}[t]{\linewidth}
        \centering
        \resizebox{\linewidth}{!}{
        \begin{tabular}{rcllr}
          \toprule
          \textbf{No.} & \textbf{ISO 639-3} & \textbf{Language} & \textbf{Region(s)} & \textbf{Population} \\
          \toprule
             \multicolumn{5}{c}{\cellcolor[HTML]{117b36}\textit{\textbf{\textcolor{white}{In SEACrowd}}}} \\
                 236 & \textsc{bdg} & Bonggi & Malaysia & <10K \\
                237 & \textsc{fau} & Fayu & Indonesia & <10K \\
                238 & \textsc{ilu} & Ili’uun & Indonesia & <10K \\
                239 & \textsc{yet} & Yetfa & Indonesia & <10K \\
                240 & \textsc{dmy} & Sowari & Indonesia & <10K \\
                241 & \textsc{ddw} & Dawera-Daweloor & Indonesia & <10K \\
\bottomrule
        \end{tabular}
        }
    \end{subtable}
    \caption{(4/6) SEA indigenous languages with <10K speakers.}
    \label{tab:sea-langs-under-10k-speakers-4}
\end{table}

\begin{table}[t]
    \centering
    \begin{subtable}[t]{\linewidth}
        \centering
        \resizebox{\linewidth}{!}{
        \begin{tabular}{rcllr}
          \toprule
          \textbf{No.} & \textbf{ISO 639-3} & \textbf{Language} & \textbf{Region(s)} & \textbf{Population} \\
          \toprule
             \multicolumn{5}{c}{\cellcolor[HTML]{D9D9D9}\textit{\textbf{Not in SEACrowd}}} \\
             320 & \textsc{krv} & Kavet & Cambodia & <10K \\
            321 & \textsc{cey} & Ekai Chin & Myanmar & <10K \\
            322 & \textsc{kjt} & Phrae Pwo Karen & Thailand & <10K \\
            323 & \textsc{kuk} & Kepo’ & Indonesia & <10K \\
            324 & \textsc{put} & Putoh & Indonesia & <10K \\
            325 & \textsc{rjg} & Rajong & Indonesia & <10K \\
\bottomrule
        \end{tabular}
        }
    \end{subtable}
    \caption{(5/6) SEA indigenous languages with <10K speakers.}
    \label{tab:sea-langs-under-10k-speakers-5}
\end{table}

\begin{table}[t]
    \centering
    \begin{subtable}[t]{\linewidth}
        \centering
        \resizebox{\linewidth}{!}{
        \begin{tabular}{rcllr}
          \toprule
          \textbf{No.} & \textbf{ISO 639-3} & \textbf{Language} & \textbf{Region(s)} & \textbf{Population} \\
          \toprule                        
             \multicolumn{5}{c}{\cellcolor[HTML]{D9D9D9}\textit{\textbf{Not in SEACrowd}}} \\
             400 & \textsc{kvh} & Komodo & Indonesia & <10K \\
            401 & \textsc{apf} & Pahanan Agta & Philippines & <10K \\
            402 & \textsc{bzb} & Andio & Indonesia & <10K \\
            403 & \textsc{jal} & Yalahatan & Indonesia & <10K \\
            404 & \textsc{mvr} & Marau & Indonesia & <10K \\
            405 & \textsc{agz} & Mt. Iriga Agta & Philippines & <10K \\
\bottomrule
        \end{tabular}
        }
    \end{subtable}
    \caption{(6/6) SEA indigenous languages with <10K speakers.}
    \label{tab:sea-langs-under-10k-speakers-6}
\end{table}

\begin{table}[t]
    \centering
    \begin{subtable}[t]{\linewidth}
        \centering
        \resizebox{\linewidth}{!}{
        \begin{tabular}{rcllr}
          \toprule
          \textbf{No.} & \textbf{ISO 639-3} & \textbf{Language} & \textbf{Region(s)} & \textbf{Population} \\
          \toprule
             \multicolumn{5}{c}{\cellcolor[HTML]{117b36}\textit{\textbf{\textcolor{white}{In SEACrowd}}}} \\
                1 & \textsc{sow} & Sowanda & Indonesia & <1K \\
                2 & \textsc{duv} & Duvle & Indonesia & <1K \\
                3 & \textsc{hmu} & Hamap & Indonesia & <1K \\
                4 & \textsc{ktt} & Ketum & Indonesia & <1K \\
                5 & \textsc{mpz} & Mpi & Thailand & <1K \\
                6 & \textsc{tvw} & Sedoa & Indonesia & <1K \\
\bottomrule
        \end{tabular}
        }
    \end{subtable}
    \caption{(1/3) SEA indigenous languages with <1K speakers.}
    \label{tab:sea-langs-under-1k-speakers-1}
\end{table}

\begin{table}[t]
    \centering
    \begin{subtable}[t]{\linewidth}
        \centering
        \resizebox{\linewidth}{!}{
        \begin{tabular}{rcllr}
          \toprule
          \textbf{No.} & \textbf{ISO 639-3} & \textbf{Language} & \textbf{Region(s)} & \textbf{Population} \\
          \toprule
             \multicolumn{5}{c}{\cellcolor[HTML]{117b36}\textit{\textbf{\textcolor{white}{In SEACrowd}}}} \\
                81 & \textsc{jel} & Yelmek & Indonesia & <1K \\
                82 & \textsc{nun} & Anong & Myanmar & <1K \\
                83 & \textsc{opk} & Kopkaka & Indonesia & <1K \\
                84 & \textsc{pas} & Papasena & Indonesia & <1K \\
                85 & \textsc{tmj} & Samarokena & Indonesia & <1K \\
                86 & \textsc{urn} & Uruangnirin & Indonesia & <1K \\
\bottomrule
        \end{tabular}
        }
    \end{subtable}
    \caption{(2/3) SEA indigenous languages with <1K speakers.}
    \label{tab:sea-langs-under-1k-speakers-2}
\end{table}

\begin{table}[t]
    \centering
    \begin{subtable}[t]{\linewidth}
        \centering
        \resizebox{\linewidth}{!}{
        \begin{tabular}{rcllr}
          \toprule
          \textbf{No.} & \textbf{ISO 639-3} & \textbf{Language} & \textbf{Region(s)} & \textbf{Population} \\
          \toprule
             \multicolumn{5}{c}{\cellcolor[HTML]{117b36}\textit{\textbf{\textcolor{white}{In SEACrowd}}}} \\
                161 & \textsc{ugo} & Ugong & Thailand & <1K \\
                162 & \textsc{wbe} & Waritai & Indonesia & <1K \\
                163 & \textsc{mra} & Mlabri & Laos, Thailand & <1K \\
                164 & \textsc{afz} & Obokuitai & Indonesia & <1K \\
                165 & \textsc{mgf} & Maklew & Indonesia & <1K \\
                166 & \textsc{ttn} & Towei & Indonesia & <1K \\
\bottomrule
        \end{tabular}
        }
    \end{subtable}
    \caption{(3/3) SEA indigenous languages with <1K speakers.}
    \label{tab:sea-langs-under-1k-speakers-3}
\end{table}

\begin{table}[t]
    \centering
    \begin{subtable}[t]{\linewidth}
        \centering
        \resizebox{\linewidth}{!}{
        \begin{tabular}{rcllr}
          \toprule
          \textbf{No.} & \textbf{ISO 639-3} & \textbf{Language} & \textbf{Region(s)} & \textbf{Population} \\
          \toprule
             \multicolumn{5}{c}{\cellcolor[HTML]{117b36}\textit{\textbf{\textcolor{white}{In SEACrowd}}}} \\
                1 & \textsc{mnu} & Mer & Indonesia & <100 \\
                2 & \textsc{itx} & Itik & Indonesia & <100 \\
                3 & \textsc{kxq} & Smärky Kanum & Indonesia & <100 \\
                4 & \textsc{lix} & Liabuku & Indonesia & <100 \\
                5 & \textsc{awr} & Awera & Indonesia & <100 \\
                6 & \textsc{bdx} & Budong-Budong & Indonesia & <100 \\
\bottomrule
        \end{tabular}
        }
    \end{subtable}
    \caption{SEA indigenous languages with <100 speakers.}
    \label{tab:sea-langs-under-100-speakers}
\end{table}

\end{document}